%% file: main.tex
\documentclass{article}

\PassOptionsToPackage{numbers, compress}{natbib}
\usepackage{ailab2026}

\input{preamble}

\title{On-Policy Distillation with \Name-Policy Rollouts}

\author{%
Jaehui Hwang\quad Dongyoon Han \quad Sangdoo Yun \quad Byeongho Heo\footnotemark[2] \\
  \\NAVER AI Lab
}

\begin{document}

\maketitle

{\renewcommand{\thefootnote}{\fnsymbol{footnote}}%
\footnotetext[2]{\scriptsize Corresponding author}}

\begin{abstract}
On-policy distillation (OPD) has been widely studied as a post-training method in which a student model obtains token-level supervision from a stronger teacher on its own rollouts.
Recent studies have improved OPD through alternative distillation reward formulations and teacher configurations, while the objective of distillation remains centered on mimicking the teacher.
However, when a stronger teacher has limited distributional overlap with the student, such positive guidance can provide insufficient learning signals.
In this work, we introduce \oursfullbf (\textbf{\ours}), which complements teacher supervision with rollouts from a lower-performing, lower-capability \name policy that serves as a negative reference for the student.
Rather than modifying the distillation reward formulation, \ours introduces the \name policy at the rollout stage, continuously supplying tokens preferred by the \name policy over the teacher so that they remain exposed to teacher supervision throughout training.
This provides an explicit negative signal through \name-policy rollouts while preserving the positive teacher supervision used in OPD.
Through extensive experiments, we show that \ours improves OPD across model scales, generation modes, reasoning domains, and different OPD variants.
Furthermore, our analyses show that \ours effectively suppresses tokens preferred by the \name policy over the teacher and moves the student away from the \name policy.
These results support our design of introducing negative signals through \name-policy rollouts and provide new insight into the role of the rollout policy in OPD. Code will be available
at \url{https://github.com/naver-ai/np-opd}.
\end{abstract}

\section{Introduction}
As large language models (LLMs) have rapidly improved, they have been widely utilized for diverse applications, such as daily chat, information search, and solving complex problems.
While pre-training through next-token prediction remains important and allows LLMs to acquire broad knowledge, the increasingly diverse usage of LLMs has led to the development of various post-training methods to tune models toward specific goals.
These methods further adapt models using additional supervision beyond pre-training.
Among representative approaches, supervised fine-tuning (SFT) trains models on curated prompt-response pairs, which are often generated by stronger LLMs~\citep{mukherjee2023orca,s1,openthought}.
Direct preference optimization (DPO) commonly uses pairs of responses with relative preferences, learning from both responses to favor one and suppress the other~\citep{rafailov2023direct}.
On the other hand, reinforcement learning (RL) methods, such as GRPO, optimize the model's own generations according to reward signals~\citep{rlhf,shao2024deepseekmath}.
More recently, on-policy distillation (OPD) has been actively studied as an alternative to RL-based post-training with the token-level training signal obtained through distillation from a teacher model~\citep{agarwal2024policy,lu2025onpolicydistillation,yang2025qwen3,xiao2026mimov2}.

Knowledge distillation~\citep{hinton2015distilling} has been applied to LLM post-training in various ways, particularly by varying how training sequences are obtained.
SeqKD~\citep{kim2016sequence} utilizes teacher-generated sequences for distillation, commonly referred to as supervised fine-tuning (SFT) in LLM post-training.
Although SeqKD plays an important role in transferring teacher knowledge, it has limitations in preserving the student's on-policy capabilities.
To address this issue, OPD performs distillation on student-generated sequences, similar to RL, to preserve the student's capabilities during distillation.
Recent studies have mainly focused on improving the token-level distillation signal through alternative loss functions or teacher configurations~\citep{yang2026learning,heo2026policy,feng2026weak,wu2026lightning2}.
Despite these advances, the objective of distillation has remained unchanged: \textit{mimicking the teacher model}.

The teacher model serves as the only positive role model in distillation.
When the student's distribution sufficiently overlaps with that of the teacher, the positive-only guidance provides enough signal for on-policy distillation.
However, distillation often employs a substantially stronger teacher whose distribution can differ from that of the student.
As a result, limited overlap between the teacher and student distributions provides insufficient supervision for OPD, as observed in previous studies~\citep{fu2026revisiting,li2026rethinking}.
In DPO~\citep{rafailov2023direct}, negative samples complement positive-sample supervision and are frequently sampled from a weaker model \citep{geng2025delta, olmo2025olmo}.
Motivated by DPO, we introduce a negative policy as a complementary learning signal for OPD.
The negative policy provides an additional source of supervision for undesirable outputs, potentially complementing the limited teacher supervision in low-overlap regions.

\begin{figure}[t]
  \centering
    \includegraphics[width=\textwidth]{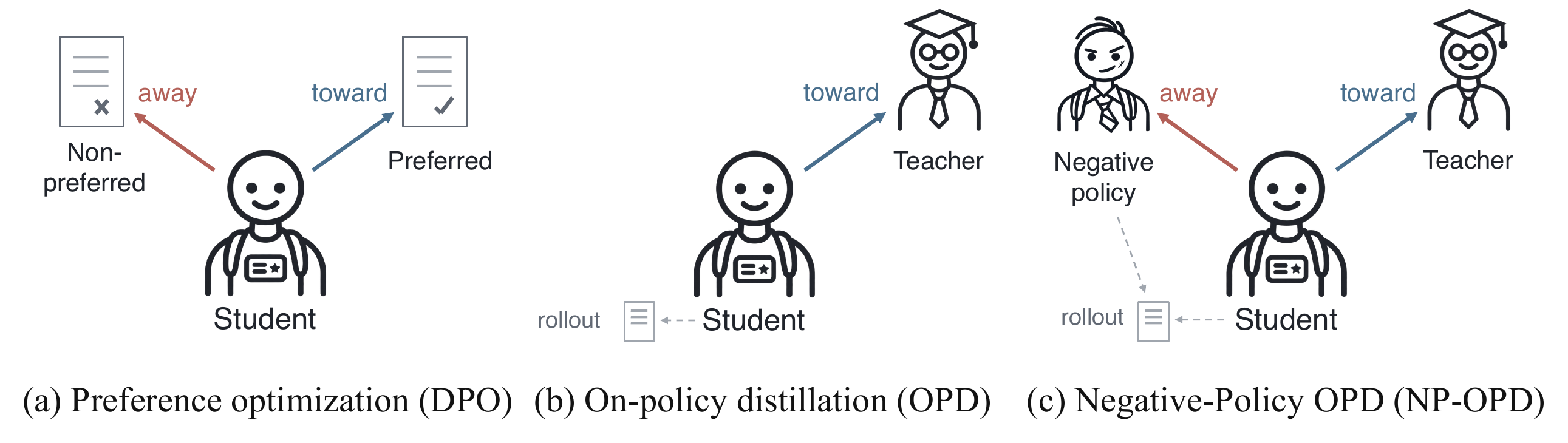}
    \vspace{-1em}
    \caption{
    \textbf{Conceptual comparison of preference optimization, on-policy distillation (OPD), and \oursfull (\ours).}
    (a) Preference optimization provides both toward and away directions through preferred and non-preferred responses, whereas (b) OPD provides an explicit positive direction through teacher supervision but no explicit negative reference indicating what the student should move away from.
    (c) \ours complements the positive signal from teacher supervision with rollouts from a lower-performing, lower-capability \name policy that serves as a negative reference.
    Here, ``\name'' refers to the role of the \name policy as a reference from which the student is intended to move away.
    In our main setting, the \name policy is selected from the same model family as the student with lower model capacity and overall reasoning performance.
    }
  \label{fig:concept}
  \vspace{-1em}
\end{figure}

We propose \oursfull (\ours) to introduce a negative policy into an on-policy distillation framework (see \cref{fig:concept}).
Existing OPD variants often modify the reward design to achieve their respective objectives.
Thus, introducing the negative policy through the reward limits its compatibility with recent advances in OPD~\citep{yang2026learning,heo2026policy}.
Instead, we introduce the negative policy to the rollout stage.
In \ours, some tokens are sampled from the negative policy and repeatedly optimized using the distillation reward.
Consequently, suppression is more concentrated on negative tokens, defined as tokens preferred by the negative policy over the teacher, than under on-policy rollouts.
In this way, \ours enhances the suppression of these tokens while preserving the original reward direction of OPD.
Therefore, \ours can be combined with recent advances in OPD to further improve their performance.

In our main setting, we choose the \name policy from the same model family as the student, with lower model capacity and lower overall reasoning performance.
To validate the effectiveness of \ours, we conduct extensive experiments using different student scales and both thinking and non-thinking modes, and further evaluate it on 13 benchmarks from three domains, math, science, and code.
\ours consistently improves OPD and remains effective when combined with recent OPD variants such as ExOPD \citep{yang2026learning} and OPD$^2$ \citep{heo2026policy}.
This demonstrates that the benefit of \ours is not tied to a specific OPD formulation and can complement existing OPD methods without modifying their reward formulations.
Beyond empirical validation, our analyses show that \ours operates as intended by concentrating suppression on \name tokens and reducing the measured overlap with the \name policy.
These effects are associated with larger performance gains, explaining why \ours improves OPD.
As an additional practical benefit, \ours can reuse pre-generated \name-policy rollouts, reducing the cost of rollout generation during training.
Furthermore, we discuss a new perspective on the success of recent OPD variants. We also show that degrading student rollouts improves OPD but is less effective than using a separate \name policy.



\section{Preliminary}
\label{sec:preliminary}

\subsection{On-Policy Distillation}
\label{sec:opd}

On-policy distillation (OPD) trains a student model $\pi_\theta$ using trajectories generated by the student itself and token-level supervision from a stronger teacher $\pi^*$~\citep{agarwal2024policy,lu2025onpolicydistillation,yang2026learning,heo2026policy}. Given a question $x\sim\mathcal{D}$, the student generates a rollout
\begin{equation}
    y\sim\pi_\theta(\cdot\mid x).
    \label{eq:opd-rollout}
\end{equation}
At each token position $t$, the teacher and student distributions are conditioned on the same prefix $(x,y_{<t})$.
As an alternative to reinforcement learning (RL)-based post-training, OPD can be implemented in an RL-style framework using sampled-token distillation signals~\citep{heo2026policy}.
For a sampled token $y_t$, the token-level signal is given by
\begin{equation}
R_t^{\mathrm{OPD}}
=
\log \pi^*(y_t\mid x,y_{<t})
-
\log \pi_\theta(y_t\mid x,y_{<t}).
\label{eq:opd-reward}
\end{equation}
This token-level signal adjusts the student toward the teacher distribution.
Although individual token probabilities can increase or decrease, the teacher remains the sole reference that determines the update direction.
The resulting RL-style update direction is written as
\begin{equation}
\nabla_\theta J_{\mathrm{OPD}}(\theta)
=
\mathbb{E}_{x\sim\mathcal{D},\,y\sim\pi_\theta(\cdot\mid x)}
\left[
\sum_{t=1}^{T}
R_t^{\mathrm{OPD}}
\nabla_\theta
\log \pi_\theta(y_t\mid x,y_{<t})
\right].
\label{eq:opd-gradient}
\end{equation}
Recent OPD variants modify this token-level reward in different ways~\citep{yang2026learning,heo2026policy}.
We provide their detailed formulations in Appendix~\ref{app:opd-variants}.

\subsection{Learning from Positive and Negative Signals}
\label{sec:dpo}

Several post-training methods train models using not only outputs that should be preferred (or correct), but also those that should be avoided (or incorrect) \citep{rafailov2023direct, shao2024deepseekmath, meng2024simpo, geng2025delta}.
For example, Direct Preference Optimization (DPO) \citep{rafailov2023direct} uses a pair of preferred and rejected responses $(y^+,y^-)$ with the following objective:
\begin{equation}
    \mathcal{L}_{\mathrm{DPO}}
    =
    -\mathbb{E}
    \left[
        \log\sigma
        \left(
            \beta
            \left[
                \log\frac{\pi_\theta(y^+\mid x)}{\pi_{\mathrm{ref}}(y^+\mid x)}
                -
                \log\frac{\pi_\theta(y^-\mid x)}{\pi_{\mathrm{ref}}(y^-\mid x)}
            \right]
        \right)
    \right].
    \label{eq:dpo}
\end{equation}
The preferred response provides a direction to move toward, while the rejected response explicitly identifies a direction to move away from.

A similar structure appears in Reinforcement Learning (RL). Group Relative Policy Optimization (GRPO)~\citep{shao2024deepseekmath} uses group-relative advantages, with the policy-gradient component
\begin{equation}
    \nabla_\theta J(\theta)
    \propto
    \mathbb{E}
    \left[
        \sum_i
        A_i
        \nabla_\theta \log\pi_\theta(y_i\mid x)
    \right],
    \label{eq:grpo-signal}
\end{equation}
where responses with positive advantage are reinforced, while those with negative advantage are suppressed. In correctness-based reasoning tasks, this typically corresponds to reinforcing correct responses and suppressing incorrect ones.

OPD follows a different structure. 
As illustrated in \cref{fig:concept}, the teacher provides a single direction that the student should move toward, but does not explicitly indicate what the student should move away from.
This contrast raises a natural question: \emph{can such a negative signal benefit OPD without changing its original teacher supervision?}

\section{Method}
\label{sec:method}

This section introduces \oursfull (\ours), which injects negative signals into OPD via the rollout policy and addresses the question raised in \cref{sec:preliminary}.
\ours augments student-generated trajectories with trajectories from an additional rollout source and applies teacher supervision to both. 
We present the formulation of \ours and explain how changing the rollout policy provides a direction for the student to move away from.

\subsection{Distillation with \Name-Policy rollouts}
\label{sec:negative-rollout}

As motivated in \cref{sec:preliminary}, we introduce \oursfullbf (\textbf{\ours}), which incorporates negative signal into OPD through the rollout distribution.
Specifically, we modify the rollout distribution to include both student-generated trajectories and trajectories sampled from an additional \name policy, $\pi_n$.
The \textit{\name policy} refers to a policy that the student needs to move away from during training, similar to non-preferred responses in DPO.
In our main setting, we choose $\pi_n$ from the same model family as the student, with lower model capacity and lower overall reasoning performance.
Each training batch consists of trajectories sampled from the student policy and the \name policy, with proportions $1-\alpha$ and $\alpha$, respectively:
\begin{equation}
\pi_{z_x} =
\begin{cases}
\pi_\theta, & z_x=0,\\
\pi_n, & z_x=1,
\end{cases}
\label{eq:mixed-rollout}
\end{equation}
where $\alpha\in[0,1]$ controls the proportion of \name-policy trajectories.

The token-level distillation signal defined in \cref{sec:opd} is applied to trajectories sampled from both rollout policies. The resulting RL-style update direction can be written as

\begin{equation}
\nabla_\theta J_{\mathrm{NP}}(\theta)
=
\mathbb{E}_{\substack{
x\sim\mathcal{D},\,
z_x\sim\mathrm{Bernoulli}(\alpha),\,
y\sim\pi_{z_x}(\cdot\mid x)
}}
\left[
\sum_{t=1}^{|y|}
R_t^{\mathrm{OPD}}
\nabla_\theta
\log \pi_\theta(y_t\mid x,y_{<t})
\right].
\label{eq:np-opd-objective}
\end{equation}

In each term, the sampled token $y_t$ and its prefix $y_{<t}$ are drawn from the corresponding rollout policy.
Thus, \ours changes the trajectory distribution exposed to teacher supervision, while the token-level teacher-student supervision remains unchanged.
When $\alpha=0$, \ours reduces to OPD with student-generated trajectories, while $\alpha=1$ uses only \name-policy trajectories.


\subsection{How \Name-Policy rollouts Provide Negative Signals}
\label{sec:negative-signal}

\ours introduces the negative policy into the rollout model. In addition to the on-policy model, the negative model is selected as the rollout model with probability $\alpha$.
This rollout implementation is designed to improve applicability to reward variants $R^{OPD}_t$ for various recent advances in OPD studies~\citep{yang2026learning,heo2026policy}. 
However, it is not straightforward how the negative policy on the rollouts makes the student model move away from the negative policy.
In this section, we explain this mathematically.

We expand the expectation in Eq.~\ref{eq:np-opd-objective} and consider the expected gradient contribution of the $t$-th token in a trajectory $y=(y_1,\ldots,y_T)$. The original OPD and \ours have different sampling probabilities for $y$. Thus, the gradients of its $t$-th token with respect to the expected OPD are given by
\begin{equation}
g_{\mathrm{OPD}}(y,t)
=
\underbrace{
\prod_{i=1}^{t}
\pi_\theta(y_i\mid x,y_{<i})
}_{\text{sampling probability}}
R_t^{\mathrm{OPD}}
\nabla_\theta
\log \pi_\theta(y_t\mid x,y_{<t})
\label{eq:opd-token-contribution}
\end{equation}
for the original OPD. On the other hand, \ours has a mixed sampling probability between the on-policy and negative policy models, which is denoted as
\begin{equation}
\small
g_{\mathrm{NP}}(y,t)
=
\underbrace{
\left[
(1-\alpha)
\prod_{i=1}^{t}
\pi_\theta(y_i\mid x,y_{<i})
+
\alpha
\prod_{i=1}^{t}
\pi_n(y_i\mid x,y_{<i})
\right]
}_{\text{sampling probability}}
R_t^{\mathrm{OPD}}
\nabla_\theta
\log \pi_\theta(y_t\mid x,y_{<t}).
\label{eq:ours-token-contribution}
\end{equation}
The difference between these two gradient contributions quantifies the gradient changes induced by sampling from the negative policy:
\begin{equation}
\small
g_{\mathrm{NP}}-g_{\mathrm{OPD}}
={}
\alpha
\left[
\prod_{i=1}^{t}
\pi_n(y_i\mid x,y_{<i})
-
\prod_{i=1}^{t}
\pi_\theta(y_i\mid x,y_{<i})
\right]
R_t^{\mathrm{OPD}}
\nabla_\theta
\log \pi_\theta(y_t\mid x,y_{<t}).
\label{eq:ours-opd-contribution-difference}
\end{equation}
Compared to OPD, \ours enhances the learning signal for negative policy trajectories, and the degree of enhancement is proportional to the probability $\alpha$.
The enhanced sampling of \ours keeps the on-policy model away from the negative policy.
This is valid under the following assumption:
\begin{equation}
D_{\mathrm{KL}}\!\left(\pi_n\|\pi^*\right)
>
D_{\mathrm{KL}}\!\left(\pi_n\|\pi_\theta\right).
\label{eq:negative-policy-assumption}
\end{equation}
The Kullback-Leibler (KL) divergence between the teacher $\pi^*$ and the negative policy $\pi_n$ is greater than that between the student $\pi_\theta$ and the negative policy.
Since we select a model with weaker performance than the student as a negative policy, this is a reasonable assumption in practice.
Under this assumption, the expected value of OPD reward in Eq.~\ref{eq:opd-reward} is presented as
\begin{equation}
\begin{aligned}
\mathbb{E}_{y_t\sim\pi_n}
\left[
R_t^{\mathrm{OPD}}
\right]
&=
\mathbb{E}_{y_t\sim\pi_n}
\left[
\log \pi^*(y_t)
-
\log \pi_\theta(y_t)
\right]
\\
&=
\mathbb{E}_{y_t\sim\pi_n}
\left[
\bigl(\log \pi^*(y_t)-\log \pi_n(y_t)\bigr)
-
\bigl(\log \pi_\theta(y_t)-\log \pi_n(y_t)\bigr)
\right]
\\
&=
-
D_{\mathrm{KL}}(\pi_n\|\pi^*)
+
D_{\mathrm{KL}}(\pi_n\|\pi_\theta)
<0.
\end{aligned}
\label{eq:negative-opd-signal}
\end{equation}
In short, the tokens generated by the negative policy have negative expected OPD rewards. 
This suggests that the OPD reward trains the student not to generate tokens from the negative policy.
Combined with \cref{eq:ours-opd-contribution-difference}, \ours increases the contribution of trajectories from the negative policy and then suppresses these tokens from being generated by the student on average.
Thus, \ours pushes the student away from the negative policy solely through rollout modification.

There are two notable aspects in interpreting \ours under this formulation.
First, \ours selectively suppresses tokens generated by the negative policy according to the teacher's preference.
While the formulation shows that \ours statistically suppresses the negative policy's rollouts, this doesn't mean that every token sampled from the negative policy is suppressed.
At the token level, tokens preferred by the teacher are still enhanced even when they have high probability under the negative policy.
Since $R^{\textrm{OPD}}$ in Eq.~\ref{eq:np-opd-objective} remains the OPD reward, \ours preserves the direction of OPD signal rather than explicitly reversing or overriding it.
Although \ours statistically trains the student away from the negative policy, it preserves the original training dynamics of OPD.
Second, this formulation naturally extends to advanced distillation rewards~\citep{yang2026learning,heo2026policy}.
Although these rewards are not directly formulated by KL divergence, tokens sampled from the \name policy are generally expected to receive lower rewards than those from the student.
Therefore, the negative expected reward in Eq.~\ref{eq:negative-opd-signal} is expected to hold for these reward variants, allowing \ours to be applied to advanced OPD methods without modifying their reward functions.

\section{Experiment and Analysis}
\label{sec:4}

We evaluate \ours across various student scales, generation modes, and OPD reward formulations, and further validate its effectiveness on diverse reasoning benchmarks, examining whether \name-policy rollouts consistently improve OPD.
We also examine the practical training efficiency of reusing pre-generated \name-policy rollouts.
Finally, to validate that \ours follows the mechanism described in \cref{sec:negative-signal}, we analyze whether \ours selectively suppresses tokens favored by the \name policy over the teacher and reduces the student's overlap with the \name policy.

\subsection{Experimental Setup}
\label{sec:4-1}

\noindent{\textbf{Models.}}
We use Qwen3-1.7B, Qwen3-4B, and Qwen3-8B~\citep{yang2025qwen3} as student models, and conduct experiments with both thinking and non-thinking modes.
Teachers are selected according to the scale of the student model and generation mode.
For thinking mode, we use Qwen3-8B as the teacher for Qwen3-1.7B, and Qwen3-30B-A3B-Thinking for Qwen3-4B and Qwen3-8B.
For non-thinking mode, we use Qwen3-4B-Instruct for Qwen3-1.7B and Qwen3-30B-A3B-Instruct for Qwen3-4B and Qwen3-8B.
We additionally evaluate \ours on the Gemma-4 family \citep{team2026gemma}, using Gemma-4-E4B-it as the student and Gemma-4-12B-it as the teacher, to examine whether the observed gains extend beyond the Qwen3 family.

\input{figs/nothink_full}

\noindent{\textbf{Training and evaluation.}}
The training set contains Math \citep{moshkov2025aimo}, Science \citep{opensciencereasoning2}, and Code \citep{ahmad2025opencodereasoning} questions mixed at a 1:1:1 ratio, and a total of 30k questions are used.
In all experiments, we use a fixed batch size of 256, a learning rate of $5\times10^{-6}$, a maximum rollout length of 16k tokens, and 100 training steps unless otherwise specified.
We vary the rollout mixing ratio as $\alpha \in \{0.0, 0.25, 0.5, 0.75, 1.0\}$ for OPD.
Note that $\alpha=0.0$ uses only on-policy rollouts (OPD), while $\alpha=1.0$ uses only \name-policy rollouts.
All rollouts are sampled with a temperature of 1.0.
For evaluation, we use a temperature of 0.7 and a maximum generation length of 32k tokens.
We evaluate reasoning performance on seven math benchmarks, three code benchmarks, and three science benchmarks.
Further details are provided in Appendix~\ref{app:detail}.

\noindent{\textbf{Rollout policy.}}
For our main experiments, we use Qwen3-0.6B, Qwen3-1.7B, and Qwen3-4B as \name-policy models for Qwen3-1.7B, Qwen3-4B, and Qwen3-8B students, respectively. For Gemma-4-E4B-it, we utilize Gemma-4-E2B-it as the \name-policy model.
We also vary the scale of the rollout model to examine the effect of rollout policy capability.

\subsection{Main Results}
\label{sec:3-2}

In non-thinking mode, \ours consistently improves over OPD across model scales and reasoning domains.
For Qwen3-1.7B with $\alpha=1.0$, \ours increases the average performance from 48.3 to 56.3 on Math, from 27.2 to 31.1 on Code, and from 37.2 to 41.5 on Science.
Similar gains are observed for Qwen3-4B, where the averages improve from 62.2 to 70.5 on Math, from 40.2 to 51.1 on Code, and from 47.8 to 51.2 on Science.
Across individual benchmarks, \ours improves performance in most settings, with particularly large gains on AIME, HMMT25, LCBv5, and RGAlgo.
The same trend extends to Qwen3-8B and to thinking-mode generation, with thinking-mode results summarized in \cref{tab:avg_reward} and full benchmark-wise results provided in Appendix~\ref{app:full}.

\noindent{\textbf{\ours with various OPD reward formulations.}}
We also test \ours with recent OPD variants that modify the token-level reward, including ExOPD~\citep{yang2026learning} and OPD$^2$~\citep{heo2026policy}.
As shown in \cref{tab:avg_reward}, \ours improves overall performance across all three formulations for Qwen3-1.7B and Qwen3-4B in thinking mode.
For Qwen3-1.7B, \ours improves the overall average by 1.98 points with OPD and 1.76 points with OPD$^2$, while maintaining comparable performance with ExOPD.
For Qwen3-4B, \ours improves all three formulations, with gains of 1.11, 1.86, and 2.10 points for OPD, ExOPD, and OPD$^2$, respectively.
This suggests that the benefit of \ours is not limited to a specific student scale or OPD reward formulation.
We report representative results in \cref{tab:avg_reward}, with full results provided in Appendix~\ref{app:full}.

\noindent{\textbf{\ours on Gemma.}}
We additionally evaluate \ours on the Gemma family to examine whether its effectiveness extends beyond Qwen3.
As shown in \cref{tab:gemma}, \ours with $\alpha=1.0$ improves the average performance of Gemma-4-E4B over OPD from 63.0 to 66.4 on Math and from 49.1 to 49.8 on Science.
On Code, although both methods remain below the base model's performance of 56.5, \ours still outperforms the on-policy OPD baseline, improving the average from 54.9 to 55.2.
Overall, \ours is also effective on Gemma, with clear gains on Math and Science.

\input{figs/reward_full}
\input{figs/gemma_full}



\noindent{\textbf{Training efficiency.}}
Training efficiency is a practical advantage of \ours.
\Name-policy rollouts can be pre-generated and reused throughout training, unlike student-generated on-policy rollouts.\footnote{We plan to release the pre-generated \name-policy rollouts with our code.}
For Qwen3-1.7B on a single 8$\times$H100 node, \ours with $\alpha=1$ reduces standard OPD training time from 9h19m to 3h34m in thinking mode, excluding rollout pre-generation.
Additional details for different values of $\alpha$ are provided in Appendix~\ref{app:chracter}.

\input{figs/gemma}

\subsection{Understanding the Effectiveness of \ours}
\label{sec:analysis}

In the previous section, we empirically showed that \ours consistently improves OPD across different student models, generation modes, and OPD variants.
We now examine whether these gains are consistent with our hypothesis that supplying \name-policy rollouts provides a negative signal during distillation.
Then, we consider two complementary questions.
First, does training suppress the intended tokens, i.e., those preferred by the \name policy over the teacher?
Second, does this targeted suppression actually reduce the student's overlap with the \name policy?
For both analyses, we compare rollout policies with different capabilities.


\noindent{\textbf{Suppressing the intended tokens.}}
We first test whether training selectively suppresses tokens associated with the negative policy.
We define a metric, \emph{\name-policy suppression precision} (NSP), as the fraction of suppressed tokens that are preferred by the \name policy over the teacher.
We evaluate each trained student on a fixed set of trajectories generated by the initial student.
For this analysis, we use Qwen3-1.7B as the student and Qwen3-8B as the teacher in thinking mode, with Qwen3-0.6B as the default \name policy.
We define NSP as
\begin{equation}
\mathrm{NSP}
=
P\!\left(
\log \pi_n(y_t\mid s_t)
>
\log \pi^*(y_t\mid s_t)
\,\middle|\,
t\in\mathcal{D}_q
\right),
\label{eq:nsp}
\end{equation}
where $\mathcal{D}_q$ contains positions with decreased log-probability among the top $q\%$ by absolute change from the initial student.
A higher $\mathrm{NSP}$ indicates that the largest probability reductions are more selectively concentrated on tokens preferred by the \name policy over the teacher.

As shown in \cref{fig:nsp}, \ours using Qwen3-0.6B as the \name policy achieves higher $\mathrm{NSP}$ than OPD using only on-policy rollouts.
This indicates that suppression during distillation is more concentrated on tokens preferred by the \name policy over the teacher, supporting the role of these rollouts as a negative reference and remaining consistent with the intended mechanism described in \cref{sec:negative-signal}.

Furthermore, $\mathrm{NSP}$ is highly correlated with average accuracy across OPD, \ours, and alternative rollout settings using Qwen3-1.7B, Qwen3-4B, and Qwen3-8B.
Using larger rollout models with higher capability and performance than the student, including the teacher, yields lower $\mathrm{NSP}$, indicating that suppression is less concentrated on negative tokens.
As shown in \cref{fig:sweep}, only the lower-capability \name policy improves over on-policy OPD, consistent with our intended use of its rollouts as a negative reference.
This also suggests that changing the rollout source alone does not explain the gains of \ours.

\input{figs/anal_ld}

\noindent{\textbf{Reducing overlap with the \name policy.}}
While NSP measures where suppression is concentrated, we next examine whether \ours reduces the student's overlap with the \name policy.
We define \emph{\name-policy overlap reduction} (NOR) as

\begin{equation}
\mathrm{NOR}(i;\pi_A)=
\frac{1}{|\mathcal{X}^{(i)}|}
\sum_{t\in\mathcal{X}^{(i)}}
\left[\Delta\pi(v_T^*\mid s_t)-\Delta\pi(v_n^*\mid s_t)\right],
\label{eq:nor}
\end{equation}
where $i$ indexes trajectories, $s_t=(x,y_{<t})$, and $\Delta\pi=\pi_A-\pi_{S_0}$ denotes the probability change after training. The tokens $v_n^*$ and $v_T^*$ are the top predictions of the \name policy and teacher at $s_t$, respectively, and $\mathcal{X}^{(i)}$ contains positions where they differ.
Here, $\pi_{S_0}$ and $\pi_A$ denote the student before and after training, respectively, and overlap reduction refers to a teacher-relative shift in top-token probabilities, rather than an absolute reduction in distributional overlap.
A larger NOR indicates a greater shift in the student's probability gap toward the teacher's top prediction relative to the \name policy's at disagreement positions.

For the main analysis, we use a fixed set of trajectories generated by the initial student from prompts used during training.
\cref{fig:away-a} (a) compares NOR under OPD and \ours for these trajectories.
Each point represents one trajectory, with the horizontal and vertical axes corresponding to \ours and OPD, respectively.
Most points lie below the diagonal, with a mean NOR of 0.019 under \ours compared with 0.005 under OPD, indicating a greater relative shift toward the teacher.
In contrast, \cref{fig:away-a} (b) shows that using teacher rollouts during training yields a mean NOR of $-0.030$, below that of OPD.
Additional results on held-out prompts and other rollout models show similar trends and are provided in Appendix~\ref{app:analysis}.
These results support our design of using \name-policy rollouts to induce the intended shift away from the \name policy.

These analyses address the two questions of whether \ours suppresses tokens preferred by the \name policy and whether it reduces the student's overlap with the \name policy.
In both cases, \ours operates as intended, consistent with the mechanism described in \cref{sec:negative-signal}.
These results show that \ours is supported not only by empirical performance gains but also by direct analyses of the effects induced by \name-policy rollouts.

\input{figs/bow_2}

\section{Discussion}
\label{sec:discussion}


\noindent{\textbf{A negative-signal perspective on recent OPD methods.}}
As shown in \cref{fig:nsp-reward}, ExOPD and OPD$^2$ achieve higher NSP than standard OPD even with on-policy rollouts.
This provides a different perspective on recent OPD variants that modify the distillation signal.
ExOPD defines its reward relative to a reference model, the teacher's base model, while OPD$^2$ explicitly uses the difference between the teacher and its pretrained base model.
This interpretation suggests that recent OPD methods may already benefit from such implicit negative signals even without explicitly targeting them.
Even with these signals, \ours further increases NSP when combined with each method, together with additional gains in average accuracy.
Thus, the negative signal introduced through \name-policy rollouts can complement the implicit negative signals in these OPD methods.

\begin{wraptable}{r}{0.53\textwidth}
    \centering
    \vspace{-12pt}
    \tabcolsep=0.15em
    \caption{\textbf{Performance with other degraded rollouts.}}
    \label{tab:synthetic-negative}
    \vspace{-.8em}
    \resizebox{\linewidth}{!}{
    \begin{tabular}{lccccc}
        \toprule
        Rollout policy
        & AIME24 & AIME25 & GPQA & LCB & \textbf{AVG} \\
        \midrule
        On-policy
        & 50.0 & 35.6 & 49.1 & 36.9 & 42.9 \\
        + high temp.
        & 52.7 & 37.1 & 48.3 & 37.1 & 43.8 \\
        + persona prompt
        & 52.1 & 36.5 & 51.2 & 35.9 & 43.9 \\
        + layer drop
        & 49.0 & 35.8 & 51.0 & 32.5 & 42.1 \\
        \midrule
        \textbf{\ours}
        & \textbf{54.2} & \textbf{38.5} & \textbf{51.4}
        & \textbf{41.0} & \textbf{46.3} \\
        \bottomrule
    \end{tabular}
    }
    \vspace{-12pt}
\end{wraptable}
\noindent{\textbf{Can negative rollouts be synthesized from the student?}}
Our method uses rollouts from a separate, lower-capability \name policy as a negative reference for the student.
Using Qwen3-1.7B as the student and Qwen3-8B as the teacher in thinking mode, we examine whether similar negative rollouts can be synthesized from the student by increasing the decoding temperature to 1.5, adding a persona prompt that asks it to solve the problem at an elementary-school level, or dropping three layers during rollout generation.
As shown in \cref{tab:synthetic-negative}, some of these modifications improve over standard on-policy rollouts, suggesting that degrading the rollout process can provide useful complementary signals.
However, their gains remain substantially smaller than those obtained with a separate lower-capability \name policy.
In particular, temperature scaling and persona prompting improve the average score by 0.9 and 1.0 points, respectively, whereas the \name policy improves the average by 3.4 points.
These results suggest that simple degradation of the student can partially synthesize useful negative rollouts, but does not match the benefit of using a separate lower-capability policy to provide a move-away direction in OPD.

\section{Conclusion}
In this work, we introduced \oursfull (\ours), which complemented positive teacher guidance with rollouts from a lower-capability \name policy serving as a negative reference for the student.
\ours introduced the \name policy at the rollout stage while preserving the original distillation reward formulation.
Across model scales, generation modes, reasoning domains, and different OPD variants, \ours improved performance over on-policy baselines.
Our analyses showed that \ours concentrated suppression on tokens preferred by the \name policy over the teacher and reduced the measured overlap with the \name policy.
Reusing pre-generated \name-policy rollouts also reduced rollout generation costs during training.
These findings provided a new perspective on the role of the rollout policy in OPD, showing how negative-policy rollouts complemented positive teacher guidance without modifying the distillation reward.

\clearpage
\bibliographystyle{plainnat}
\bibliography{references}

\clearpage
\appendix
\counterwithin{table}{section}
\counterwithin{figure}{section}

\section*{Appendix}

\section{Related Work}

\subsection{Post-Training for LLMs}
Post-training adapts pretrained LLMs to instruction following and reasoning through additional supervision.
Supervised fine-tuning (SFT) trains models on curated prompt-response pairs, often using responses generated by stronger models~\citep{mukherjee2023orca,s1,openthought}.
Preference-based methods learn from comparisons between responses.
RLHF uses a learned reward model to guide policy optimization~\citep{rlhf}, while DPO directly optimizes the relative likelihood of preferred and rejected responses with respect to a reference policy~\citep{rafailov2023direct}.
For reasoning tasks, reinforcement learning with verifiable rewards uses task-specific feedback, such as answer correctness, to improve model-generated responses~\citep{shao2024deepseekmath, yu2026dapo}.
These methods illustrate complementary roles for positive and negative signals, encouraging preferred behavior while discouraging less desirable outputs.
Our work studies a related role for negative references in OPD, introducing them through the rollout policy while preserving the existing distillation reward.

\subsection{On-Policy Distillation}
Knowledge distillation (KD) transfers the predictive behavior of a teacher model to a student~\citep{hinton2015distilling}.
For autoregressive generation, the source of training sequences distinguishes two common forms of distillation.
Sequence-KD uses teacher-generated sequences~\citep{kim2016sequence}, which is now commonly realized as SFT on model-generated responses~\citep{s1,openthought}.
OPD instead generates trajectories from the student and obtains token-level supervision from the teacher~\citep{agarwal2024policy,gu2023minillm,ko2024distillm}.
This reduces the train-inference mismatch of teacher-generated supervision and has shown competitive performance for reasoning post-training~\citep{lu2025onpolicydistillation,yang2025qwen3,xiao2026mimov2}.

Recent studies extend OPD along several directions.
A major line modifies the distillation signal, including uncertainty-aware weighting~\citep{jin2026entropy}, extrapolated teacher signals~\citep{yang2026learning}, and the teacher-to-base delta used in OPD$^2$~\citep{heo2026policy}.
Other studies modify the trajectories used during distillation.
Lightning OPD precomputes teacher supervision on fixed SFT rollouts to enable efficient offline training~\citep{wu2026lightning}, while Relay-OPD lets the teacher temporarily take over failed student prefixes to construct more reliable training trajectories~\citep{xu2026pass}.
These approaches mainly introduce alternative trajectories to improve supervision quality or training efficiency.
In contrast, we study a different role of the rollout policy: rollouts from the \name policy are intentionally supplied as a negative reference, similar in role to rejected responses in preference optimization. Our work revisits the rollout source as a means of supplying negative signals and analyzes how it shapes token-level updates under the existing distillation reward.

\section{OPD Reward Formulations}
\label{app:opd-variants}

In our experiments, we additionally use two variants of OPD to examine the effectiveness of \ours across different reward formulations.
Here, we describe the token-level reward formulations of ExOPD~\citep{yang2026learning} and OPD$^2$~\citep{heo2026policy}, following the notation introduced in \cref{sec:opd}.
For compactness, we denote the token context by $s_t=(x,y_{<t})$.
As described in \cref{sec:opd}, standard OPD uses
\begin{equation}
R_t^{\mathrm{OPD}}
=
\log \pi^*(y_t\mid s_t)
-
\log \pi_\theta(y_t\mid s_t).
\end{equation}

\paragraph{ExOPD.}
ExOPD~\citep{yang2026learning} introduces a reference policy $\pi_{\mathrm{ref}}$ and extrapolates the reward from the reference toward the teacher.
Its token-level reward can be written as
\begin{equation}
\begin{aligned}
R_t^{\mathrm{ExOPD}}
={}&
\lambda
\left[
\log \pi^*(y_t\mid s_t)
-
\log \pi_{\mathrm{ref}}(y_t\mid s_t)
\right]
\\
&-
\left[
\log \pi_\theta(y_t\mid s_t)
-
\log \pi_{\mathrm{ref}}(y_t\mid s_t)
\right],
\end{aligned}
\label{eq:exopd-reward}
\end{equation}
where $\lambda$ controls the degree of reward extrapolation.
Equivalently,
\begin{equation}
R_t^{\mathrm{ExOPD}}
=
R_t^{\mathrm{OPD}}
+
(\lambda-1)
\left[
\log \pi^*(y_t\mid s_t)
-
\log \pi_{\mathrm{ref}}(y_t\mid s_t)
\right].
\end{equation}
Thus, $\lambda=1$ recovers standard OPD, while $\lambda>1$ extrapolates the reward along the direction from the reference policy toward the teacher.
Following the main setting of ExOPD, we use the teacher's base model as the reference policy $\pi_{\mathrm{ref}}$ in our experiments.

\paragraph{OPD$^2$.}
OPD$^2$~\citep{heo2026policy} instead uses the probability change from the teacher's base model to the post-trained teacher.
Denoting the corresponding base model by $\pi^*_{\mathrm{base}}$, the delta reward is
\begin{equation}
R_t^{\Delta}
=
\log \pi^*(y_t\mid s_t)
-
\log \pi^*_{\mathrm{base}}(y_t\mid s_t).
\label{eq:opd2-delta}
\end{equation}
OPD$^2$ centers both the original OPD reward and the delta reward under the current student distribution:
\begin{align}
A_t^{\mathrm{OPD}}
&=
R_t^{\mathrm{OPD}}
-
\mathbb{E}_{\tilde y_t\sim\pi_\theta(\cdot\mid s_t)}
\left[
R_t^{\mathrm{OPD}}(\tilde y_t)
\right],
\\
A_t^{\Delta}
&=
R_t^{\Delta}
-
\mathbb{E}_{\tilde y_t\sim\pi_\theta(\cdot\mid s_t)}
\left[
R_t^{\Delta}(\tilde y_t)
\right].
\end{align}
The final OPD$^2$ signal is
\begin{equation}
A_t^{\mathrm{OPD}^2}
=
\begin{cases}
A_t^{\Delta},
&
A_t^{\Delta}A_t^{\mathrm{OPD}}>0,
\\
0,
&
\text{otherwise}.
\end{cases}
\label{eq:opd2-reward}
\end{equation}
This conditioning applies the delta signal only when its update direction agrees with that of standard OPD.

\section{Details and Additional Results of Section \ref{sec:4}}
\label{app:full}


\subsection{Training and Evaluation Details}
\label{app:detail}
\noindent{\textbf{Training data and optimization.}}
We train on 30K reasoning prompts with the corresponding system prompts.
Unless otherwise specified, all models are trained for 100 steps with a global batch size of 256.
We use AdamW with a learning rate of $5\times10^{-6}$ and cosine decay with a minimum learning-rate ratio of 0.1.
We use a warmup ratio of 0.1, gradient clipping at 1.0, bf16 precision, and ZeRO-3.
Rollouts are generated with a maximum length of 16,384 tokens and a sampling temperature of 1.0.
For ExOPD and OPD$^2$, we use the base model corresponding to each teacher, following their respective formulations.
We set the KL regularization coefficient $\beta$ to 0 for Qwen models and 0.04 for Gemma models.

\noindent{\textbf{Additional filtering and regularization.}}
In the non-thinking mode of Qwen, we observe that some rollouts terminate with very short answers.
Since such responses provide little token-level supervision for distillation, we filter rollouts shorter than 20 tokens from training.
For Gemma, we observe substantially less stable OPD training than for Qwen.
We therefore apply KL regularization with $\beta=0.04$ for all Gemma experiments.

\noindent{\textbf{Infrastructure.}}
Qwen experiments are conducted on a single node with eight NVIDIA H100 GPUs, with vLLM used concurrently for rollout generation.
For Qwen3-8B, we use tensor parallelism with a tensor-parallel size of 4.
Gemma experiments are conducted on two nodes.

\noindent{\textbf{Evaluation.}}
To evaluate reasoning performance, we utilize AIME24~\citep{AIME2024}, AIME25~\citep{AIME2025}, AMC23~\citep{AMC}, HMMT25~\citep{dekoninck2026matharena}, MATH500~\citep{hendrycks2021}, OlympiadBench~\citep{he2024olympiad}, and ReasoningGym Math (RGMath)~\citep{stojanovski2025reasoninggym} for math. For coding, we use CodeForces~\citep{openthought}, LiveCodeBench~\citep{jain2025livecodebench}, and ReasoningGym Algorithm (RGAlgo)~\citep{stojanovski2025reasoninggym}. For science, we use GPQA~\citep{rein2023gpqa}, SuperGPQA~\citep{pteam2025supergpqascalingllmevaluation}, and SciBench~\citep{wang2024scibench}.
We report avg@16 for AIME24 and AIME25, avg@8 for AMC23, HMMT25, MATH500, and GPQA, and avg@3 for the others.
We use Evalchemy \citep{Evalchemy} and \texttt{lm-evaluation-harness} \citep{eval-harness} for benchmark evaluation, depending on the benchmark implementation.
We use the same evaluation implementation and decoding configuration across all compared methods.

\subsection{Full Results of Section \ref{sec:3-2}}
\label{app:full_results}

For \cref{tab:nothink_math_results}, \cref{tab:avg_reward}, \cref{tab:qwen_summary_nothink}, and \cref{tab:qwen_summary_think}, we report a single representative $\alpha$ per setting, preferring the value that improves all three domains over the on-policy baseline and, among such values, the one with the highest overall average.
Across the evaluated nonzero $\alpha$ values, \ours generally improves the overall average over the corresponding on-policy baseline in both thinking and non-thinking modes.
Full results for all evaluated $\alpha$ values are provided below.

\noindent{\textbf{Full results of non-thinking mode with OPD variants.}}
\cref{tab:qwen_summary_nothink} summarizes the non-thinking results, showing that \ours improves domain-average performance in nearly all combinations of student scale, reasoning domain, and OPD reward formulation.
\cref{tab:nothink_math_results_v2_opd,tab:nothink_code_science_results_v2_opd} provide benchmark-wise results for OPD, \cref{tab:nothink_math_results_v2_exopd,tab:nothink_code_science_results_v2_exopd} for ExOPD, and \cref{tab:nothink_math_results_v2_opd2,tab:nothink_code_science_results_v2_opd2} for OPD$^2$.
Larger $\alpha$ generally yields stronger results, particularly for OPD and ExOPD, while mixed rollouts also achieve competitive or better performance in some OPD$^2$ settings.
These results show that incorporating \name-policy rollouts benefits multiple OPD reward formulations, with gains observed at different values of $\alpha$.

\input{figs/summary_nothink}
\input{figs/nothink_rewards}
\clearpage

\noindent{\textbf{Full results of thinking mode with OPD variants.}}
\cref{tab:qwen_summary_think} summarizes the thinking-mode results, showing broad improvements with \ours across student scales, reasoning domains, and OPD reward formulations.
\cref{tab:think_math_results_v2_opd,tab:think_code_science_results_v2_opd} provide benchmark-wise results for OPD, \cref{tab:think_math_results_v2_opd2,tab:think_code_science_results_v2_opd2} for OPD$^2$, and \cref{tab:think_code_science_results_v2_exopd,tab:think_math_results_v2_exopd} for ExOPD.
Gains are observed with both mixed and fully \name-policy rollouts, indicating that the benefit is not limited to a single value of $\alpha$.
The effect of increasing $\alpha$ varies across settings, with larger values often benefiting Math and Science and intermediate values sometimes yielding stronger Code results.

\input{figs/summary_think}
\input{figs/think_rewards}

\clearpage

\subsection{Additional Experiments}
\label{app:chracter}

\noindent{\textbf{Training efficiency.}}
\cref{tab:training-efficiency} reports the wall-clock training time of Qwen3-1.7B on a single node with 8 H100 GPUs.
The \name-policy rollouts are generated in advance, whereas the student portion of each batch requires on-policy rollout generation during training.
Accordingly, the training time generally decreases as $\alpha$ increases.
For standard OPD, using only \name-policy rollouts ($\alpha=1$) reduces the training time from 9h19m to 3h34m in thinking mode and from 5h15m to 1h16m in non-thinking mode.

\input{figs/training_efficiency}

\noindent{\textbf{Generation length.}
As shown in \cref{tab:generation-length}, on-policy training with OPD, ExOPD, and OPD$^2$ increases generation length relative to the base models on both AIME24 and AIME25.
Compared with on-policy training, \ours reduces generation length for Qwen3-4B and Qwen3-8B across all three reward formulations, while slightly increasing it for Qwen3-1.7B.

\begin{table}[h]
\centering
\caption{
\textbf{Mean generation length (in thousands of tokens) in thinking mode.}
Entries for OPD variants show training with on-policy rollouts ($\alpha$=0.0) $\rightarrow$ \ours with $\alpha$=1.0.
}
\label{tab:generation-length}
\small
\begin{tabular}{llccc}
\toprule
Benchmark & Method & Qwen3-1.7B & Qwen3-4B & Qwen3-8B \\
\midrule
AIME24 & Base & 17.0 & 14.2 & 15.1 \\
& OPD      & $20.9 \to 21.5$ & $18.5 \to 15.3$ & $17.0 \to 15.1$ \\
& ExOPD    & $20.7 \to 21.1$ & $18.7 \to 16.0$ & $18.1 \to 15.9$ \\
& OPD$^2$  & $20.1 \to 21.1$ & $19.0 \to 16.6$ & $18.6 \to 16.1$ \\
\midrule
AIME25 & Base & 18.0 & 18.1 & 18.3 \\
& OPD      & $22.9 \to 23.5$ & $20.5 \to 18.3$ & $19.9 \to 18.0$ \\
& ExOPD    & $22.4 \to 22.9$ & $20.7 \to 18.7$ & $20.6 \to 18.6$ \\
& OPD$^2$  & $22.4 \to 22.7$ & $21.1 \to 19.2$ & $20.7 \to 18.8$ \\
\bottomrule
\end{tabular}
\end{table}

\noindent{\textbf{Model merging.}}
For Qwen3-8B in thinking mode, the best rollout mixing ratio varies across domains.
Motivated by this observation, we merge models trained with $\alpha=0$ and $\alpha=1$ by averaging their parameters with equal weights.
For ExOPD, \cref{tab:think8b_merge_math_exopd,tab:think8b_merge_other_exopd} show improvements over the on-policy baseline from 74.2 to 76.3 on Math, from 60.3 to 60.7 on Code, and from 53.6 to 54.7 on Science.
For OPD$^2$, \cref{tab:think8b_merge_math_opd2,tab:think8b_merge_other_opd2} show similar improvements, from 75.0 to 77.4 on Math, from 61.4 to 62.3 on Code, and from 52.5 to 54.2 on Science.
These results suggest that model merging can combine some of the benefits of using on-policy rollouts and \name rollouts.

\input{figs/merge}


\subsection{Details on Section \ref{sec:analysis}}
\label{app:analysis}

\noindent{\textbf{Details of NSP analysis.}}
We compute NSP using changes in the student's log-probability assigned to each realized token relative to the initial student.
Within each trajectory, we select the top 5\% of token positions by absolute log-probability change and retain only those with negative changes.
NSP is the fraction of retained positions where the \name policy (Qwen3-0.6B) assigns higher probability to the token than the teacher model (Qwen3-8B).
To clearly contrast different rollout sources, we analyze \ours and alternative rollout settings with $\alpha=1$ and compare them with the on-policy baseline ($\alpha=0$).
NSP values are averaged over three repetitions of analysis-trajectory sampling, with standard errors of the mean (SEMs) ranging from 0.0004 to 0.0033 across settings.
Error bars are omitted from the figures for visual clarity.

\noindent{\textbf{Details of NOR analysis.}}
We use a fixed set of 300 training prompts, with 100 prompts from each of Math, Code, and Science.
The negative reference and teacher are fixed to Qwen3-0.6B and Qwen3-8B, respectively, so their top predictions and the disagreement positions are shared across all settings.
To clearly contrast different rollout sources, we analyze \ours and alternative rollout settings with $\alpha=1$ and compare them with the on-policy baseline ($\alpha=0$).

\noindent{\textbf{NOR analysis with different rollouts.}}
We extend the NOR analysis to rollout models of different sizes within the Qwen3 family.
As shown in \cref{app:away-ladder}, the mean NOR decreases from 0.019 with Qwen3-0.6B rollouts to $-0.014$, $-0.023$, and $-0.030$ with Qwen3-1.7B, Qwen3-4B, and Qwen3-8B rollouts, respectively.
Only the lower-capability \name policy yields a higher mean NOR than on-policy OPD (0.005).
This pattern supports our use of lower-capability rollouts as a negative reference and suggests that changing the rollout source alone does not explain the observed shift.

\input{figs/heldin_full}

\noindent{\textbf{NOR analysis on held-out prompts.}}
We further evaluate NOR on held-out prompts to examine whether the reduction in overlap with the \name policy also appears beyond the prompts used during training.
We use 300 prompts from nine benchmarks that are not included in the training data: 120 math prompts, with 30 each from AIME24, AIME25, HMMT, and OlympiadBench; 120 science prompts, with 45 from GPQA-Diamond, 40 from SciBench, and 35 from SuperGPQA; and 60 code prompts, with 35 from CodeForces and 25 from RGAlgorithmic.
Using the same model configuration and checkpoints as in \cref{sec:analysis}, we generate one trajectory per prompt from the initial student, yielding 300 trajectories in total.
As shown in \cref{app:away-heldout}, the same trend observed on training prompts also holds for these held-out examples.
Most trajectories lie below the diagonal, indicating greater relative overlap reduction under NP-OPD than under OPD on held-out prompts.
These results show that the effect is not limited to the prompts used during training and generalizes to unseen benchmark examples.

\input{figs/heldout}

\section{Details of Section \ref{sec:discussion}}
\label{app:discussion}

\noindent{\textbf{Persona prompt. }}For the persona-prompt setting in \cref{sec:discussion}, we use the following prompt to instruct the student model to respond as an elementary school student:

\begin{tcolorbox}[
    title={Persona prompt},
    colback=gray!8,
    colbacktitle=gray!15,
    coltitle=black,
    colframe=gray!30,
    boxrule=0.4pt,
    arc=2pt,
    fonttitle=\bfseries,
    fontupper=\small,
    left=8pt,
    right=8pt,
    top=6pt,
    bottom=6pt
]
You are an elementary school student. You have only learned basic school material: simple arithmetic, fractions, basic shapes, and everyday science facts. You do not know advanced methods like algebra tricks, calculus, formal proofs, or real programming, only very simple ideas. Think and write the way a young student would: use simple words, take small easy steps, and always pick the easiest method you know. If a problem seems too hard, just make your best guess using what you know.
\end{tcolorbox}

\noindent{\textbf{Layer drop.}} For the layer-drop setting in \cref{sec:discussion}, we remove the last three transformer layers from the student model.

\end{document}

%% file: preamble.tex
\usepackage[T1]{fontenc}

\usepackage{microtype}  
\usepackage{url}        
\usepackage{xspace}     

\usepackage{amsmath}
\usepackage{amsfonts}   
\usepackage{amssymb}

\usepackage{graphicx}
\usepackage{subcaption}
\graphicspath{{figs/}}  

\usepackage{booktabs}   
\usepackage{multirow}
\usepackage{makecell}
\usepackage{nicefrac}   

\usepackage{enumitem}

\usepackage{hyperref}
\definecolor{darkblue}{rgb}{0, 0, 0.5}

\hypersetup{colorlinks=true, linkcolor=darkblue, citecolor=darkblue, urlcolor=darkblue}
\usepackage{cleveref}

\crefname{section}{Sec.}{Secs.}
\Crefname{section}{Section}{Sections}
\crefname{figure}{Fig.}{Figs.}
\Crefname{figure}{Figure}{Figures}
\crefname{table}{Tab.}{Tabs.}
\Crefname{table}{Table}{Tables}
\crefname{equation}{Eq.}{Eqs.}
\Crefname{equation}{Equation}{Equations}

\usepackage{lipsum}

\usepackage{kotex}
\usepackage{wrapfig}

\newcommand\name{{negative}\xspace}
\newcommand\Name{{Negative}\xspace}
\newcommand\ours{{NP-OPD}\xspace}
\newcommand\oursfull{{Negative-Policy OPD}\xspace}
\newcommand\oursfullbf{{\textbf{N}egative-\textbf{P}olicy \textbf{OPD}}\xspace}
\usepackage{tcolorbox}

\usepackage{tikz}
\usepackage{tikz,amsmath}
\usetikzlibrary{arrows.meta}
\usepackage{pgfplots}
\pgfplotsset{compat=1.18}
\usepgfplotslibrary{fillbetween}
\definecolor{selfc}{HTML}{9AA0A6}
\definecolor{weakc}{HTML}{1F5FA9}

\usepackage[table]{xcolor}
\definecolor{ngreenlight}{HTML}{e5ebfb}   

\definecolor{ink}{HTML}{1A1F24}\definecolor{mut}{HTML}{8A929B}
\definecolor{red1}{HTML}{B3261E}\definecolor{redbg}{HTML}{FDECEA}
\definecolor{blue1}{HTML}{1F5FA8}\definecolor{gfill}{HTML}{D9DDE3}

\newsavebox{\tblbox}\newsavebox{\figbox}

\definecolor{ink}{HTML}{1A1F24}  \definecolor{mut}{HTML}{8A929B}
\definecolor{mag}{HTML}{B03AA0}

\usepackage{pgfplots}
\pgfplotsset{compat=1.18}
\usetikzlibrary{arrows.meta,decorations.pathmorphing}
\usepgfplotslibrary{groupplots}

\definecolor{ink}{HTML}{22262B}   
\definecolor{mut}{HTML}{6E767E}   
\definecolor{trd}{HTML}{B4BBC2}   
\definecolor{lad}{HTML}{C2562A}   
\definecolor{rew}{HTML}{1E5F80}   

\pgfplotsset{
  ldacc/.style={
    axis line style   = {mut!50, line width=0.8pt},
    major tick style  = {mut!50, line width=0.8pt},
    tick align        = outside,
    tick pos          = left,
    grid              = major,
    grid style        = {mut!20, line width=0.6pt},
    label style       = {font=\footnotesize, color=ink},
    tick label style  = {font=\footnotesize, color=mut},
    xlabel near ticks, ylabel near ticks,
    scaled x ticks=false,
    xticklabel style={/pgf/number format/.cd, fixed, fixed zerofill, precision=2},
    legend style      = {font=\footnotesize, at={(0.5,1.02)}, anchor=south,
                         legend columns=3, draw=none, fill=none,
                         /tikz/every even column/.append style={column sep=14pt}},
    legend cell align = left,
    clip              = false,
  },
  ldbar/.style={
    ybar=0.7pt, bar width=0.32cm, legbar,
    xmin=0.45, xmax=3.55, ymin=0, ymax=80,
    xtick={1,2,3}, xticklabels={1.7B,4B,8B},
    ytick={0,20,40,60,80},
    axis line style  = {mut!50, line width=0.8pt},
    major tick style = {mut!50, line width=0.8pt},
    tick align=outside, tick pos=left,
    ymajorgrids, grid style={mut!20, line width=0.6pt},
    label style      = {font=\footnotesize, color=ink},
    tick label style = {font=\footnotesize, color=mut},
    title style      = {font=\footnotesize\bfseries, color=ink, yshift=1pt},
    ylabel style     = {align=center, font=\footnotesize, color=ink},
    legend style     = {font=\footnotesize, draw=none, fill=none,
                        legend columns=3,
                        /tikz/every even column/.append style={column sep=12pt}},
    legend cell align=left,
  },
  barbase/.style={fill=mut!32, draw=none},
  barself/.style={fill=rew!80, draw=none},
  barweak/.style={fill=lad,    draw=none},
  ladder/.style={
    lad!45, line width=1.8pt,
    mark=*, mark size=3.1pt,
    mark options={solid, fill=lad, draw=white, line width=1.0pt},
    error bars/x dir=both,
    error bars/x explicit,
    error bars/error bar style={lad!75, line width=1.1pt},
    error bars/error mark options={rotate=90, mark size=2.2pt, line width=1.1pt, lad!75},
  },
  selfmark/.style={
    only marks, mark=*, mark size=2pt,
    mark options={solid, fill=white, draw=rew, line width=1.5pt},
    error bars/x dir=both,
    error bars/x explicit,
    error bars/error bar style={rew!70, line width=1.1pt},
    error bars/error mark options={rotate=90, mark size=2.2pt, line width=1.1pt, rew!70},
  },
  weakmark/.style={
    only marks, mark=square*, mark size=3.0pt,
    mark options={solid, fill=lad, draw=white, line width=1.0pt},
    error bars/x dir=both,
    error bars/x explicit,
    error bars/error bar style={lad!75, line width=1.1pt},
    error bars/error mark options={rotate=90, mark size=2.2pt, line width=1.1pt, lad!75},
  },
}

\tikzset{ybreakwave/.style={decoration={snake, amplitude=0.8pt, segment length=6.0pt}}}

\tikzset{
  gaplead/.style={lad!55!black, line width=0.5pt, dotted},
  gaparr/.style ={lad!85!black, line width=0.7pt,
                  {Stealth[length=2.6pt,width=2.2pt]}-{Stealth[length=2.6pt,width=2.2pt]}},
}

\tikzset{
  lb/.style  = {font=\scriptsize, color=ink, inner sep=1.6pt},          
  lbs/.style = {font=\scriptsize, color=rew!80!black, inner sep=1.6pt},  
  lbw/.style = {font=\scriptsize, color=lad!80!black, inner sep=1.6pt},  
  lbo/.style = {font=\scriptsize\bfseries, color=lad!80!black,           
                inner sep=1.6pt},
  an/.style  = {font=\scriptsize, color=mut, inner sep=1.6pt},          
  dlt/.style = {font=\tiny\bfseries, color=lad!85!black, inner sep=1pt}, 
  dltn/.style= {font=\tiny\bfseries, color=mut!90!black, inner sep=1pt}, 
  nav/.style = {font=\tiny, color=mut!80, inner sep=1pt},               
  pairlink/.style={mut!60, line width=1.0pt, shorten <=5pt, shorten >=5pt,
                   -{Stealth[length=5pt,width=3.6pt]}},
  spot/.style= {circle, fill=lad!16, minimum size=17pt, inner sep=0pt},
}

\definecolor{lad1}{HTML}{EFB392}   
\definecolor{lad2}{HTML}{DC8352}   
\definecolor{lad3}{HTML}{C2562A}   
\pgfplotsset{
  ldframe/.style={
    axis line style  = {mut!50, line width=0.8pt},
    major tick style = {mut!50, line width=0.8pt},
    tick align=outside, tick pos=left,
    label style      = {font=\footnotesize, color=ink},
    tick label style = {font=\footnotesize, color=mut},
    title style      = {font=\footnotesize\bfseries, color=ink, yshift=1pt},
    ylabel style     = {align=center, font=\footnotesize, color=ink},
    legend style     = {font=\footnotesize, draw=none, fill=none, legend columns=-1,
                        /tikz/every even column/.append style={column sep=12pt}},
    legend cell align=left,
  },
  legbar/.style={legend image code/.code={\draw[##1,draw=none] (0cm,-0.085cm) rectangle (0.30cm,0.085cm);}},
  legmark/.style={legend image code/.code={\draw[##1] plot coordinates {(0.15cm,0cm)};}},
  gain/.style={
    ldframe, ybar=0.8pt, bar width=0.90cm, legbar,
    xmin=0.45, xmax=3.55, ymin=-1.9, ymax=10.2,
    ytick={0,2,4,6,8,10},
    xtick={1,2,3}, xticklabels={OPD,ExOPD,OPD$^2$},
    nodes near coords, point meta=y,
    every node near coord/.append style={font=\tiny\bfseries, color=ink!85,
        inner sep=1.6pt, /pgf/number format/.cd, fixed, fixed zerofill, precision=1, print sign},
    nodes near coords align={vertical},
  },
  barsC/.style={
    ldbar, ybar=1.2pt, bar width=0.44cm, legbar,
    ymin=25, ymax=71, ytick={30,40,50,60,70},
  },
  dumb/.style={
    ldframe, legmark,
    ymin=0.4, ymax=9.6, ytick={1,2,3,4,5,6,7,8,9},
    xmajorgrids, grid style={mut!20, line width=0.6pt},
    y tick label style={font=\scriptsize, color=ink},
    major tick style={draw=none},
  },
}
\tikzset{
  link/.style   = {lad!32, line width=3.0pt, line cap=round},
  linkneg/.style= {mut!35, line width=3.0pt, line cap=round},
  gtag/.style   = {font=\tiny\bfseries, color=lad!85!black, inner sep=1.5pt},
  gtagn/.style  = {font=\tiny\bfseries, color=mut!90!black, inner sep=1.5pt},
}

%% file: figs/nothink_full.tex
\begin{table}[t]
\centering
\caption{\textbf{Results of Qwen3 models on math, code, and science benchmarks (non-thinking mode).} The best and second-best results are shown in bold and underlined, respectively.}
\label{tab:nothink_math_results}
\small
\setlength{\tabcolsep}{5pt}
\resizebox{\textwidth}{!}{
\begin{tabular}{lcccccccc}
\toprule
& \multicolumn{8}{c}{\textbf{Math}} \\
\textbf{Model} & AIME24 & AIME25 & AMC23 & HMMT25 & MATH500 & Olympiad & RGMath & \textbf{Avg} \\
\midrule
Qwen3-1.7B   & \textcolor{gray}{11.5} & \textcolor{gray}{10.0} & \textcolor{gray}{45.9} & \textcolor{gray}{5.4} & \textcolor{gray}{69.0} & \textcolor{gray}{25.9} & \textcolor{gray}{82.5} & \textcolor{gray}{35.7} \\
\quad + OPD   & \underline{31.7} & \underline{21.7} & \underline{65.3} & \underline{13.3} & \underline{80.4} & \underline{36.0} & \underline{89.9} & \underline{48.3} \\
\rowcolor{ngreenlight}
\quad + \ours   & \textbf{43.1} & \textbf{33.3} & \textbf{78.4} & \textbf{20.0} & \textbf{86.0} & \textbf{40.6} & \textbf{92.3} & \textbf{56.3} \\
\midrule
Qwen3-4B   & \textcolor{gray}{24.2} & \textcolor{gray}{20.0} & \textcolor{gray}{69.4} & \textcolor{gray}{11.7} & \textcolor{gray}{79.0} & \textcolor{gray}{34.7} & \textcolor{gray}{88.9} & \textcolor{gray}{46.8} \\
\quad + OPD   & \underline{53.8} & \underline{42.3} & \underline{90.0} & \underline{26.2} & \underline{88.2} & \underline{43.7} & \underline{90.9} & \underline{62.2} \\
\rowcolor{ngreenlight}
\quad + \ours   & \textbf{67.9} & \textbf{58.8} & \textbf{95.3} & \textbf{43.8} & \textbf{90.6} & \textbf{45.5} & \textbf{91.7} & \textbf{70.5} \\
\bottomrule
\end{tabular}
\label{tab:qwen_nothink_math}
}

\setlength{\tabcolsep}{8.5pt}
\resizebox{\textwidth}{!}{
\begin{tabular}{lcccc|cccc}
\toprule
& \multicolumn{4}{c|}{\textbf{Code}} & \multicolumn{4}{c}{\textbf{Science}} \\
\textbf{Model}
& \begin{tabular}[c]{@{}c@{}}Code\\Forces\end{tabular}
& LCBv5
& \begin{tabular}[c]{@{}c@{}}RG\\Algo\end{tabular}
& \textbf{Avg}
& GPQA
& \begin{tabular}[c]{@{}c@{}}Super\\GPQA\end{tabular}
& \begin{tabular}[c]{@{}c@{}}Sci\\Bench\end{tabular}
& \textbf{Avg} \\
\midrule
Qwen3-1.7B   & \textcolor{gray}{9.2} & \textcolor{gray}{8.3} & \textcolor{gray}{19.5} & \textcolor{gray}{12.3}
               & \textcolor{gray}{40.9} & \textcolor{gray}{23.8} & \textcolor{gray}{27.6} & \textcolor{gray}{30.8} \\
\quad + OPD   & \textbf{19.1} & \underline{29.9} & \underline{32.7} & \underline{27.2}
               & \underline{48.9} & \underline{26.2} & \underline{36.5} & \underline{37.2} \\
\rowcolor{ngreenlight}
\quad + \ours   & \underline{18.5} & \textbf{33.6} & \textbf{41.3} & \textbf{31.1}
               & \textbf{54.9} & \textbf{28.5} & \textbf{41.2} & \textbf{41.5} \\
\midrule
Qwen3-4B   & \textcolor{gray}{16.1} & \textcolor{gray}{26.6} & \textcolor{gray}{32.0} & \textcolor{gray}{24.9}
               & \textcolor{gray}{43.1} & \textcolor{gray}{31.5} & \textcolor{gray}{46.6} & \textcolor{gray}{40.4} \\
\quad + OPD   & \underline{30.5} & \underline{42.0} & \underline{48.2} & \underline{40.2}
               & \underline{56.3} & \underline{36.5} & \underline{50.5} & \underline{47.8} \\
\rowcolor{ngreenlight}
\quad + \ours   & \textbf{36.4} & \textbf{56.1} & \textbf{60.7} & \textbf{51.1}
               & \textbf{60.5} & \textbf{38.4} & \textbf{54.8} & \textbf{51.2} \\
\bottomrule
\end{tabular}
\label{tab:qwen_nothink_other}
}
\end{table}

%% file: figs/reward_full.tex
\begin{table}[t]
\centering
\caption{
\textbf{Effect of \ours across OPD variants (thinking mode).}
Results where \ours improves over the on-policy baseline are shown in bold. ``All'' denotes the average across all 13 benchmarks.
}
\label{tab:avg_reward}
\small
\setlength{\tabcolsep}{3.5pt}
\resizebox{\textwidth}{!}{
\begin{tabular}{l|ccc|ccc|ccc|ccc}
\toprule
& \multicolumn{3}{c|}{\textbf{Math}} & \multicolumn{3}{c|}{\textbf{Code}} & \multicolumn{3}{c|}{\textbf{Science}} & \multicolumn{3}{c}{\textbf{All}} \\
  & - & \cellcolor{ngreenlight}\scriptsize{\ours} & \scriptsize{$\Delta$} & - & \cellcolor{ngreenlight}\scriptsize{\ours} & \scriptsize{$\Delta$} & - & \cellcolor{ngreenlight}\scriptsize{\ours} & \scriptsize{$\Delta$} & - & \cellcolor{ngreenlight}\scriptsize{\ours} & \scriptsize{$\Delta$} \\
\midrule
\multicolumn{13}{c}{{Qwen3-1.7B}} \\
\midrule
\quad OPD & 60.0 & \cellcolor{ngreenlight}\textbf{61.8} & \textbf{+1.83} & 34.3 & \cellcolor{ngreenlight}\textbf{37.6} & \textbf{+3.24} & 40.9 & \cellcolor{ngreenlight}\textbf{41.9} & \textbf{+1.06} & 49.7 & \cellcolor{ngreenlight}\textbf{51.6} & \textbf{+1.98} \\
\quad ExOPD & 63.1 & \cellcolor{ngreenlight}\textbf{63.4} & \textbf{+0.27} & 38.2 & \cellcolor{ngreenlight}\textbf{39.0} & \textbf{+0.77} & 43.4 & \cellcolor{ngreenlight}43.2 & -0.12 & 52.8 & \cellcolor{ngreenlight}\textbf{53.1} & \textbf{+0.30} \\
\quad OPD$^2$ & 64.3 & \cellcolor{ngreenlight}\textbf{65.3} & \textbf{+1.00} & 38.5 & \cellcolor{ngreenlight}\textbf{43.0} & \textbf{+4.48} & 44.2 & \cellcolor{ngreenlight}\textbf{45.0} & \textbf{+0.83} & 53.7 & \cellcolor{ngreenlight}\textbf{55.5} & \textbf{+1.76} \\
\midrule
\multicolumn{13}{c}{{Qwen3-4B}} \\
\midrule
\quad OPD & 70.7 & \cellcolor{ngreenlight}\textbf{72.2} & \textbf{+1.51} & 52.9 & \cellcolor{ngreenlight}\textbf{53.4} & \textbf{+0.49} & 49.8 & \cellcolor{ngreenlight}\textbf{50.6} & \textbf{+0.80} & 61.7 & \cellcolor{ngreenlight}\textbf{62.9} & \textbf{+1.11} \\
\quad ExOPD & 72.8 & \cellcolor{ngreenlight}\textbf{76.0} & \textbf{+3.20} & 58.7 & \cellcolor{ngreenlight}57.6 & -1.10 & 51.5 & \cellcolor{ngreenlight}\textbf{53.2} & \textbf{+1.70} & 64.6 & \cellcolor{ngreenlight}\textbf{66.5} & \textbf{+1.86} \\
\quad OPD$^2$ & 74.0 & \cellcolor{ngreenlight}\textbf{76.8} & \textbf{+2.85} & 58.4 & \cellcolor{ngreenlight}\textbf{60.6} & \textbf{+2.17} & 52.3 & \cellcolor{ngreenlight}\textbf{52.6} & \textbf{+0.28} & 65.4 & \cellcolor{ngreenlight}\textbf{67.5} & \textbf{+2.10} \\
\bottomrule
\end{tabular}
}
\end{table}

%% file: figs/gemma_full.tex
\begin{table}[t]
\centering
\caption{
\textbf{Results of Gemma-4-E4B models on math, code, and science benchmarks (thinking mode).} The best and second-best results are shown in bold and underlined, respectively.}
\label{tab:gemma_think_math_results}
\small
\setlength{\tabcolsep}{5pt}
\resizebox{\textwidth}{!}{
\begin{tabular}{lcccccccc}
\toprule
& \multicolumn{8}{c}{\textbf{Math}} \\
\textbf{Model} & AIME24 & AIME25 & AMC23 & HMMT25 & MATH500 & Olympiad & RGMath & \textbf{Avg} \\
\midrule
Gemma-4-E4B   & \textcolor{gray}{55.0} & \textcolor{gray}{35.0} & \textcolor{gray}{\underline{88.8}} & \textcolor{gray}{28.7} & \textcolor{gray}{\underline{88.0}} & \textcolor{gray}{\underline{42.6}} & \textcolor{gray}{89.7} & \textcolor{gray}{61.1} \\
\quad + OPD   & \underline{57.9} & \underline{42.1} & 81.2 & \underline{35.8} & 87.4 & 42.1 & \textbf{94.1} & \underline{63.0} \\
\rowcolor{ngreenlight}
\quad + \ours   & \textbf{61.9} & \textbf{45.6} & \textbf{91.6} & \textbf{37.5} & \textbf{89.8} & \textbf{44.7} & \underline{93.9} & \textbf{66.4} \\
\bottomrule
\end{tabular}
\label{tab:gemma}
}

\setlength{\tabcolsep}{8.5pt}
\resizebox{\textwidth}{!}{%
\begin{tabular}{lcccc|cccc}
\toprule
& \multicolumn{4}{c|}{\textbf{Code}} & \multicolumn{4}{c}{\textbf{Science}} \\
\textbf{Model}
& \begin{tabular}[c]{@{}c@{}}Code\\Forces\end{tabular}
& LCBv5
& \begin{tabular}[c]{@{}c@{}}RG\\Algo\end{tabular}
& \textbf{AVG}
& GPQA
& \begin{tabular}[c]{@{}c@{}}Super\\GPQA\end{tabular}
& \begin{tabular}[c]{@{}c@{}}Sci\\Bench\end{tabular}
& \textbf{AVG} \\
\midrule
Gemma-4-E4B   & \textcolor{gray}{\textbf{44.1}} & \textcolor{gray}{\textbf{61.2}} & \textcolor{gray}{\underline{64.1}} & \textcolor{gray}{\textbf{56.5}}
               & \textcolor{gray}{59.5} & \textcolor{gray}{36.8} & \textcolor{gray}{47.8} & \textcolor{gray}{48.0} \\
\quad + OPD   & \underline{41.9} & \underline{59.1} & 63.7 & 54.9
               & \underline{60.4} & \underline{37.6} & \underline{49.4} & \underline{49.1} \\
\rowcolor{ngreenlight}
\quad + \ours   & 40.8 & 58.4 & \textbf{66.4} & \underline{55.2}
               & \textbf{62.0} & \textbf{37.9} & \textbf{49.6} & \textbf{49.8} \\
\bottomrule
\end{tabular}
}
\end{table}

%% file: figs/gemma.tex

%% file: figs/anal_ld.tex
\definecolor{ink}{HTML}{22262B}   
\definecolor{mut}{HTML}{6E767E}   
\definecolor{trd}{HTML}{B4BBC2}   
\definecolor{lad}{HTML}{C2562A}   
\definecolor{rew}{HTML}{1E5F80}   

\pgfplotsset{
  pairax/.style={
    axis line style   = {mut!50, line width=0.8pt},
    major tick style  = {mut!50, line width=0.8pt},
    tick align        = outside,
    tick pos          = left,
    grid              = major,
    grid style        = {mut!20, line width=0.6pt},
    label style       = {font=\scriptsize, color=ink},
    tick label style  = {font=\footnotesize, color=mut},
    xlabel near ticks, ylabel near ticks,
    scaled x ticks    = false,
    legend style      = {font=\footnotesize, at={(0.5,1.02)}, anchor=south,
                         draw=none, fill=none, legend columns=-1,
                         /tikz/every even column/.append style={column sep=14pt}},
    legend cell align = left,
    clip              = false,
  },
  ladder/.style={
    only marks, mark=*, mark size=3.1pt,
    mark options={solid, fill=lad, draw=lad},
    error bars/.cd, x dir=both, x explicit,
    error bar style={lad!75, line width=1.1pt},
    error mark options={rotate=90, mark size=2.2pt, line width=1.1pt, lad!75},
  },
  selfmark/.style={
    only marks, mark=*, mark size=3.0pt,
    mark options={solid, fill=rew, draw=rew},
    error bars/.cd, x dir=both, x explicit,
    error bar style={rew!70, line width=1.1pt},
    error mark options={rotate=90, mark size=2.2pt, line width=1.1pt, rew!70},
  },
  inlegend/.style={
    legend columns=1,
    legend style={at={(0.975,0.035)}, anchor=south east,   
                  font=\scriptsize,
                  draw=gray!30, fill=gray!4, rounded corners=2pt,
                  column sep=2pt,
                  /tikz/every even column/.append style={column sep=5pt}},
    legend image code/.code={\draw[##1] (0cm,-0.04cm) rectangle (0.16cm,0.09cm);},
  },
}
\tikzset{
  lb/.style  = {font=\scriptsize, color=ink, inner sep=1.6pt},           
  lbo/.style = {font=\scriptsize\bfseries, color=lad!80!black,           
                inner sep=1.6pt},
  an/.style  = {font=\scriptsize, color=mut, inner sep=1.6pt},           
  spot/.style= {circle, fill=lad!16, minimum size=17pt, inner sep=0pt},  
}
\newcommand{\ladpt}[3]{%
  \addplot[ladder, /pgfplots/.cd, mark=*, forget plot,
           mark options={solid, fill=#1, draw=#1},
           /tikz/mark size=2.6pt, error bars/x dir=none]
    coordinates {(#2,#3)};}

\begin{figure}[t]
\centering
\begin{minipage}[t]{0.6\textwidth}
\centering
\begin{tikzpicture}
\begin{axis}[pairax,
  width=\linewidth, height=5cm,
  xmin=0.376, xmax=0.580, ymin=42.2, ymax=52.4,
  xtick={0.40,0.45,0.50,0.55},
  xticklabel style={/pgf/number format/.cd, fixed, fixed zerofill, precision=2},
  ytick={43,45,47,49,51},
  xlabel={\Name-policy suppression precision (NSP, $\uparrow$)},
  ylabel={Average accuracy},
  label style={font=\footnotesize},
  ticklabel style={font=\scriptsize},
  inlegend
]
  \addplot[trd, line width=1.1pt, forget plot, domain=0.380:0.575, samples=2]
          {29.268 + 39.20*x};
  \node[spot] at (axis cs:0.5587,51.65) {};
  \addplot[selfmark, /pgfplots/.cd, mark=*, mark options={solid, fill=rew, draw=rew},
           draw=rew, fill=rew,   
           /tikz/mark size=2.6pt, error bars/x dir=none] coordinates
           {(0.5393,49.67)};   
  \addlegendentry{On-policy}
  \ladpt{mut}{0.4005}{44.28}{0.0005}   
  \ladpt{mut!85}{0.4187}{46.36}{0.0022}
  \ladpt{mut!70}{0.4524}{47.26}{0.0008}      
  \ladpt{lad}{0.5587}{51.65}  
  \addlegendimage{draw=lad, fill=lad}
  \addlegendentry{\ours}
  \addlegendimage{draw=mut, fill=mut}
  \addlegendentry{Other rollouts}
  \node[lb,  anchor=north west, xshift=3pt, yshift=-2pt] at (axis cs:0.4005,44.28) {8B (teacher)};
  \node[lb,  anchor=north west, xshift=3pt, yshift=-2pt] at (axis cs:0.4187,46.36) {4B};
  \node[lb,  anchor=south east, xshift=-3pt, yshift=2pt]  at (axis cs:0.4524,47.26) {1.7B (student)};
  \node[lbo, anchor=east,       xshift=-12pt]            at (axis cs:0.5587,51.65) {\ours};
\end{axis}
\end{tikzpicture}
\caption{
\textbf{\Name-policy suppression precision (NSP) for OPD and different rollout policies.}
\ours with a lower-capability rollout policy than the student yields higher $\mathrm{NSP}$ and higher reasoning performance, while using higher-capability rollouts decreases accuracy.
}
\label{fig:nsp}
\end{minipage}
\hfill
\begin{minipage}[t]{0.375\textwidth}
\centering
\begin{tikzpicture}
\begin{axis}[pairax,
  width=\linewidth, height=5cm,
  xmin=0.42, xmax=4.5, ymin=-6.6, ymax=3.9,
  xtick={1,2,3,4}, xticklabels={0.6B,1.7B,4B,8B},
  ytick={-6,-4,-2,0,2},
  xlabel={Rollout model (Qwen3)},
  ylabel={$\Delta$ Average accuracy},
  label style={font=\footnotesize},
  ticklabel style={font=\scriptsize},
]
  \addplot[mut!60, dash pattern=on 3pt off 2pt, line width=0.8pt, mark=none,
           forget plot] coordinates {(0.42,0) (4.5,0)};
  \node[an, anchor=south east] at (axis cs:4.45,0.22) {OPD};
  \node[spot] at (axis cs:1,1.98) {};
  \addplot[only marks, mark=*, mark size=2.6pt, forget plot,
           mark options={solid, fill=mut, draw=mut}]  coordinates {(4,-5.39)};   
  \addplot[only marks, mark=*, mark size=2.6pt, forget plot,
           mark options={solid, fill=mut!85, draw=mut!85}] coordinates {(3,-3.31)};
  \addplot[only marks, mark=*, mark size=2.6pt, forget plot,
           mark options={solid, fill=mut!70, draw=mut!70}] coordinates {(2,-2.41)};
  \addplot[only marks, mark=*, mark size=2.6pt, forget plot,
           mark options={solid, fill=lad, draw=lad}]         coordinates {(1,1.98)};
  \node[lbo, anchor=south west, xshift=5pt, yshift=5pt] at (axis cs:1,1.98) {\Name policy};
\end{axis}
\end{tikzpicture}
\caption{
\textbf{Role of lower-capability rollouts.} When changing the rollout policy, only Qwen3-0.6B, a negative policy under our definition, improves over OPD.
}
\label{fig:sweep}
\end{minipage}
\end{figure}

%% file: figs/bow_2.tex
\definecolor{ink}{HTML}{1A1F24}  \definecolor{mut}{HTML}{8A929B}
\definecolor{mutt}{HTML}{E7E8EA}
\definecolor{grn}{HTML}{3F9D5A}  \definecolor{org}{HTML}{E05A1F}
\definecolor{blu}{HTML}{3F9DD8}  \definecolor{mag}{HTML}{B03AA0}
 
\newcommand{\panelw}{3.5cm}    
\newcommand{\panelh}{2.5cm}    
\newcommand{\ylabsep}{-5pt}
 
\pgfplotsset{
  ptmark/.style={only marks, color=mut, mark=*, mark size=1.1pt,
                 mark options={solid, draw=mut, fill=mut, opacity=0.45}},
  idl/.style={mut, dashed, line width=1.0pt},
  zline/.style={mut, opacity=0.55, dotted, line width=0.8pt},
  chalo/.style={only marks, color=white, mark=*, mark size=3.4pt,
                mark options={draw=white, fill=white, line width=0.6pt}},
  cmark/.style={only marks, color=mag, mark=+, mark size=6.2pt,
                mark options={mag, line width=1.9pt}},
  cdot/.style={only marks, color=mag, mark=*, mark size=2.4pt,
               mark options={draw=white, line width=0.9pt, fill=mag}},
}
 
\begin{figure}[t]
\centering
\begin{minipage}[t]{0.65\textwidth}
\centering
\begin{subfigure}[b]{0.525\linewidth}
\centering
\begin{tikzpicture}
\begin{axis}[ldacc,
  scale only axis=true, width=\panelw, height=\panelh,
  xmin=-0.105, xmax=0.108, ymin=-0.05, ymax=0.078,
  xtick={-0.08,0,0.08},
  ytick={-0.04,0,0.04},
  scaled ticks=false,
  xticklabel style={/pgf/number format/fixed, /pgf/number format/precision=2},
  yticklabel style={/pgf/number format/fixed, /pgf/number format/precision=2},
  enlargelimits=false,
  xlabel={NOR ($\uparrow$)},
  ylabel={NOR under OPD ($\uparrow$)},
  ylabel style={yshift=\ylabsep},
  clip mode=individual,
  label style={font=\footnotesize},
  ticklabel style={font=\scriptsize}
]
\addplot[zline, forget plot] coordinates {(-0.105,0) (0.108,0)};
\addplot[zline, forget plot] coordinates {(0,-0.05) (0,0.078)};
\addplot[idl, forget plot, domain=-0.05:0.078, samples=2] {x}
        node[pos=0.8, sloped, above, mut, font=\scriptsize, inner sep=2pt] {$y=x$};
\addplot[ptmark, forget plot] coordinates {
  (0.01465,0.01465) (0.02167,0.00292) (-0.01166,-0.00741) (0.00435,-0.00476) (0.03213,0.02301)
  (0.01820,0.00190) (-0.00052,-0.00874) (0.01174,-0.00555) (0.03057,0.02750) (0.00219,-0.00160)
  (0.03455,0.00427) (-0.01053,-0.01921) (0.00941,-0.00149) (0.05727,0.02291) (0.02446,0.00259)
  (0.02220,-0.00352) (0.03833,0.00356) (0.00832,-0.00557) (0.01779,0.00592) (0.02365,0.00895)
  (-0.00778,0.00157) (0.02255,-0.01289) (0.02223,0.01349) (-0.00536,-0.01507) (0.03462,0.01043)
  (0.00451,0.01347) (0.01750,-0.00712) (0.01370,-0.00962) (0.01750,0.00796) (0.06379,0.05794)
  (0.00967,-0.00253) (0.01384,0.00009) (0.01348,0.00421) (0.01857,-0.00689) (0.01820,-0.00594)
  (-0.00335,-0.00867) (0.02212,0.00981) (0.01445,0.00700) (0.02998,0.02262) (0.00978,0.00077)
  (0.02630,0.00552) (0.03739,0.02247) (0.01645,-0.00670) (-0.00240,-0.00639) (0.01853,0.00345)
  (0.04530,0.02454) (0.07385,0.03766) (0.01677,0.01201) (0.00851,0.00015) (0.01332,0.00567)
  (0.01011,-0.00334) (0.00348,0.00912) (0.02610,0.01182) (0.04513,0.02263) (0.01947,0.00395)
  (-0.02888,-0.04388) (0.01426,0.00515) (0.05614,0.03764) (0.00502,0.00558) (0.02387,-0.00080)
  (-0.00869,-0.02869) (0.01949,-0.00754) (0.02640,0.02208) (0.00668,-0.01851) (0.00853,-0.01123)
  (0.00411,-0.02175) (0.02081,0.00012) (0.03696,0.02035) (0.01911,0.00309) (0.03083,0.01272)
  (0.02325,-0.00271) (0.02005,-0.01676) (-0.01659,-0.00106) (-0.00326,-0.00460) (0.03774,0.00644)
  (0.02588,0.01480) (0.02487,0.01460) (0.02852,0.00247) (0.02641,-0.01073) (0.01943,-0.01343)
  (-0.01937,-0.01555) (0.04252,0.02745) (0.02870,0.01504) (-0.00583,-0.01257) (0.03188,0.01547)
  (0.03326,0.04230) (0.01776,0.01990) (0.01785,-0.00297) (-0.01431,-0.03138) (0.01125,-0.01494)
  (0.02735,-0.00386) (0.00943,-0.00836) (-0.02137,-0.03379) (0.01718,0.00266) (0.02009,-0.00759)
  (-0.00351,-0.02069) (0.03625,-0.00055) (0.01963,0.00459) (-0.00746,-0.01988) (-0.01472,-0.02016)
  (0.00145,-0.01468) (0.00307,-0.00990) (0.02859,0.00906) (-0.00519,-0.01763) (0.01599,-0.01116)
  (-0.01076,-0.01107) (-0.01528,-0.02937) (0.01350,-0.00817) (-0.00038,-0.01429) (0.01287,-0.01123)
  (0.01319,0.00482) (0.01863,-0.00345) (0.01258,0.01660) (0.00490,0.00342) (-0.00899,-0.01287)
  (0.03719,0.02839) (0.05705,0.05753) (0.00103,0.00539) (0.00385,-0.00772) (0.02540,0.00793)
  (0.00162,-0.01184) (0.01170,0.00222) (0.01378,-0.01151) (-0.02224,-0.02024) (0.00417,-0.00680)
  (0.01033,0.01032) (-0.01987,-0.02841) (0.02002,0.01506) (0.06236,0.02303) (0.01176,-0.00147)
  (0.00771,0.00188) (0.04193,0.02049) (0.01894,0.00373) (0.01184,-0.00125) (0.00812,0.00414)
  (0.01740,-0.00751) (0.03267,0.01411) (0.03076,0.00538) (0.02371,0.00975) (0.02573,0.00662)
  (0.00546,0.00816) (0.02511,0.00782) (0.00396,-0.01457) (-0.02808,-0.03580) (0.01327,0.00940)
  (0.02066,0.00358) (-0.00529,-0.02329) (0.03512,0.00221) (0.03196,-0.00259) (0.00838,-0.00044)
  (0.01627,-0.00191) (0.02085,-0.00519) (0.01538,-0.00354) (0.00772,-0.00670) (0.01345,-0.00734)
  (0.00640,-0.00477) (0.01642,0.00140) (0.03106,0.00419) (0.00333,-0.01387) (0.00299,0.00771)
  (0.01600,0.00263) (0.01554,0.00426) (0.00511,-0.00435) (0.00617,0.01972) (0.01085,-0.00823)
  (-0.01117,-0.01730) (0.00689,-0.01253) (0.00936,-0.00795) (0.01743,0.00411) (0.00257,-0.02489)
  (0.00804,0.00729) (-0.00120,-0.00114) (-0.00947,-0.01078) (0.00338,-0.00442) (0.01627,-0.00504)
  (0.01165,-0.01257) (0.01782,-0.00288) (0.05415,0.02032) (0.02797,0.00263) (0.00616,-0.00347)
  (0.01252,0.01848) (0.03200,0.00313) (0.00067,-0.00624) (0.01190,-0.00047) (0.01986,-0.00653)
  (0.02863,0.01767) (0.00498,0.00953) (-0.00386,-0.02191) (0.04733,0.01334) (0.00075,-0.00573)
  (0.02839,0.00255) (0.00962,-0.00239) (0.00618,-0.00275) (0.02289,0.00978) (0.02546,0.01273)
  (0.03785,0.00772) (0.02137,-0.01121) (0.02245,0.00215) (0.00676,0.01002) (0.02613,0.00395)
  (-0.01835,-0.02323) (0.02783,0.01093) (0.06490,0.05821) (0.02062,0.00444) (0.03070,0.01260)
  (0.00288,-0.00074) (0.02706,0.00702) (0.01634,0.00157) (0.07507,0.07163) (0.02841,0.01274)
  (0.02475,0.01285) (0.03559,0.01712) (-0.00094,-0.01098) (0.03745,0.01572) (0.03097,0.00841)
  (0.02524,0.01048) (0.00649,-0.00340) (0.05202,0.01531) (0.04371,0.01152) (0.06865,0.03995)
  (0.03615,0.01768) (0.03978,0.03227) (0.02179,0.02227) (0.03168,0.00801) (0.03776,0.01354)
  (0.04343,0.02351) (0.01431,-0.00468) (0.00704,-0.01042) (0.02239,0.00998) (0.03625,0.03243)
  (0.02352,0.00039) (0.04490,0.02350) (0.00265,0.00804) (0.01885,0.01518) (0.02175,0.00982)
  (0.04928,0.03963) (0.01328,0.01379) (0.00918,0.01526) (0.02100,-0.00043) (-0.00079,-0.01266)
  (0.01290,0.00869) (-0.00392,-0.00268) (0.01787,0.00377) (0.05711,0.04560) (0.05844,0.03990)
  (0.10166,0.06194) (0.06990,0.05676) (0.05158,0.04086) (0.02759,0.01475) (0.00151,-0.00653)
  (0.04089,0.02998) (0.01701,0.01963) (-0.00443,-0.00875) (0.02013,0.00627) (0.04420,0.02824)
  (0.02669,0.02028) (0.03392,0.01576) (0.03956,0.03081) (0.01196,0.01520) (0.00296,-0.00439)
  (0.01868,0.02484) (0.00404,-0.00326) (0.01548,0.00630) (0.03692,0.01757) (-0.00516,-0.03179)
  (0.04356,0.03369) (0.03863,0.01382) (0.03761,0.02446) (0.04238,0.02535) (0.03134,0.01893)
  (0.02546,0.02504) (0.01834,0.01887) (0.03017,0.00901) (0.05262,0.01876) (0.03442,0.01471)
  (0.01649,0.02210) (0.03004,0.01077) (0.01479,0.01606) (0.04755,0.02177) (0.03197,0.00234)
  (0.01503,0.01837) (-0.01150,-0.02700) (0.06625,0.03710) (0.01626,0.02782) (0.03256,0.02102)
  (0.04107,0.01284) (0.03981,0.00825) (0.01040,-0.00763) (0.03304,0.03092) (0.00770,-0.00143)
  (0.02466,0.01070) (0.01040,0.02275) (0.02963,0.01833) (0.01494,0.02247) (0.01247,-0.00519)
  (0.02883,0.01625) (0.00494,-0.00651) (-0.00063,0.00235) (0.03526,0.02742) (-0.00831,-0.01629)
};
\addplot[zline, mag, opacity=0.85, line width=1.3pt, forget plot]
        coordinates {(-0.105,0.00521) (0.01905,0.00521)};
\addplot[zline, mag, opacity=0.85, line width=1.3pt, forget plot]
        coordinates {(0.01905,-0.05) (0.01905,0.00521)};
\addplot[chalo, forget plot] coordinates {(0.01905,0.00521)};
\addplot[cmark, forget plot] coordinates {(0.01905,0.00521)};
\addplot[cdot, forget plot]  coordinates {(0.01905,0.00521)};
\node[font=\scriptsize, anchor=north west, inner sep=2pt] at (axis cs:-0.1015,0.0755)
     {\textcolor{mag}{$(0.019,\,0.005)$}};
\end{axis}
\end{tikzpicture}
\vspace{-1.7em}
\subcaption{\Name rollouts}\label{fig:away-ours}
\end{subfigure}
\begin{subfigure}[b]{0.405\linewidth}
\centering
\begin{tikzpicture}
\begin{axis}[ldacc,
  scale only axis=true, width=\panelw, height=\panelh,
  xmin=-0.105, xmax=0.108, ymin=-0.05, ymax=0.078,
  xtick={-0.08,0,0.08},
  ytick={-0.04,0,0.04},
  scaled ticks=false,
  xticklabel style={/pgf/number format/fixed, /pgf/number format/precision=2},
  yticklabel style={/pgf/number format/fixed, /pgf/number format/precision=2},
  enlargelimits=false,
  xlabel={NOR ($\uparrow$)},
  yticklabels={},
  clip mode=individual,
  label style={font=\footnotesize},
  ticklabel style={font=\scriptsize},  
]
\addplot[zline, forget plot] coordinates {(-0.105,0) (0.108,0)};
\addplot[zline, forget plot] coordinates {(0,-0.05) (0,0.078)};
\addplot[idl, forget plot, domain=-0.05:0.078, samples=2] {x}
        node[pos=0.8, sloped, below, mut, font=\scriptsize, inner sep=2pt] {$y=x$};
\addplot[ptmark, forget plot] coordinates {
  (-0.00757,0.01465) (-0.03713,0.00292) (-0.06304,-0.00741) (-0.06716,-0.00476) (-0.03096,0.02301)
  (-0.06399,0.00190) (-0.07953,-0.00874) (-0.04232,-0.00555) (-0.00952,0.02750) (-0.05304,-0.00160)
  (-0.05058,0.00427) (-0.06106,-0.01921) (-0.01261,-0.00149) (0.00626,0.02291) (-0.03307,0.00259)
  (-0.01248,-0.00352) (-0.00148,0.00356) (-0.08504,-0.00557) (-0.00885,0.00592) (-0.00070,0.00895)
  (0.00834,0.00157) (-0.05324,-0.01289) (-0.03550,0.01349) (-0.07439,-0.01507) (-0.01417,0.01043)
  (-0.01089,0.01347) (-0.03694,-0.00712) (-0.04716,-0.00962) (-0.03849,0.00796) (0.02341,0.05794)
  (-0.05356,-0.00253) (-0.06776,0.00009) (-0.04822,0.00421) (-0.06224,-0.00689) (-0.05321,-0.00594)
  (-0.04739,-0.00867) (0.00516,0.00981) (-0.02329,0.00700) (-0.01071,0.02262) (-0.04708,0.00077)
  (-0.02570,0.00552) (-0.03300,0.02247) (-0.03218,-0.00670) (-0.06251,-0.00639) (-0.02789,0.00345)
  (-0.00266,0.02454) (-0.00738,0.03766) (-0.02662,0.01201) (-0.07599,0.00015) (-0.03628,0.00567)
  (-0.04600,-0.00334) (-0.08986,0.00912) (-0.02679,0.01182) (0.00952,0.02263) (-0.05523,0.00395)
  (-0.07317,-0.04388) (-0.05711,0.00515) (-0.00260,0.03764) (-0.03253,0.00558) (0.00784,-0.00080)
  (-0.03421,-0.02869) (-0.06212,-0.00754) (-0.01254,0.02208) (-0.04375,-0.01851) (-0.04447,-0.01123)
  (-0.03922,-0.02175) (-0.03000,0.00012) (-0.00788,0.02035) (-0.03733,0.00309) (-0.01468,0.01272)
  (-0.02711,-0.00271) (-0.04839,-0.01676) (-0.09156,-0.00106) (-0.04363,-0.00460) (-0.02633,0.00644)
  (-0.03070,0.01480) (-0.03620,0.01460) (-0.07000,0.00247) (-0.07638,-0.01073) (-0.07515,-0.01343)
  (-0.05567,-0.01555) (-0.00742,0.02745) (-0.02940,0.01504) (-0.05170,-0.01257) (-0.02494,0.01547)
  (-0.00169,0.04230) (-0.05431,0.01990) (-0.04197,-0.00297) (-0.06162,-0.03138) (-0.01929,-0.01494)
  (-0.02702,-0.00386) (-0.05573,-0.00836) (-0.07351,-0.03379) (-0.07016,0.00266) (-0.05978,-0.00759)
  (-0.03054,-0.02069) (-0.02496,-0.00055) (-0.02904,0.00459) (-0.08354,-0.01988) (-0.06617,-0.02016)
  (-0.04652,-0.01468) (-0.01889,-0.00990) (-0.00174,0.00906) (-0.05949,-0.01763) (-0.03679,-0.01116)
  (-0.08683,-0.01107) (-0.04881,-0.02937) (-0.03522,-0.00817) (-0.06182,-0.01429) (-0.08661,-0.01123)
  (-0.04600,0.00482) (-0.05674,-0.00345) (-0.03506,0.01660) (-0.08694,0.00342) (-0.05064,-0.01287)
  (-0.02595,0.02839) (0.03600,0.05753) (-0.06725,0.00539) (-0.05389,-0.00772) (0.02073,0.00793)
  (-0.05615,-0.01184) (-0.03395,0.00222) (-0.03153,-0.01151) (-0.04271,-0.02024) (-0.05713,-0.00680)
  (-0.00314,0.01032) (-0.09808,-0.02841) (-0.03394,0.01506) (-0.00326,0.02303) (-0.05011,-0.00147)
  (-0.06866,0.00188) (-0.01043,0.02049) (-0.05644,0.00373) (-0.05923,-0.00125) (-0.04173,0.00414)
  (-0.06134,-0.00751) (0.00618,0.01411) (-0.02544,0.00538) (-0.08317,0.00975) (-0.04170,0.00662)
  (-0.08385,0.00816) (-0.06816,0.00782) (-0.04668,-0.01457) (-0.07280,-0.03580) (-0.05079,0.00940)
  (-0.04292,0.00358) (-0.07075,-0.02329) (-0.04271,0.00221) (-0.05278,-0.00259) (-0.02696,-0.00044)
  (-0.02854,-0.00191) (-0.01853,-0.00519) (-0.01441,-0.00354) (-0.04574,-0.00670) (-0.06615,-0.00734)
  (-0.01439,-0.00477) (-0.04211,0.00140) (-0.06665,0.00419) (-0.01045,-0.01387) (-0.02866,0.00771)
  (-0.06374,0.00263) (-0.05349,0.00426) (-0.03951,-0.00435) (-0.05541,0.01972) (-0.08002,-0.00823)
  (-0.08112,-0.01730) (-0.01214,-0.01253) (-0.01737,-0.00795) (-0.05614,0.00411) (-0.05905,-0.02489)
  (-0.03836,0.00729) (-0.03063,-0.00114) (-0.04778,-0.01078) (-0.03709,-0.00442) (-0.03484,-0.00504)
  (-0.03912,-0.01257) (0.01255,-0.00288) (-0.02082,0.02032) (-0.02713,0.00263) (-0.04523,-0.00347)
  (-0.06680,0.01848) (-0.03852,0.00313) (-0.05963,-0.00624) (-0.05188,-0.00047) (-0.02517,-0.00653)
  (0.01283,0.01767) (-0.05056,0.00953) (-0.07902,-0.02191) (0.01051,0.01334) (-0.07954,-0.00573)
  (-0.03877,0.00255) (-0.06039,-0.00239) (-0.07357,-0.00275) (-0.04491,0.00978) (-0.01594,0.01273)
  (-0.00417,0.00772) (-0.07133,-0.01121) (-0.04558,0.00215) (-0.05023,0.01002) (-0.00292,0.00395)
  (-0.06013,-0.02323) (0.02032,0.01093) (0.02521,0.05821) (-0.02537,0.00444) (-0.03008,0.01260)
  (-0.03713,-0.00074) (-0.00541,0.00702) (0.00436,0.00157) (0.04629,0.07163) (-0.02603,0.01274)
  (0.02297,0.01285) (0.01176,0.01712) (-0.03311,-0.01098) (-0.01044,0.01572) (-0.01721,0.00841)
  (-0.02399,0.01048) (-0.01603,-0.00340) (-0.00731,0.01531) (-0.02389,0.01152) (0.00572,0.03995)
  (0.03061,0.01768) (0.01152,0.03227) (0.00417,0.02227) (-0.00805,0.00801) (-0.02094,0.01354)
  (-0.00294,0.02351) (-0.02650,-0.00468) (-0.01760,-0.01042) (-0.00348,0.00998) (0.01129,0.03243)
  (-0.01623,0.00039) (-0.01220,0.02350) (-0.01876,0.00804) (-0.01052,0.01518) (0.00185,0.00982)
  (0.00649,0.03963) (-0.02515,0.01379) (-0.00766,0.01526) (0.01144,-0.00043) (-0.01838,-0.01266)
  (-0.02446,0.00869) (-0.04705,-0.00268) (-0.00820,0.00377) (0.01868,0.04560) (0.03175,0.03990)
  (0.00519,0.06194) (0.03158,0.05676) (0.03030,0.04086) (-0.00286,0.01475) (0.00212,-0.00653)
  (-0.01996,0.02998) (-0.05122,0.01963) (-0.01759,-0.00875) (-0.01612,0.00627) (0.00064,0.02824)
  (0.00090,0.02028) (-0.02640,0.01576) (0.01750,0.03081) (-0.01147,0.01520) (-0.01217,-0.00439)
  (0.00116,0.02484) (-0.02665,-0.00326) (0.00075,0.00630) (-0.01939,0.01757) (-0.02839,-0.03179)
  (-0.00013,0.03369) (-0.03023,0.01382) (-0.01096,0.02446) (0.02103,0.02535) (0.01614,0.01893)
  (0.00185,0.02504) (0.00081,0.01887) (-0.01011,0.00901) (-0.02543,0.01876) (-0.01125,0.01471)
  (-0.00934,0.02210) (-0.02014,0.01077) (-0.00826,0.01606) (0.02361,0.02177) (-0.02247,0.00234)
  (-0.03737,0.01837) (-0.07229,-0.02700) (0.01167,0.03710) (-0.02840,0.02782) (-0.02031,0.02102)
  (-0.01919,0.01284) (-0.02474,0.00825) (-0.05373,-0.00763) (-0.00076,0.03092) (-0.02490,-0.00143)
  (-0.02980,0.01070) (-0.03254,0.02275) (-0.02911,0.01833) (-0.01197,0.02247) (-0.01071,-0.00519)
  (0.00944,0.01625) (-0.03751,-0.00651) (-0.01754,0.00235) (-0.02942,0.02742) (-0.00404,-0.01629)
};
\addplot[zline, mag, opacity=0.85, line width=1.3pt, forget plot]
        coordinates {(-0.105,0.00521) (-0.03028,0.00521)};
\addplot[zline, mag, opacity=0.85, line width=1.3pt, forget plot]
        coordinates {(-0.03028,-0.05) (-0.03028,0.00521)};
\addplot[chalo, forget plot] coordinates {(-0.03028,0.00521)};
\addplot[cmark, forget plot] coordinates {(-0.03028,0.00521)};
\addplot[cdot, forget plot]  coordinates {(-0.03028,0.00521)};
\node[font=\scriptsize, anchor=north west, inner sep=2pt] at (axis cs:-0.1015,0.0755)
     {\textcolor{mag}{$(-0.030,\,0.005)$}};
\end{axis}
\end{tikzpicture}
\vspace{-1.7em}
\subcaption{Teacher rollouts}\label{fig:away-8b}
\end{subfigure}
\vspace{-0.3em}
\caption{
\textbf{Reduction in overlap with the \name policy.}
Each point compares trajectory-level NOR under OPD (y-axis) and (a) negative-policy or (b) teacher rollouts (x-axis). Points below the diagonal indicate greater overlap reduction.
}
\label{fig:away-a}
\end{minipage}\hfill
\begin{minipage}[t]{0.33\textwidth}
\centering
\vspace{-11.5em}
\begin{tikzpicture}
\begin{axis}[ldacc,
  scale only axis=true, width=\panelw, height=\panelh,
  xmin=0.530, xmax=0.610, ymin=47.5, ymax=56.2,   
  xtick={0.54,0.56,0.58,0.60},
  ytick={48,52,56},
  scaled ticks=false,
  xticklabel style={/pgf/number format/fixed zerofill, /pgf/number format/precision=2},
  xlabel={NSP ($\uparrow$)},
  ylabel={Average accuracy},
  ylabel style={yshift=\ylabsep},
  legend style={at={(0.985,0.03)}, anchor=south east, legend columns=1,
                font=\tiny, draw=gray!30, fill=gray!4, rounded corners=2pt,
                inner xsep=2.5pt, inner ysep=1.5pt, row sep=-4pt,
                /tikz/every even column/.append style={column sep=-4pt}},
  legend image code/.code={\draw[#1] (0cm,-0.035cm) rectangle (0.13cm,0.075cm);},
  label style={font=\footnotesize},
  ticklabel style={font=\scriptsize}
]

\addplot[trd, line width=1.0pt, forget plot, domain=0.530:0.60785, samples=2]   
        {-0.937 + 94.00*x};

\addplot[selfmark, /pgfplots/.cd, mark=*,
         mark options={solid, fill=rew, draw=rew},
         draw=rew, fill=rew,   
         /tikz/mark size=2.0pt, error bars/x dir=none] coordinates {
  (0.5393,49.67)   
  (0.5795,52.82)   
  (0.5742,53.71)   
};
\addlegendentry{On-policy}

\addplot[weakmark, /pgfplots/.cd, mark=*,   
         mark options={solid, fill=lad, draw=lad},
         draw=lad, fill=lad,   
         /tikz/mark size=2.0pt, error bars/x dir=none] coordinates {
  (0.5587,51.65)   
  (0.5824,53.12)   
  (0.5843,54.73)   
};
\addlegendentry{\ours}

\node[lbs, anchor=north west, xshift=1pt,  yshift=-2pt] at (axis cs:0.5393,49.67) {OPD};
\node[lbs, anchor=north west, xshift=1pt,  yshift=-4pt] at (axis cs:0.5795,52.82) {ExOPD};
\node[lbs, anchor=south east, xshift=-1pt, yshift=2pt]  at (axis cs:0.5742,53.71) {OPD$^2$};
\node[lbw, anchor=south east, xshift=-1pt, yshift=2pt]  at (axis cs:0.5587,51.65) {OPD};
\node[lbw, anchor=west,       xshift=2pt]               at (axis cs:0.5824,53.12) {ExOPD};
\node[lbw, anchor=west,       xshift=2pt]               at (axis cs:0.5843,54.73) {OPD$^2$};

\end{axis}
\end{tikzpicture}
\vspace{-.3em}
\caption{
\textbf{NSP across OPD variants.}
Each pair compares on-policy training with \ours for three formulations; OPD, ExOPD, and OPD$^2$.
}
\label{fig:nsp-reward}
\end{minipage}

\end{figure}

%% file: figs/summary_nothink.tex
\begin{table}[h]
\centering
\caption{\textbf{Summary of Qwen3 results in non-thinking mode.}
The table reports average performance across Math, Code, and Science benchmarks. The best and second-best results are shown in bold and underlined, respectively.
}
\label{tab:qwen_summary_nothink}
\small
\setlength{\tabcolsep}{5pt}

\begin{tabular}{lccc|ccc|ccc}
\toprule
& \multicolumn{3}{c|}{\textbf{Qwen3-1.7B}}
& \multicolumn{3}{c|}{\textbf{Qwen3-4B}}
& \multicolumn{3}{c}{\textbf{Qwen3-8B}} \\
\cmidrule(lr){2-4} \cmidrule(lr){5-7} \cmidrule(lr){8-10}

\textbf{Model}
& Math & Code & Sci.
& Math & Code & Sci.
& Math & Code & Sci. \\
\midrule
Baseline
& 35.7 & 12.3 & 30.8
& 46.8 & 24.9 & 40.4
& 45.7 & 29.8 & 43.0 \\
\quad OPD
& 48.3 & 27.2 & 37.2
& 62.2 & 40.2 & 47.8
& 65.1 & 44.0 & 50.1 \\
\rowcolor{ngreenlight}
\quad + \ours
& \textbf{56.3} & 31.1 & 41.5
& \underline{70.5} & 51.1 & \underline{51.2}
& 69.4 & 44.3 & 52.6 \\
\midrule
\quad ExOPD
& 53.1 & 30.4 & 38.2
& 65.8 & 44.5 & 48.6
& 68.5 & 45.6 & 51.6 \\
\rowcolor{ngreenlight}
\quad + \ours
& \underline{55.5} & 30.2 & \underline{41.5}
& \textbf{71.3} & \textbf{51.2} & \textbf{51.5}
& \underline{71.7} & \underline{47.2} & \textbf{54.2} \\
\midrule
\quad OPD$^2$
& 53.1 & \underline{31.7} & 39.5
& 66.3 & 46.8 & 49.4
& 68.3 & 42.7 & 52.6 \\
\rowcolor{ngreenlight}
\quad + \ours
& 54.7 & \textbf{31.9} & \textbf{43.1}
& 69.9 & \underline{51.2} & 51.1
& \textbf{72.9} & \textbf{53.0} & \underline{53.2} \\
\bottomrule
\end{tabular}%
\vspace{-4pt}
\end{table}

%% file: figs/nothink_rewards.tex
\begin{table}[h]
\centering
\caption{\textbf{Full results of Qwen3 models on mathematical reasoning benchmarks with OPD (non-thinking mode).} The best and second-best results among the OPD variants are shown in bold and underlined, respectively.}
\label{tab:nothink_math_results_v2_opd}
\small
\setlength{\tabcolsep}{5pt}
\resizebox{\textwidth}{!}{
\begin{tabular}{lcccccccc}
\toprule
\textbf{Model} & AIME24 & AIME25 & AMC23 & HMMT25 & MATH500 & Olympiad & RGMath & \textbf{Avg} \\
\midrule
Qwen3-1.7B   & \textcolor{gray}{11.5} & \textcolor{gray}{10.0} & \textcolor{gray}{45.9} & \textcolor{gray}{5.4} & \textcolor{gray}{69.0} & \textcolor{gray}{25.9} & \textcolor{gray}{82.5} & \textcolor{gray}{35.7} \\
\quad OPD   & 31.7 & 21.7 & 65.3 & 13.3 & 80.4 & 36.0 & 89.9 & 48.3 \\
\rowcolor{ngreenlight}
\quad + \ours ($\alpha$=0.25)   & 32.3 & 25.6 & 72.8 & 15.4 & 82.8 & 36.8 & 91.1 & 51.0 \\
\rowcolor{ngreenlight}
\quad + \ours ($\alpha$=0.5)   & 34.6 & 26.0 & 70.0 & 15.8 & 83.2 & 37.6 & 90.3 & 51.1 \\
\rowcolor{ngreenlight}
\quad + \ours ($\alpha$=0.75)   & \underline{38.5} & \underline{26.7} & \underline{74.1} & \underline{17.9} & \underline{83.4} & \underline{38.6} & \underline{92.0} & \underline{53.0} \\
\rowcolor{ngreenlight}
\quad + \ours ($\alpha$=1)   & \textbf{43.1} & \textbf{33.3} & \textbf{78.4} & \textbf{20.0} & \textbf{86.0} & \textbf{40.6} & \textbf{92.3} & \textbf{56.3} \\
\midrule
Qwen3-4B   & \textcolor{gray}{24.2} & \textcolor{gray}{20.0} & \textcolor{gray}{69.4} & \textcolor{gray}{11.7} & \textcolor{gray}{79.0} & \textcolor{gray}{34.7} & \textcolor{gray}{88.9} & \textcolor{gray}{46.8} \\
\quad OPD   & 53.8 & 42.3 & 90.0 & 26.2 & 88.2 & 43.7 & 90.9 & 62.2 \\
\rowcolor{ngreenlight}
\quad + \ours ($\alpha$=0.25)   & 59.2 & 45.8 & 90.6 & 28.3 & 89.8 & 44.2 & 91.5 & 64.2 \\
\rowcolor{ngreenlight}
\quad + \ours ($\alpha$=0.5)   & 59.8 & 45.8 & \underline{94.4} & 30.8 & 88.6 & 43.9 & \underline{92.6} & 65.1 \\
\rowcolor{ngreenlight}
\quad + \ours ($\alpha$=0.75)   & \underline{63.5} & \underline{48.8} & 94.1 & \underline{36.7} & \underline{90.0} & \underline{44.6} & \textbf{92.9} & \underline{67.2} \\
\rowcolor{ngreenlight}
\quad + \ours ($\alpha$=1)   & \textbf{67.9} & \textbf{58.8} & \textbf{95.3} & \textbf{43.8} & \textbf{90.6} & \textbf{45.5} & 91.7 & \textbf{70.5} \\
\midrule
Qwen3-8B   & \textcolor{gray}{26.5} & \textcolor{gray}{18.5} & \textcolor{gray}{66.9} & \textcolor{gray}{5.4} & \textcolor{gray}{78.8} & \textcolor{gray}{34.3} & \textcolor{gray}{89.6} & \textcolor{gray}{45.7} \\
\quad OPD   & 61.5 & 46.5 & 91.9 & 28.3 & 89.6 & 44.4 & 93.5 & 65.1 \\
\rowcolor{ngreenlight}
\quad + \ours ($\alpha$=0.25)   & 67.3 & 46.0 & 90.0 & 29.2 & \textbf{91.0} & 44.6 & 94.2 & 66.0 \\
\rowcolor{ngreenlight}
\quad + \ours ($\alpha$=0.5)   & 68.1 & 51.2 & 93.4 & 27.5 & 90.8 & 44.8 & \textbf{95.5} & 67.3 \\
\rowcolor{ngreenlight}
\quad + \ours ($\alpha$=0.75)   & \underline{71.7} & \underline{53.3} & \underline{93.4} & \underline{37.5} & 90.4 & \underline{45.2} & \underline{94.4} & \underline{69.4} \\
\rowcolor{ngreenlight}
\quad + \ours ($\alpha$=1)   & \textbf{73.8} & \textbf{61.7} & \textbf{94.4} & \textbf{37.9} & \underline{90.8} & \textbf{45.7} & 93.5 & \textbf{71.1} \\
\bottomrule
\end{tabular}}
\label{tab:qwen_nothink_math_v2_opd}
\end{table}

\begin{table}[h]
\centering
\caption{\textbf{Full results of Qwen3 models on code and science benchmarks with OPD (non-thinking mode).} The best and second-best results among the OPD variants are shown in bold and underlined, respectively. \textbf{Avg} is the per-domain average.}
\label{tab:nothink_code_science_results_v2_opd}
\small
\setlength{\tabcolsep}{7.8pt}
\begin{tabular}{lcccc|cccc}
\toprule
& \multicolumn{4}{c|}{\textbf{Code}} & \multicolumn{4}{c}{\textbf{Science}} \\
\cmidrule(lr){2-5} \cmidrule(lr){6-9}
\textbf{Model}
& \begin{tabular}[c]{@{}c@{}}Code\\Forces\end{tabular}
& LCBv5
& \begin{tabular}[c]{@{}c@{}}RG\\Algo\end{tabular}
& \textbf{Avg}
& GPQA
& \begin{tabular}[c]{@{}c@{}}Super\\GPQA\end{tabular}
& \begin{tabular}[c]{@{}c@{}}Sci\\Bench\end{tabular}
& \textbf{Avg} \\
\midrule
Qwen3-1.7B   & \textcolor{gray}{9.2} & \textcolor{gray}{8.3} & \textcolor{gray}{19.5} & \textcolor{gray}{12.3}
               & \textcolor{gray}{40.9} & \textcolor{gray}{23.8} & \textcolor{gray}{27.6} & \textcolor{gray}{30.8} \\
\quad OPD   & \underline{19.1} & 29.9 & 32.7 & 27.2
               & 48.9 & 26.2 & 36.5 & 37.2 \\
\rowcolor{ngreenlight}
\quad + \ours ($\alpha$=0.25)   & \textbf{19.4} & 30.5 & 34.1 & 28.0
               & 49.6 & 26.5 & 37.5 & 37.8 \\
\rowcolor{ngreenlight}
\quad + \ours ($\alpha$=0.5)   & 16.8 & 30.5 & 34.0 & 27.1
               & 51.3 & \underline{27.9} & 36.9 & 38.7 \\
\rowcolor{ngreenlight}
\quad + \ours ($\alpha$=0.75)   & 18.8 & \underline{31.9} & \underline{37.7} & \underline{29.4}
               & \underline{51.8} & 27.8 & \underline{39.4} & \underline{39.7} \\
\rowcolor{ngreenlight}
\quad + \ours ($\alpha$=1)   & 18.5 & \textbf{33.6} & \textbf{41.3} & \textbf{31.1}
               & \textbf{54.9} & \textbf{28.5} & \textbf{41.2} & \textbf{41.5} \\
\midrule
Qwen3-4B   & \textcolor{gray}{16.1} & \textcolor{gray}{26.6} & \textcolor{gray}{32.0} & \textcolor{gray}{24.9}
               & \textcolor{gray}{43.1} & \textcolor{gray}{31.5} & \textcolor{gray}{46.6} & \textcolor{gray}{40.4} \\
\quad OPD   & \underline{30.5} & 42.0 & 48.2 & 40.2
               & 56.3 & 36.5 & 50.5 & 47.8 \\
\rowcolor{ngreenlight}
\quad + \ours ($\alpha$=0.25)   & 28.3 & 45.3 & 49.2 & 40.9
               & 57.3 & 36.4 & 53.7 & 49.1 \\
\rowcolor{ngreenlight}
\quad + \ours ($\alpha$=0.5)   & 28.7 & 45.0 & 51.5 & 41.7
               & 57.6 & 36.3 & \underline{54.5} & 49.4 \\
\rowcolor{ngreenlight}
\quad + \ours ($\alpha$=0.75)   & 28.3 & \underline{49.9} & \underline{54.1} & \underline{44.1}
               & \underline{57.7} & \underline{37.4} & 53.6 & \underline{49.6} \\
\rowcolor{ngreenlight}
\quad + \ours ($\alpha$=1)   & \textbf{36.4} & \textbf{56.1} & \textbf{60.7} & \textbf{51.1}
               & \textbf{60.5} & \textbf{38.4} & \textbf{54.8} & \textbf{51.2} \\
\midrule
Qwen3-8B   & \textcolor{gray}{21.1} & \textcolor{gray}{28.5} & \textcolor{gray}{39.9} & \textcolor{gray}{29.8}
               & \textcolor{gray}{52.1} & \textcolor{gray}{31.4} & \textcolor{gray}{45.6} & \textcolor{gray}{43.0} \\
\quad OPD   & 31.3 & \textbf{49.9} & 50.8 & 44.0
               & 60.4 & 38.3 & 51.6 & 50.1 \\
\rowcolor{ngreenlight}
\quad + \ours ($\alpha$=0.25)   & 30.2 & 42.5 & 50.1 & 41.0
               & \underline{63.9} & \textbf{38.7} & 52.1 & \underline{51.6} \\
\rowcolor{ngreenlight}
\quad + \ours ($\alpha$=0.5)   & \underline{31.3} & 46.9 & 52.5 & 43.6
               & 63.5 & 38.2 & 52.3 & 51.3 \\
\rowcolor{ngreenlight}
\quad + \ours ($\alpha$=0.75)   & \textbf{31.6} & \underline{48.0} & \underline{53.4} & \underline{44.3}
               & \textbf{65.8} & \underline{38.6} & \underline{53.3} & \textbf{52.6} \\
\rowcolor{ngreenlight}
\quad + \ours ($\alpha$=1)   & 30.2 & 47.2 & \textbf{57.7} & \textbf{45.0}
               & 56.8 & 36.6 & \textbf{53.3} & 48.9 \\
\bottomrule
\end{tabular}
\label{tab:qwen_nothink_other_v2_opd}
\end{table}

\begin{table}[h]
\centering
\caption{\textbf{Full results of Qwen3 models on mathematical reasoning benchmarks with ExOPD (non-thinking mode).} The best and second-best results among the OPD variants are shown in bold and underlined, respectively.}
\label{tab:nothink_math_results_v2_exopd}
\small
\setlength{\tabcolsep}{5pt}
\resizebox{\textwidth}{!}{
\begin{tabular}{lcccccccc}
\toprule
\textbf{Model} & AIME24 & AIME25 & AMC23 & HMMT25 & MATH500 & Olympiad & RGMath & \textbf{Avg} \\
\midrule
Qwen3-1.7B   & \textcolor{gray}{11.5} & \textcolor{gray}{10.0} & \textcolor{gray}{45.9} & \textcolor{gray}{5.4} & \textcolor{gray}{69.0} & \textcolor{gray}{25.9} & \textcolor{gray}{82.5} & \textcolor{gray}{35.7} \\
\quad ExOPD   & \underline{41.2} & 26.7 & \underline{75.0} & \underline{18.3} & 82.0 & 37.7 & \underline{90.8} & \underline{53.1} \\
\rowcolor{ngreenlight}
\quad + \ours ($\alpha$=0.5)   & 36.9 & \underline{27.5} & 73.8 & 17.5 & \underline{82.8} & \underline{37.8} & 90.1 & 52.3 \\
\rowcolor{ngreenlight}
\quad + \ours ($\alpha$=1)   & \textbf{41.2} & \textbf{31.5} & \textbf{80.9} & \textbf{19.6} & \textbf{83.2} & \textbf{39.6} & \textbf{92.2} & \textbf{55.5} \\
\midrule
Qwen3-4B   & \textcolor{gray}{24.2} & \textcolor{gray}{20.0} & \textcolor{gray}{69.4} & \textcolor{gray}{11.7} & \textcolor{gray}{79.0} & \textcolor{gray}{34.7} & \textcolor{gray}{88.9} & \textcolor{gray}{46.8} \\
\quad ExOPD   & \underline{59.6} & \underline{51.0} & \underline{93.8} & 30.4 & 89.6 & \underline{44.9} & 91.4 & 65.8 \\
\rowcolor{ngreenlight}
\quad + \ours ($\alpha$=0.5)   & 59.2 & 49.2 & 93.1 & \underline{34.2} & \underline{89.8} & 44.1 & \textbf{92.6} & \underline{66.0} \\
\rowcolor{ngreenlight}
\quad + \ours ($\alpha$=1)   & \textbf{69.6} & \textbf{62.7} & \textbf{96.2} & \textbf{40.8} & \textbf{91.4} & \textbf{45.8} & \underline{92.5} & \textbf{71.3} \\
\midrule
Qwen3-8B   & \textcolor{gray}{26.5} & \textcolor{gray}{18.5} & \textcolor{gray}{66.9} & \textcolor{gray}{5.4} & \textcolor{gray}{78.8} & \textcolor{gray}{34.3} & \textcolor{gray}{89.6} & \textcolor{gray}{45.7} \\
\quad ExOPD   & \underline{70.2} & \underline{54.4} & \textbf{95.6} & 30.4 & \underline{90.8} & 44.8 & \textbf{93.4} & \underline{68.5} \\
\rowcolor{ngreenlight}
\quad + \ours ($\alpha$=0.5)   & 68.3 & 50.6 & 93.4 & \underline{34.2} & \textbf{91.2} & \underline{45.4} & \underline{93.3} & 68.1 \\
\rowcolor{ngreenlight}
\quad + \ours ($\alpha$=1)   & \textbf{74.2} & \textbf{62.9} & \underline{95.3} & \textbf{41.2} & 90.2 & \textbf{45.7} & 92.2 & \textbf{71.7} \\
\bottomrule
\end{tabular}}
\label{tab:qwen_nothink_math_v2_exopd}
\end{table}

\begin{table}[h]
\centering
\caption{\textbf{Full results of Qwen3 models on code and science benchmarks with ExOPD (non-thinking mode).} The best and second-best results among the OPD variants are shown in bold and underlined, respectively. \textbf{Avg} is the per-domain average.}
\label{tab:nothink_code_science_results_v2_exopd}
\small
\setlength{\tabcolsep}{7.8pt}
\begin{tabular}{lcccc|cccc}
\toprule
& \multicolumn{4}{c|}{\textbf{Code}} & \multicolumn{4}{c}{\textbf{Science}} \\
\cmidrule(lr){2-5} \cmidrule(lr){6-9}
\textbf{Model}
& \begin{tabular}[c]{@{}c@{}}Code\\Forces\end{tabular}
& LCBv5
& \begin{tabular}[c]{@{}c@{}}RG\\Algo\end{tabular}
& \textbf{Avg}
& GPQA
& \begin{tabular}[c]{@{}c@{}}Super\\GPQA\end{tabular}
& \begin{tabular}[c]{@{}c@{}}Sci\\Bench\end{tabular}
& \textbf{Avg} \\
\midrule
Qwen3-1.7B   & \textcolor{gray}{9.2} & \textcolor{gray}{8.3} & \textcolor{gray}{19.5} & \textcolor{gray}{12.3}
               & \textcolor{gray}{40.9} & \textcolor{gray}{23.8} & \textcolor{gray}{27.6} & \textcolor{gray}{30.8} \\
\quad ExOPD   & \textbf{21.0} & \underline{33.3} & 36.9 & \textbf{30.4}
               & 47.9 & \textbf{29.4} & 37.5 & 38.2 \\
\rowcolor{ngreenlight}
\quad + \ours ($\alpha$=0.5)   & \underline{15.2} & \textbf{34.7} & \underline{37.0} & 29.0
               & \underline{52.5} & 27.9 & \underline{38.0} & \underline{39.4} \\
\rowcolor{ngreenlight}
\quad + \ours ($\alpha$=1)   & 13.5 & 32.5 & \textbf{44.6} & \underline{30.2}
               & \textbf{56.4} & \underline{29.0} & \textbf{39.2} & \textbf{41.5} \\
\midrule
Qwen3-4B   & \textcolor{gray}{16.1} & \textcolor{gray}{26.6} & \textcolor{gray}{32.0} & \textcolor{gray}{24.9}
               & \textcolor{gray}{43.1} & \textcolor{gray}{31.5} & \textcolor{gray}{46.6} & \textcolor{gray}{40.4} \\
\quad ExOPD   & \underline{31.3} & \underline{46.3} & 55.8 & \underline{44.5}
               & 56.6 & \underline{37.1} & 52.0 & 48.6 \\
\rowcolor{ngreenlight}
\quad + \ours ($\alpha$=0.5)   & 29.6 & 45.7 & \underline{56.1} & 43.8
               & \underline{57.3} & 36.6 & \textbf{57.2} & \underline{50.4} \\
\rowcolor{ngreenlight}
\quad + \ours ($\alpha$=1)   & \textbf{41.9} & \textbf{51.5} & \textbf{60.1} & \textbf{51.2}
               & \textbf{60.0} & \textbf{38.7} & \underline{55.8} & \textbf{51.5} \\
\midrule
Qwen3-8B   & \textcolor{gray}{21.1} & \textcolor{gray}{28.5} & \textcolor{gray}{39.9} & \textcolor{gray}{29.8}
               & \textcolor{gray}{52.1} & \textcolor{gray}{31.4} & \textcolor{gray}{45.6} & \textcolor{gray}{43.0} \\
\quad ExOPD   & 31.6 & 46.3 & \underline{58.8} & \underline{45.6}
               & 63.0 & \underline{39.3} & \underline{52.6} & \underline{51.6} \\
\rowcolor{ngreenlight}
\quad + \ours ($\alpha$=0.5)   & \textbf{33.8} & \textbf{49.8} & 52.4 & 45.4
               & \underline{63.1} & 37.7 & 52.5 & 51.1 \\
\rowcolor{ngreenlight}
\quad + \ours ($\alpha$=1)   & \underline{32.5} & \underline{47.4} & \textbf{61.8} & \textbf{47.2}
               & \textbf{67.4} & \textbf{41.3} & \textbf{53.9} & \textbf{54.2} \\
\bottomrule
\end{tabular}
\label{tab:qwen_nothink_other_v2_exopd}
\end{table}

\begin{table}[h]
\centering
\caption{\textbf{Full results of Qwen3 models on mathematical reasoning benchmarks with OPD$^2$ (non-thinking mode).} The best and second-best results among the OPD variants are shown in bold and underlined, respectively.}
\label{tab:nothink_math_results_v2_opd2}
\small
\setlength{\tabcolsep}{5pt}
\resizebox{\textwidth}{!}{
\begin{tabular}{lcccccccc}
\toprule
\textbf{Model} & AIME24 & AIME25 & AMC23 & HMMT25 & MATH500 & Olympiad & RGMath & \textbf{Avg} \\
\midrule
Qwen3-1.7B   & \textcolor{gray}{11.5} & \textcolor{gray}{10.0} & \textcolor{gray}{45.9} & \textcolor{gray}{5.4} & \textcolor{gray}{69.0} & \textcolor{gray}{25.9} & \textcolor{gray}{82.5} & \textcolor{gray}{35.7} \\
\quad OPD$^2$   & \underline{39.2} & 28.1 & \underline{76.9} & 13.8 & 84.2 & 38.7 & 90.9 & 53.1 \\
\rowcolor{ngreenlight}
\quad + \ours ($\alpha$=0.5)   & \textbf{39.2} & \underline{30.0} & \textbf{77.2} & \textbf{19.2} & \underline{85.2} & \textbf{39.8} & \textbf{92.3} & \textbf{54.7} \\
\rowcolor{ngreenlight}
\quad + \ours ($\alpha$=1)   & 38.1 & \textbf{31.0} & 76.6 & \underline{18.8} & \textbf{85.4} & \underline{39.5} & \underline{91.1} & \underline{54.4} \\
\midrule
Qwen3-4B   & \textcolor{gray}{24.2} & \textcolor{gray}{20.0} & \textcolor{gray}{69.4} & \textcolor{gray}{11.7} & \textcolor{gray}{79.0} & \textcolor{gray}{34.7} & \textcolor{gray}{88.9} & \textcolor{gray}{46.8} \\
\quad OPD$^2$   & 60.8 & 50.4 & 92.8 & 31.7 & \underline{91.0} & \underline{45.4} & \underline{92.3} & 66.3 \\
\rowcolor{ngreenlight}
\quad + \ours ($\alpha$=0.5)   & \underline{65.6} & \textbf{59.4} & \textbf{96.9} & \underline{38.7} & 90.4 & \textbf{45.6} & \textbf{94.0} & \textbf{70.1} \\
\rowcolor{ngreenlight}
\quad + \ours ($\alpha$=1)   & \textbf{67.7} & \underline{58.1} & \underline{96.2} & \textbf{39.2} & \textbf{91.4} & 45.3 & 91.5 & \underline{69.9} \\
\midrule
Qwen3-8B   & \textcolor{gray}{26.5} & \textcolor{gray}{18.5} & \textcolor{gray}{66.9} & \textcolor{gray}{5.4} & \textcolor{gray}{78.8} & \textcolor{gray}{34.3} & \textcolor{gray}{89.6} & \textcolor{gray}{45.7} \\
\quad OPD$^2$   & 67.5 & 54.2 & 95.0 & 32.1 & \underline{91.8} & \underline{46.1} & 91.4 & 68.3 \\
\rowcolor{ngreenlight}
\quad + \ours ($\alpha$=0.5)   & \underline{73.8} & \underline{60.0} & \underline{95.6} & \underline{41.2} & 90.6 & \textbf{46.6} & \underline{94.7} & \underline{71.8} \\
\rowcolor{ngreenlight}
\quad + \ours ($\alpha$=1)   & \textbf{75.6} & \textbf{62.9} & \textbf{97.2} & \textbf{41.7} & \textbf{92.2} & 45.7 & \textbf{94.9} & \textbf{72.9} \\
\bottomrule
\end{tabular}
}
\label{tab:qwen_nothink_math_v2_opd2}
\end{table}

\begin{table}[h]
\centering
\caption{\textbf{Full results of Qwen3 models on code and science benchmarks with OPD$^2$ (non-thinking mode).} The best and second-best results among the OPD variants are shown in bold and underlined, respectively. \textbf{Avg} is the per-domain average.}
\label{tab:nothink_code_science_results_v2_opd2}
\small
\setlength{\tabcolsep}{7.8pt}
\begin{tabular}{lcccc|cccc}
\toprule
& \multicolumn{4}{c|}{\textbf{Code}} & \multicolumn{4}{c}{\textbf{Science}} \\
\cmidrule(lr){2-5} \cmidrule(lr){6-9}
\textbf{Model}
& \begin{tabular}[c]{@{}c@{}}Code\\Forces\end{tabular}
& LCBv5
& \begin{tabular}[c]{@{}c@{}}RG\\Algo\end{tabular}
& \textbf{Avg}
& GPQA
& \begin{tabular}[c]{@{}c@{}}Super\\GPQA\end{tabular}
& \begin{tabular}[c]{@{}c@{}}Sci\\Bench\end{tabular}
& \textbf{Avg} \\
\midrule
Qwen3-1.7B   & \textcolor{gray}{9.2} & \textcolor{gray}{8.3} & \textcolor{gray}{19.5} & \textcolor{gray}{12.3}
               & \textcolor{gray}{40.9} & \textcolor{gray}{23.8} & \textcolor{gray}{27.6} & \textcolor{gray}{30.8} \\
\quad OPD$^2$   & \textbf{21.3} & \underline{33.5} & \underline{40.3} & \underline{31.7}
               & 51.0 & \textbf{28.8} & 38.6 & 39.5 \\
\rowcolor{ngreenlight}
\quad + \ours ($\alpha$=0.5)   & \underline{19.6} & \textbf{36.0} & 40.2 & \textbf{31.9}
               & \textbf{60.3} & \underline{28.7} & \textbf{40.4} & \textbf{43.1} \\
\rowcolor{ngreenlight}
\quad + \ours ($\alpha$=1)   & 15.3 & 30.9 & \textbf{40.6} & 28.9
               & \underline{55.1} & 28.2 & \underline{38.9} & \underline{40.7} \\
\midrule
Qwen3-4B   & \textcolor{gray}{16.1} & \textcolor{gray}{26.6} & \textcolor{gray}{32.0} & \textcolor{gray}{24.9}
               & \textcolor{gray}{43.1} & \textcolor{gray}{31.5} & \textcolor{gray}{46.6} & \textcolor{gray}{40.4} \\
\quad OPD$^2$   & 30.9 & \underline{46.6} & 62.7 & 46.8
               & 57.5 & 36.6 & 53.9 & 49.4 \\
\rowcolor{ngreenlight}
\quad + \ours ($\alpha$=0.5)   & \textbf{35.9} & 44.6 & \underline{63.1} & \underline{47.9}
               & \underline{58.0} & \underline{38.2} & \textbf{57.9} & \textbf{51.4} \\
\rowcolor{ngreenlight}
\quad + \ours ($\alpha$=1)   & \underline{35.1} & \textbf{50.9} & \textbf{67.5} & \textbf{51.2}
               & \textbf{59.2} & \textbf{39.6} & \underline{54.5} & \underline{51.1} \\
\midrule
Qwen3-8B   & \textcolor{gray}{21.1} & \textcolor{gray}{28.5} & \textcolor{gray}{39.9} & \textcolor{gray}{29.8}
               & \textcolor{gray}{52.1} & \textcolor{gray}{31.4} & \textcolor{gray}{45.6} & \textcolor{gray}{43.0} \\
\quad OPD$^2$   & 33.6 & 44.4 & 50.1 & 42.7
               & 64.3 & \underline{40.2} & \textbf{53.2} & \underline{52.6} \\
\rowcolor{ngreenlight}
\quad + \ours ($\alpha$=0.5)   & \underline{34.1} & \underline{48.6} & \underline{62.7} & \underline{48.5}
               & \underline{66.3} & 39.5 & \underline{51.4} & 52.4 \\
\rowcolor{ngreenlight}
\quad + \ours ($\alpha$=1)   & \textbf{37.3} & \textbf{57.2} & \textbf{64.4} & \textbf{53.0}
               & \textbf{67.5} & \textbf{41.1} & 51.0 & \textbf{53.2} \\
\bottomrule
\end{tabular}
\label{tab:qwen_nothink_other_v2_opd2}
\end{table}

%% file: figs/summary_think.tex
\begin{table}[h]
\centering
\caption{\textbf{Summary of Qwen3 results in thinking mode.}
The table reports average performance across Math, Code, and Science benchmarks. The best and second-best results are shown in bold and underlined, respectively.
}
\label{tab:qwen_summary_think}
\small
\setlength{\tabcolsep}{5pt}

\begin{tabular}{lccc|ccc|ccc}
\toprule
& \multicolumn{3}{c|}{\textbf{Qwen3-1.7B}}
& \multicolumn{3}{c|}{\textbf{Qwen3-4B}}
& \multicolumn{3}{c}{\textbf{Qwen3-8B}} \\
\cmidrule(lr){2-4} \cmidrule(lr){5-7} \cmidrule(lr){8-10}

\textbf{Model}
& Math & Code & Sci.
& Math & Code & Sci.
& Math & Code & Sci. \\
\midrule
Baseline
& 60.1 & 33.5 & 39.5
& 72.9 & 53.2 & 51.8
& 73.3 & 52.7 & 53.6 \\
\quad OPD
& 60.0 & 34.3 & 40.9
& 70.7 & 52.9 & 49.8
& 71.7 & 56.4 & 52.2 \\
\rowcolor{ngreenlight}
\quad + \ours
& 61.8 & 37.6 & 41.9
& 72.2 & 53.4 & 50.6
& 74.3 & 54.0 & 53.1 \\
\midrule
\quad ExOPD
& 63.1 & 38.2 & 43.4
& 72.8 & \underline{58.7} & 51.5
& 74.2 & 60.3 & 53.6 \\
\rowcolor{ngreenlight}
\quad + \ours
& 63.4 & \underline{39.0} & 43.2
& \underline{76.0} & 57.6 & \textbf{53.2}
& \underline{76.1} & 59.6 & \textbf{55.9} \\
\midrule
\quad OPD$^2$
& \underline{64.3} & 38.5 & \underline{44.2}
& 74.0 & 58.4 & 52.3
& 75.0 & \underline{61.4} & 52.5 \\
\rowcolor{ngreenlight}
\quad + \ours
& \textbf{65.3} & \textbf{43.0} & \textbf{45.0}
& \textbf{76.8} & \textbf{60.6} & \underline{52.6}
& \textbf{77.1} & \textbf{61.7} & \underline{55.3} \\
\bottomrule
\end{tabular}%
\vspace{-4pt}
\end{table}

%% file: figs/think_rewards.tex
\begin{table}[h]
\centering
\caption{\textbf{Full results of Qwen3 models on mathematical reasoning benchmarks with OPD (thinking mode).} The best and second-best results among the OPD variants are shown in bold and underlined, respectively.}
\label{tab:think_math_results_v2_opd}
\small
\setlength{\tabcolsep}{5pt}
\resizebox{\textwidth}{!}{
\begin{tabular}{lcccccccc}
\toprule
\textbf{Model} & AIME24 & AIME25 & AMC23 & HMMT25 & MATH500 & Olympiad & RGMath & \textbf{Avg} \\
\midrule
Qwen3-1.7B   & \textcolor{gray}{53.1} & \textcolor{gray}{35.0} & \textcolor{gray}{85.0} & \textcolor{gray}{22.9} & \textcolor{gray}{87.2} & \textcolor{gray}{39.7} & \textcolor{gray}{97.8} & \textcolor{gray}{60.1} \\
\quad OPD   & 50.0 & 35.6 & 84.4 & 25.4 & \underline{87.4} & 39.8 & 97.5 & 60.0 \\
\rowcolor{ngreenlight}
\quad + \ours ($\alpha$=0.25)   & 51.0 & 36.5 & 84.1 & 25.4 & 86.4 & 40.5 & \textbf{98.1} & 60.3 \\
\rowcolor{ngreenlight}
\quad + \ours ($\alpha$=0.5)   & 51.5 & \textbf{39.2} & 82.8 & 25.4 & 86.6 & \underline{40.8} & 96.1 & 60.3 \\
\rowcolor{ngreenlight}
\quad + \ours ($\alpha$=0.75)   & \underline{52.9} & 36.5 & \underline{84.4} & \underline{26.7} & 87.0 & 40.7 & 97.0 & \underline{60.7} \\
\rowcolor{ngreenlight}
\quad + \ours ($\alpha$=1)   & \textbf{54.2} & \underline{38.5} & \textbf{86.6} & \textbf{27.1} & \textbf{87.4} & \textbf{41.3} & \underline{97.9} & \textbf{61.8} \\
\midrule
Qwen3-4B   & \textcolor{gray}{72.9} & \textcolor{gray}{61.9} & \textcolor{gray}{95.9} & \textcolor{gray}{44.2} & \textcolor{gray}{91.2} & \textcolor{gray}{45.1} & \textcolor{gray}{99.1} & \textcolor{gray}{72.9} \\
\quad OPD   & 67.9 & 54.4 & 97.5 & 39.6 & 91.0 & 46.1 & 98.2 & 70.7 \\
\rowcolor{ngreenlight}
\quad + \ours ($\alpha$=0.25)   & 69.6 & \underline{60.8} & 96.9 & 39.6 & 90.4 & 45.9 & 98.5 & 71.7 \\
\rowcolor{ngreenlight}
\quad + \ours ($\alpha$=0.5)   & 70.4 & 57.5 & \underline{97.5} & 43.3 & 91.2 & 46.3 & \textbf{99.0} & 72.2 \\
\rowcolor{ngreenlight}
\quad + \ours ($\alpha$=0.75)   & \underline{74.2} & 59.0 & \textbf{98.1} & \underline{45.0} & \underline{91.2} & \underline{46.8} & 97.7 & \underline{73.1} \\
\rowcolor{ngreenlight}
\quad + \ours ($\alpha$=1)   & \textbf{75.0} & \textbf{63.3} & 95.9 & \textbf{47.1} & \textbf{91.4} & \textbf{46.9} & \underline{98.7} & \textbf{74.1} \\
\midrule
Qwen3-8B   & \textcolor{gray}{75.6} & \textcolor{gray}{65.6} & \textcolor{gray}{94.1} & \textcolor{gray}{46.7} & \textcolor{gray}{91.8} & \textcolor{gray}{39.7} & \textcolor{gray}{99.4} & \textcolor{gray}{73.3} \\
\quad OPD   & 72.3 & 57.1 & 97.2 & 37.5 & \textbf{91.8} & 46.2 & \underline{100.0} & 71.7 \\
\rowcolor{ngreenlight}
\quad + \ours ($\alpha$=0.25)   & 72.9 & 61.0 & 97.2 & 42.9 & 91.2 & 47.2 & 99.1 & 73.1 \\
\rowcolor{ngreenlight}
\quad + \ours ($\alpha$=0.5)   & 73.1 & 60.6 & \textbf{97.8} & 42.1 & 91.4 & 46.8 & 99.7 & 73.1 \\
\rowcolor{ngreenlight}
\quad + \ours ($\alpha$=0.75)   & \underline{74.2} & \textbf{62.3} & \underline{97.2} & \textbf{48.8} & 90.6 & \underline{47.6} & 99.8 & \textbf{74.3} \\
\rowcolor{ngreenlight}
\quad + \ours ($\alpha$=1)   & \textbf{74.4} & \underline{61.7} & 95.3 & \underline{44.2} & \underline{91.6} & \textbf{48.0} & \textbf{100.0} & \underline{73.6} \\
\bottomrule
\end{tabular}
}
\label{tab:qwen_think_math_v2_opd}
\end{table}

\begin{table}[h]
\centering
\caption{\textbf{Full results of Qwen3 models on code and science benchmarks with OPD (thinking mode).} The best and second-best results among the OPD variants are shown in bold and underlined, respectively. }
\label{tab:think_code_science_results_v2_opd}
\small
\setlength{\tabcolsep}{7.8pt}
\begin{tabular}{lcccc|cccc}
\toprule
& \multicolumn{4}{c|}{\textbf{Code}} & \multicolumn{4}{c}{\textbf{Science}} \\
\cmidrule(lr){2-5} \cmidrule(lr){6-9}
\textbf{Model}
& \begin{tabular}[c]{@{}c@{}}Code\\Forces\end{tabular}
& LCBv5
& \begin{tabular}[c]{@{}c@{}}RG\\Algo\end{tabular}
& \textbf{Avg}
& GPQA
& \begin{tabular}[c]{@{}c@{}}Super\\GPQA\end{tabular}
& \begin{tabular}[c]{@{}c@{}}Sci\\Bench\end{tabular}
& \textbf{Avg} \\
\midrule
Qwen3-1.7B   & \textcolor{gray}{22.5} & \textcolor{gray}{33.5} & \textcolor{gray}{44.6} & \textcolor{gray}{33.5}
               & \textcolor{gray}{47.5} & \textcolor{gray}{28.7} & \textcolor{gray}{42.4} & \textcolor{gray}{39.5} \\
\quad OPD   & 22.5 & 36.9 & 43.6 & 34.3
               & 49.1 & \textbf{29.8} & 43.8 & 40.9 \\
\rowcolor{ngreenlight}
\quad + \ours ($\alpha$=0.25)   & \underline{25.3} & 38.5 & 42.9 & 35.6
               & 49.2 & 27.6 & 43.4 & 40.1 \\
\rowcolor{ngreenlight}
\quad + \ours ($\alpha$=0.5)   & \textbf{25.6} & \underline{39.2} & 45.8 & \underline{36.9}
               & 49.2 & 28.7 & 43.7 & 40.5 \\
\rowcolor{ngreenlight}
\quad + \ours ($\alpha$=0.75)   & 24.7 & 37.3 & \underline{46.2} & 36.1
               & \underline{50.7} & 28.8 & \underline{43.9} & \underline{41.1} \\
\rowcolor{ngreenlight}
\quad + \ours ($\alpha$=1)   & 25.1 & \textbf{41.0} & \textbf{46.6} & \textbf{37.6}
               & \textbf{51.4} & \underline{29.6} & \textbf{44.8} & \textbf{41.9} \\
\midrule
Qwen3-4B   & \textcolor{gray}{39.7} & \textcolor{gray}{58.9} & \textcolor{gray}{60.8} & \textcolor{gray}{53.2}
               & \textcolor{gray}{57.0} & \textcolor{gray}{38.5} & \textcolor{gray}{60.0} & \textcolor{gray}{51.8} \\
\quad OPD   & 38.9 & \textbf{59.3} & 60.5 & 52.9
               & 52.8 & 36.9 & 59.6 & 49.8 \\
\rowcolor{ngreenlight}
\quad + \ours ($\alpha$=0.25)   & \underline{39.0} & \underline{58.4} & \textbf{62.8} & \textbf{53.4}
               & 53.4 & 38.3 & 58.6 & 50.1 \\
\rowcolor{ngreenlight}
\quad + \ours ($\alpha$=0.5)   & \textbf{39.1} & 58.2 & \underline{62.8} & \underline{53.4}
               & \underline{54.4} & 37.5 & 59.8 & 50.6 \\
\rowcolor{ngreenlight}
\quad + \ours ($\alpha$=0.75)   & 38.6 & 56.8 & 61.1 & 52.2
               & 53.2 & \underline{38.7} & \textbf{62.4} & \underline{51.4} \\
\rowcolor{ngreenlight}
\quad + \ours ($\alpha$=1)   & 37.6 & 55.8 & 55.4 & 49.6
               & \textbf{54.9} & \textbf{39.3} & \underline{61.9} & \textbf{52.0} \\
\midrule
Qwen3-8B   & \textcolor{gray}{40.1} & \textcolor{gray}{50.9} & \textcolor{gray}{67.1} & \textcolor{gray}{52.7}
               & \textcolor{gray}{63.2} & \textcolor{gray}{39.5} & \textcolor{gray}{58.0} & \textcolor{gray}{53.6} \\
\quad OPD   & \textbf{41.6} & \textbf{59.6} & \textbf{68.0} & \textbf{56.4}
               & \underline{60.0} & 39.3 & 57.3 & 52.2 \\
\rowcolor{ngreenlight}
\quad + \ours ($\alpha$=0.25)   & 41.1 & \underline{57.2} & 63.9 & \underline{54.1}
               & 59.6 & 40.1 & 58.2 & 52.6 \\
\rowcolor{ngreenlight}
\quad + \ours ($\alpha$=0.5)   & \underline{41.2} & 56.7 & 60.5 & 52.8
               & 59.9 & \underline{40.9} & 58.7 & \underline{53.2} \\
\rowcolor{ngreenlight}
\quad + \ours ($\alpha$=0.75)   & 39.0 & 56.2 & \underline{66.8} & 54.0
               & 60.0 & 40.7 & \underline{58.7} & 53.1 \\
\rowcolor{ngreenlight}
\quad + \ours ($\alpha$=1)   & 37.6 & 47.6 & 65.8 & 50.3
               & \textbf{60.5} & \textbf{41.8} & \textbf{60.8} & \textbf{54.4} \\
\bottomrule
\end{tabular}
\label{tab:qwen_think_other_v2_opd}
\end{table}

\begin{table}[h]
\centering
\caption{\textbf{Full results of Qwen3 models on mathematical reasoning benchmarks with ExOPD (thinking mode).} The best and second-best results among the OPD variants are shown in bold and underlined, respectively.}
\label{tab:think_math_results_v2_exopd}
\small
\setlength{\tabcolsep}{5pt}
\resizebox{\textwidth}{!}{
\begin{tabular}{lcccccccc}
\toprule
\textbf{Model} & AIME24 & AIME25 & AMC23 & HMMT25 & MATH500 & Olympiad & RGMath & \textbf{Avg} \\
\midrule
Qwen3-1.7B   & \textcolor{gray}{53.1} & \textcolor{gray}{35.0} & \textcolor{gray}{85.0} & \textcolor{gray}{22.9} & \textcolor{gray}{87.2} & \textcolor{gray}{39.7} & \textcolor{gray}{97.8} & \textcolor{gray}{60.1} \\
\quad ExOPD   & \textbf{59.0} & 40.6 & \underline{88.4} & 25.8 & \underline{87.6} & \textbf{42.0} & \underline{98.5} & 63.1 \\
\rowcolor{ngreenlight}
\quad + \ours ($\alpha$=0.5)   & 56.7 & \underline{40.8} & 87.5 & \textbf{32.1} & 87.4 & 41.6 & 98.2 & \textbf{63.5} \\
\rowcolor{ngreenlight}
\quad + \ours ($\alpha$=1)   & \underline{57.9} & \textbf{41.5} & \textbf{88.8} & \underline{27.5} & \textbf{87.8} & \underline{41.8} & \textbf{98.6} & \underline{63.4} \\
\midrule
Qwen3-4B   & \textcolor{gray}{72.9} & \textcolor{gray}{61.9} & \textcolor{gray}{95.9} & \textcolor{gray}{44.2} & \textcolor{gray}{91.2} & \textcolor{gray}{45.1} & \textcolor{gray}{99.1} & \textcolor{gray}{72.9} \\
\quad ExOPD   & 69.6 & \underline{63.3} & 97.8 & 41.7 & \underline{90.6} & 47.2 & \textbf{99.7} & 72.8 \\
\rowcolor{ngreenlight}
\quad + \ours ($\alpha$=0.5)   & \underline{73.3} & 63.1 & \underline{98.4} & \underline{47.5} & 89.6 & \underline{47.4} & \underline{99.2} & \underline{74.1} \\
\rowcolor{ngreenlight}
\quad + \ours ($\alpha$=1)   & \textbf{78.8} & \textbf{68.1} & \textbf{99.4} & \textbf{48.8} & \textbf{90.8} & \textbf{47.8} & 98.7 & \textbf{76.0} \\
\midrule
Qwen3-8B   & \textcolor{gray}{75.6} & \textcolor{gray}{65.6} & \textcolor{gray}{94.1} & \textcolor{gray}{46.7} & \textcolor{gray}{91.8} & \textcolor{gray}{39.7} & \textcolor{gray}{99.4} & \textcolor{gray}{73.3} \\
\quad ExOPD   & \underline{76.5} & 63.3 & 96.6 & 44.2 & \textbf{92.2} & 47.4 & \underline{99.3} & 74.2 \\
\rowcolor{ngreenlight}
\quad + \ours ($\alpha$=0.5)   & 76.2 & \underline{66.0} & \underline{97.5} & \underline{49.2} & \underline{91.6} & \underline{47.5} & \textbf{99.7} & \underline{75.4} \\
\rowcolor{ngreenlight}
\quad + \ours ($\alpha$=1)   & \textbf{78.5} & \textbf{69.8} & \textbf{97.8} & \textbf{50.0} & 90.2 & \textbf{47.7} & 99.0 & \textbf{76.1} \\
\bottomrule
\end{tabular}
}
\label{tab:qwen_think_math_v2_exopd}
\end{table}

\begin{table}[h]
\centering
\caption{\textbf{Full results of Qwen3 models on code and science benchmarks with ExOPD (thinking mode).} The best and second-best results among the OPD variants are shown in bold and underlined, respectively. }
\label{tab:think_code_science_results_v2_exopd}
\small
\setlength{\tabcolsep}{7.8pt}
\begin{tabular}{lcccc|cccc}
\toprule
& \multicolumn{4}{c|}{\textbf{Code}} & \multicolumn{4}{c}{\textbf{Science}} \\
\cmidrule(lr){2-5} \cmidrule(lr){6-9}
\textbf{Model}
& \begin{tabular}[c]{@{}c@{}}Code\\Forces\end{tabular}
& LCBv5
& \begin{tabular}[c]{@{}c@{}}RG\\Algo\end{tabular}
& \textbf{Avg}
& GPQA
& \begin{tabular}[c]{@{}c@{}}Super\\GPQA\end{tabular}
& \begin{tabular}[c]{@{}c@{}}Sci\\Bench\end{tabular}
& \textbf{Avg} \\
\midrule
Qwen3-1.7B   & \textcolor{gray}{22.5} & \textcolor{gray}{33.5} & \textcolor{gray}{44.6} & \textcolor{gray}{33.5}
               & \textcolor{gray}{47.5} & \textcolor{gray}{28.7} & \textcolor{gray}{42.4} & \textcolor{gray}{39.5} \\
\quad ExOPD   & 26.2 & \textbf{40.6} & 47.9 & 38.2
               & \underline{54.5} & \underline{30.4} & \textbf{45.2} & \textbf{43.4} \\
\rowcolor{ngreenlight}
\quad + \ours ($\alpha$=0.5)   & \textbf{27.7} & 39.7 & \textbf{50.8} & \textbf{39.4}
               & 52.8 & \textbf{30.6} & 44.5 & 42.6 \\
\rowcolor{ngreenlight}
\quad + \ours ($\alpha$=1)   & \underline{27.5} & \underline{40.3} & \underline{49.2} & \underline{39.0}
               & \textbf{55.7} & 29.3 & \underline{44.7} & \underline{43.2} \\
\midrule
Qwen3-4B   & \textcolor{gray}{39.7} & \textcolor{gray}{58.9} & \textcolor{gray}{60.8} & \textcolor{gray}{53.2}
               & \textcolor{gray}{57.0} & \textcolor{gray}{38.5} & \textcolor{gray}{60.0} & \textcolor{gray}{51.8} \\
\quad ExOPD   & \textbf{44.7} & \textbf{64.3} & \underline{67.1} & \textbf{58.7}
               & \underline{56.3} & \textbf{39.3} & 58.9 & 51.5 \\
\rowcolor{ngreenlight}
\quad + \ours ($\alpha$=0.5)   & 41.2 & 57.0 & 63.4 & 53.9
               & 55.2 & 38.7 & \underline{60.9} & \underline{51.6} \\
\rowcolor{ngreenlight}
\quad + \ours ($\alpha$=1)   & \underline{43.1} & \underline{61.1} & \textbf{68.5} & \underline{57.6}
               & \textbf{58.3} & \underline{39.0} & \textbf{62.3} & \textbf{53.2} \\
\midrule
Qwen3-8B   & \textcolor{gray}{40.1} & \textcolor{gray}{50.9} & \textcolor{gray}{67.1} & \textcolor{gray}{52.7}
               & \textcolor{gray}{63.2} & \textcolor{gray}{39.5} & \textcolor{gray}{58.0} & \textcolor{gray}{53.6} \\
\quad ExOPD   & \textbf{47.2} & \underline{63.6} & \textbf{70.1} & \textbf{60.3}
               & \underline{63.9} & 41.0 & 56.0 & \underline{53.6} \\
\rowcolor{ngreenlight}
\quad + \ours ($\alpha$=0.5)   & 43.7 & 61.5 & 68.4 & 57.9
               & 60.5 & \underline{41.5} & \underline{58.4} & 53.5 \\
\rowcolor{ngreenlight}
\quad + \ours ($\alpha$=1)   & \underline{44.8} & \textbf{63.9} & \underline{70.1} & \underline{59.6}
               & \textbf{64.1} & \textbf{42.8} & \textbf{60.7} & \textbf{55.9} \\
\bottomrule
\end{tabular}
\label{tab:qwen_think_other_v2_exopd}
\end{table}

\begin{table}[h]
\centering
\caption{\textbf{Full results of Qwen3 models on mathematical reasoning benchmarks with OPD$^2$ (thinking mode).} The best and second-best results among the OPD variants are shown in bold and underlined, respectively.}
\label{tab:think_math_results_v2_opd2}
\small
\setlength{\tabcolsep}{5pt}
\resizebox{\textwidth}{!}{
\begin{tabular}{lcccccccc}
\toprule
\textbf{Model} & AIME24 & AIME25 & AMC23 & HMMT25 & MATH500 & Olympiad & RGMath & \textbf{Avg} \\
\midrule
Qwen3-1.7B   & \textcolor{gray}{53.1} & \textcolor{gray}{35.0} & \textcolor{gray}{85.0} & \textcolor{gray}{22.9} & \textcolor{gray}{87.2} & \textcolor{gray}{39.7} & \textcolor{gray}{97.8} & \textcolor{gray}{60.1} \\
\quad OPD$^2$   & \textbf{62.7} & 40.0 & 88.8 & 29.2 & 87.8 & \underline{42.6} & \underline{99.0} & 64.3 \\
\rowcolor{ngreenlight}
\quad + \ours ($\alpha$=0.5)   & \underline{61.9} & \textbf{43.8} & \underline{88.8} & \textbf{32.1} & \textbf{88.6} & 42.5 & \textbf{99.5} & \textbf{65.3} \\
\rowcolor{ngreenlight}
\quad + \ours ($\alpha$=1)   & 60.8 & \underline{42.9} & \textbf{90.6} & \underline{30.4} & \underline{88.2} & \textbf{42.8} & 98.7 & \underline{64.9} \\
\midrule
Qwen3-4B   & \textcolor{gray}{72.9} & \textcolor{gray}{61.9} & \textcolor{gray}{95.9} & \textcolor{gray}{44.2} & \textcolor{gray}{91.2} & \textcolor{gray}{45.1} & \textcolor{gray}{99.1} & \textcolor{gray}{72.9} \\
\quad OPD$^2$   & 70.6 & 64.2 & 96.2 & 48.3 & \underline{91.6} & \underline{47.8} & 99.2 & 74.0 \\
\rowcolor{ngreenlight}
\quad + \ours ($\alpha$=0.5)   & \underline{77.7} & \underline{68.8} & \underline{98.4} & \textbf{53.8} & \textbf{91.8} & \textbf{48.2} & \underline{99.2} & \textbf{76.8} \\
\rowcolor{ngreenlight}
\quad + \ours ($\alpha$=1)   & \textbf{77.9} & \textbf{70.4} & \textbf{98.8} & \underline{50.0} & 91.4 & 47.0 & \textbf{99.3} & \underline{76.4} \\
\midrule
Qwen3-8B   & \textcolor{gray}{75.6} & \textcolor{gray}{65.6} & \textcolor{gray}{94.1} & \textcolor{gray}{46.7} & \textcolor{gray}{91.8} & \textcolor{gray}{39.7} & \textcolor{gray}{99.4} & \textcolor{gray}{73.3} \\
\quad OPD$^2$   & 73.8 & 66.5 & \underline{97.8} & 47.9 & \underline{91.8} & 47.6 & \textbf{99.7} & 75.0 \\
\rowcolor{ngreenlight}
\quad + \ours ($\alpha$=0.5)   & \textbf{81.7} & \underline{70.6} & \textbf{98.4} & \underline{50.4} & \textbf{91.8} & \textbf{48.7} & 98.3 & \underline{77.1} \\
\rowcolor{ngreenlight}
\quad + \ours ($\alpha$=1)   & \underline{81.5} & \textbf{71.2} & 97.8 & \textbf{52.9} & 90.6 & \underline{48.5} & \underline{99.3} & \textbf{77.4} \\
\bottomrule
\end{tabular}
}
\label{tab:qwen_think_math_v2_opd2}
\end{table}

\begin{table}[h]
\centering
\caption{\textbf{Full results of Qwen3 models on code and science benchmarks with OPD$^2$ (thinking mode).} The best and second-best results among the OPD variants are shown in bold and underlined, respectively. }
\label{tab:think_code_science_results_v2_opd2}
\small
\setlength{\tabcolsep}{7.8pt}
\begin{tabular}{lcccc|cccc}
\toprule
& \multicolumn{4}{c|}{\textbf{Code}} & \multicolumn{4}{c}{\textbf{Science}} \\
\cmidrule(lr){2-5} \cmidrule(lr){6-9}
\textbf{Model}
& \begin{tabular}[c]{@{}c@{}}Code\\Forces\end{tabular}
& LCBv5
& \begin{tabular}[c]{@{}c@{}}RG\\Algo\end{tabular}
& \textbf{Avg}
& GPQA
& \begin{tabular}[c]{@{}c@{}}Super\\GPQA\end{tabular}
& \begin{tabular}[c]{@{}c@{}}Sci\\Bench\end{tabular}
& \textbf{Avg} \\
\midrule
Qwen3-1.7B   & \textcolor{gray}{22.5} & \textcolor{gray}{33.5} & \textcolor{gray}{44.6} & \textcolor{gray}{33.5}
               & \textcolor{gray}{47.5} & \textcolor{gray}{28.7} & \textcolor{gray}{42.4} & \textcolor{gray}{39.5} \\
\quad OPD$^2$   & 26.3 & 37.7 & 51.5 & 38.5
               & \underline{58.5} & \underline{29.6} & 44.5 & 44.2 \\
\rowcolor{ngreenlight}
\quad + \ours ($\alpha$=0.5)   & \textbf{30.2} & \textbf{43.9} & \textbf{54.8} & \textbf{43.0}
               & \textbf{60.2} & \textbf{30.2} & \underline{44.7} & \textbf{45.0} \\
\rowcolor{ngreenlight}
\quad + \ours ($\alpha$=1)   & \underline{28.0} & \underline{42.5} & \underline{52.8} & \underline{41.1}
               & 58.3 & 29.4 & \textbf{45.9} & \underline{44.6} \\
\midrule
Qwen3-4B   & \textcolor{gray}{39.7} & \textcolor{gray}{58.9} & \textcolor{gray}{60.8} & \textcolor{gray}{53.2}
               & \textcolor{gray}{57.0} & \textcolor{gray}{38.5} & \textcolor{gray}{60.0} & \textcolor{gray}{51.8} \\
\quad OPD$^2$   & \underline{44.0} & \underline{63.1} & 68.1 & 58.4
               & \textbf{57.5} & \underline{40.0} & 59.4 & 52.3 \\
\rowcolor{ngreenlight}
\quad + \ours ($\alpha$=0.5)   & \textbf{44.5} & \textbf{64.8} & \underline{72.4} & \textbf{60.6}
               & 56.0 & \textbf{40.5} & \underline{61.2} & \underline{52.6} \\
\rowcolor{ngreenlight}
\quad + \ours ($\alpha$=1)   & 43.5 & 62.9 & \textbf{73.0} & \underline{59.8}
               & \underline{57.3} & 39.3 & \textbf{62.7} & \textbf{53.1} \\
\midrule
Qwen3-8B   & \textcolor{gray}{40.1} & \textcolor{gray}{50.9} & \textcolor{gray}{67.1} & \textcolor{gray}{52.7}
               & \textcolor{gray}{63.2} & \textcolor{gray}{39.5} & \textcolor{gray}{58.0} & \textcolor{gray}{53.6} \\
\quad OPD$^2$   & 46.5 & \underline{65.5} & \textbf{72.1} & \underline{61.4}
               & 62.3 & 41.3 & 53.8 & 52.5 \\
\rowcolor{ngreenlight}
\quad + \ours ($\alpha$=0.5)   & \textbf{48.0} & \textbf{66.0} & 71.0 & \textbf{61.7}
               & \underline{63.8} & \textbf{42.4} & \underline{59.6} & \underline{55.3} \\
\rowcolor{ngreenlight}
\quad + \ours ($\alpha$=1)   & \underline{46.7} & 62.8 & \underline{71.4} & 60.3
               & \textbf{64.2} & \underline{42.1} & \textbf{60.6} & \textbf{55.6} \\
\bottomrule
\end{tabular}
\label{tab:qwen_think_other_v2_opd2}
\end{table}

%% file: figs/training_efficiency.tex
\newcommand{\spd}[1]{\,{\scriptsize\textcolor{gray}{(#1$\times$)}}}
\newcommand{\na}{\textcolor{gray!50}{--}}
\definecolor{hl}{RGB}{232,240,254}

\begin{table}[h]
\centering
\caption{\textbf{Wall-clock training time of Qwen3-1.7B on a single node with 8$\times$H100 GPUs.}
$\alpha$ denotes the proportion of \name-policy rollouts.
Gray numbers show the speedup relative to $\alpha{=}0$.}
\label{tab:training-efficiency}
\small
\setlength{\tabcolsep}{5pt}
\renewcommand{\arraystretch}{1.15}
\begin{tabular}{ll rrrr >{}r}
\toprule
& & \multicolumn{5}{c}{Proportion of \name-policy rollouts ($\alpha$)} \\
\cmidrule(lr){3-7}
Mode & - & \multicolumn{1}{c}{$0$} & \multicolumn{1}{c}{$0.25$} & \multicolumn{1}{c}{$0.5$} & \multicolumn{1}{c}{$0.75$} & \multicolumn{1}{c}{$1$} \\
\midrule
\multirow{1}{*}{Thinking}
& OPD      & 9h 19m  & 8h 09m\spd{1.14}  & 6h 57m\spd{1.34}  & 6h 04m\spd{1.54} & {3h 34m}\spd{2.61} \\
\multirow{1}{*}{Non-thinking}
& OPD      & 5h 15m  & 5h 11m\spd{1.01}  & 4h 34m\spd{1.15}  & 3h 50m\spd{1.37} & {1h 16m}\spd{4.14} \\
\bottomrule
\end{tabular}
\end{table}

%% file: figs/merge.tex
\begin{table}[h]
\centering
\caption{\textbf{Model merging with ExOPD on Qwen3-8B (thinking mode).} \textit{merge} averages the weights of the ExOPD model ($\alpha$=0) and the \ours ($\alpha$=1) model with equal weights. The best and second-best results among the ExOPD variants are shown in bold and underlined, respectively.}
\label{tab:think8b_merge_math_exopd}
\small
\setlength{\tabcolsep}{5pt}
\resizebox{\textwidth}{!}{
\begin{tabular}{lcccccccc}
\toprule
\textbf{Model} & AIME24 & AIME25 & AMC23 & HMMT25 & MATH500 & Olympiad & RGMath & \textbf{Avg} \\
\midrule
Qwen3-8B   & \textcolor{gray}{75.6} & \textcolor{gray}{65.6} & \textcolor{gray}{94.1} & \textcolor{gray}{46.7} & \textcolor{gray}{91.8} & \textcolor{gray}{39.7} & \textcolor{gray}{99.4} & \textcolor{gray}{73.3} \\
\quad ExOPD   & 76.5 & 63.3 & 96.6 & 44.2 & \textbf{92.2} & 47.4 & \underline{99.3} & 74.2 \\
\quad + \ours ($\alpha$=1)   & \underline{78.5} & \textbf{69.8} & \textbf{97.8} & \textbf{50.0} & 90.2 & \underline{47.7} & 99.0 & \underline{76.1} \\
\rowcolor{ngreenlight}
\quad + merge ($\alpha$=0 $\oplus$ 1)   & \textbf{80.0} & \underline{67.9} & \textbf{97.8} & \underline{49.6} & \underline{91.8} & \textbf{47.9} & \underline{99.3} & \textbf{76.3} \\
\bottomrule
\end{tabular}
}
\label{tab:qwen_think8b_merge_math_exopd}
\end{table}

\begin{table}[h]
\centering
\caption{\textbf{Model merging with ExOPD on Qwen3-8B: code and science benchmarks (thinking mode).} \textit{merge} averages the weights of the ExOPD model ($\alpha$=0) and the \ours ($\alpha$=1) model with equal weights. The best and second-best results among the ExOPD variants are shown in bold and underlined, respectively.}
\label{tab:think8b_merge_other_exopd}
\small
\setlength{\tabcolsep}{7.8pt}
\begin{tabular}{lcccc|cccc}
\toprule
& \multicolumn{4}{c|}{\textbf{Code}} & \multicolumn{4}{c}{\textbf{Science}} \\
\cmidrule(lr){2-5} \cmidrule(lr){6-9}
\textbf{Model}
& \begin{tabular}[c]{@{}c@{}}Code\\Forces\end{tabular}
& LCBv5
& \begin{tabular}[c]{@{}c@{}}RG\\Algo\end{tabular}
& \textbf{Avg}
& GPQA
& \begin{tabular}[c]{@{}c@{}}Super\\GPQA\end{tabular}
& \begin{tabular}[c]{@{}c@{}}Sci\\Bench\end{tabular}
& \textbf{Avg} \\
\midrule
Qwen3-8B   & \textcolor{gray}{40.1} & \textcolor{gray}{50.9} & \textcolor{gray}{67.1} & \textcolor{gray}{52.7}
               & \textcolor{gray}{63.2} & \textcolor{gray}{39.5} & \textcolor{gray}{58.0} & \textcolor{gray}{53.6} \\
\quad ExOPD   & \textbf{47.2} & 63.6 & \underline{70.1} & \underline{60.3}
               & \underline{63.9} & 41.0 & 56.0 & 53.6 \\
\quad + \ours ($\alpha$=1)   & 44.8 & \underline{63.9} & \underline{70.1} & 59.6
               & \textbf{64.1} & \textbf{42.8} & \textbf{60.7} & \textbf{55.9} \\
\rowcolor{ngreenlight}
\quad + merge ($\alpha$=0 $\oplus$ 1)   & \underline{46.6} & \textbf{65.0} & \textbf{70.7} & \textbf{60.7}
               & 63.6 & 41.4 & \underline{59.3} & \underline{54.7} \\
\bottomrule
\end{tabular}
\label{tab:qwen_think8b_merge_other_exopd}
\end{table}

\begin{table}[h]
\centering
\caption{\textbf{Model merging with OPD$^2$ on Qwen3-8B: mathematical reasoning benchmarks (thinking mode).} \textit{merge} averages the weights of the OPD$^2$ model ($\alpha$=0) and the \ours ($\alpha$=1) model with equal weights. The best and second-best results among the OPD$^2$ variants are shown in bold and underlined, respectively.}
\label{tab:think8b_merge_math_opd2}
\small
\resizebox{\textwidth}{!}{
\setlength{\tabcolsep}{5pt}
\begin{tabular}{lcccccccc}
\toprule
\textbf{Model} & AIME24 & AIME25 & AMC23 & HMMT25 & MATH500 & Olympiad & RGMath & \textbf{Avg} \\
\midrule
Qwen3-8B   & \textcolor{gray}{75.6} & \textcolor{gray}{65.6} & \textcolor{gray}{94.1} & \textcolor{gray}{46.7} & \textcolor{gray}{91.8} & \textcolor{gray}{39.7} & \textcolor{gray}{99.4} & \textcolor{gray}{73.3} \\
\quad OPD$^2$   & 73.8 & 66.5 & \underline{97.8} & 47.9 & \textbf{91.8} & 47.6 & \textbf{99.7} & 75.0 \\
\quad + \ours ($\alpha$=1)   & 81.5 & \textbf{71.2} & \underline{97.8} & \textbf{52.9} & 90.6 & \underline{48.5} & \underline{99.3} & \textbf{77.4} \\
\rowcolor{ngreenlight}
\quad + merge ($\alpha$=0 $\oplus$ 1)   & \textbf{82.1} & 70.0 & \textbf{98.4} & \underline{51.7} & \underline{91.6} & 48.0 & \textbf{99.7} & \textbf{77.4} \\
\bottomrule
\end{tabular}
}
\label{tab:qwen_think8b_merge_math_opd2}
\end{table}

\begin{table}[!htbp]
\centering
\caption{\textbf{Model merging with OPD$^2$ on Qwen3-8B: code and science benchmarks (thinking mode).} \textit{merge} averages the weights of the OPD$^2$ model ($\alpha$=0) and the \ours ($\alpha$=1) model with equal weights. The best and second-best results among the OPD$^2$ variants are shown in bold and underlined, respectively.}
\label{tab:think8b_merge_other_opd2}
\small
\setlength{\tabcolsep}{7.8pt}
\begin{tabular}{lcccc|cccc}
\toprule
& \multicolumn{4}{c|}{\textbf{Code}} & \multicolumn{4}{c}{\textbf{Science}} \\
\cmidrule(lr){2-5} \cmidrule(lr){6-9}
\textbf{Model}
& \begin{tabular}[c]{@{}c@{}}Code\\Forces\end{tabular}
& LCBv5
& \begin{tabular}[c]{@{}c@{}}RG\\Algo\end{tabular}
& \textbf{Avg}
& GPQA
& \begin{tabular}[c]{@{}c@{}}Super\\GPQA\end{tabular}
& \begin{tabular}[c]{@{}c@{}}Sci\\Bench\end{tabular}
& \textbf{Avg} \\
\midrule
Qwen3-8B   & \textcolor{gray}{40.1} & \textcolor{gray}{50.9} & \textcolor{gray}{67.1} & \textcolor{gray}{52.7}
               & \textcolor{gray}{63.2} & \textcolor{gray}{39.5} & \textcolor{gray}{58.0} & \textcolor{gray}{53.6} \\
\quad OPD$^2$   & 46.5 & \underline{65.5} & \underline{72.1} & 61.4
               & 62.3 & 41.3 & 53.8 & 52.5 \\
\quad + \ours ($\alpha$=1)   & 46.7 & 62.8 & 71.4 & 60.3
               & \textbf{64.2} & \underline{42.1} & \textbf{60.6} & \textbf{55.6} \\
\rowcolor{ngreenlight}
\quad + merge ($\alpha$=0 $\oplus$ 1)   & \underline{47.8} & 65.0 & \textbf{74.0} & \textbf{62.3}
               & 63.4 & 41.7 & 57.4 & 54.2 \\
\bottomrule
\end{tabular}
\label{tab:qwen_think8b_merge_other_opd2}
\end{table}

%% file: figs/heldin_full.tex
\definecolor{ink}{HTML}{1A1F24}  \definecolor{mut}{HTML}{8A929B}
\definecolor{mag}{HTML}{B03AA0}

\pgfplotsset{
  ptmark/.style={only marks, color=mut, mark=*, mark size=1.1pt,
                 mark options={solid, draw=mut, fill=mut, opacity=0.45}},
  idl/.style={mut, dashed, line width=1.0pt},
  zline/.style={mut, opacity=0.55, dotted, line width=0.8pt},
  chalo/.style={only marks, color=white, mark=*, mark size=3.4pt,
                mark options={draw=white, fill=white, line width=0.6pt}},
  cmark/.style={only marks, color=mag, mark=+, mark size=6.2pt,
                mark options={mag, line width=1.9pt}},
  cdot/.style={only marks, color=mag, mark=*, mark size=2.4pt,
               mark options={draw=white, line width=0.9pt, fill=mag}},
}

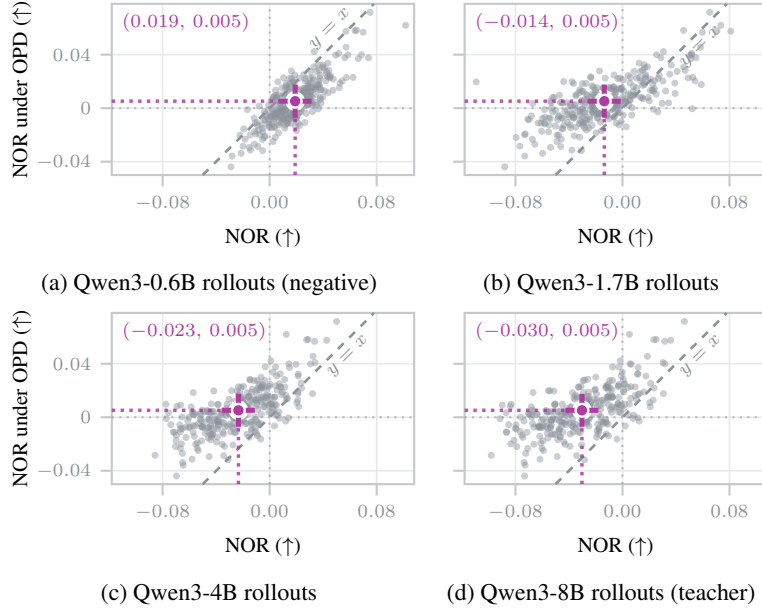
\begin{figure}[h]
\centering
\begin{subfigure}[b]{0.40\textwidth}
\centering
\begin{tikzpicture}
\begin{axis}[ldacc,
  scale only axis=true, width=4.0cm, height=2.265cm,
  xmin=-0.118, xmax=0.108, ymin=-0.05, ymax=0.078,
  xtick={-0.08,0,0.08},
  ytick={-0.04,0,0.04},
  scaled ticks=false,
  xticklabel style={/pgf/number format/fixed, /pgf/number format/precision=2},
  yticklabel style={/pgf/number format/fixed, /pgf/number format/precision=2},
  enlargelimits=false,
  xlabel={NOR ($\uparrow$)},
  ylabel={NOR under OPD ($\uparrow$)},
  ylabel style={yshift=-5pt},
  clip mode=individual,
  label style={font=\footnotesize},
  ticklabel style={font=\scriptsize}
]
\addplot[zline, forget plot] coordinates {(-0.118,0) (0.108,0)};
\addplot[zline, forget plot] coordinates {(0,-0.05) (0,0.078)};
\addplot[idl, forget plot, domain=-0.05:0.078, samples=2] {x}
        node[pos=0.8, sloped, above, mut, font=\scriptsize, inner sep=2pt] {$y=x$};
\addplot[ptmark, forget plot] coordinates {
  (0.01627,-0.00191) (0.00307,-0.00990) (0.01258,0.01660) (-0.02224,-0.02024) (0.00546,0.00816)
  (0.01345,-0.00734) (0.01743,0.00411) (0.02863,0.01767) (0.03785,0.00772) (0.01863,-0.00345)
  (0.01184,-0.00125) (0.01642,0.00140) (0.01782,-0.00288) (-0.01166,-0.00741) (0.05727,0.02291)
  (0.06379,0.05794) (0.02998,0.02262) (0.00348,0.00912) (0.02640,0.02208) (-0.00326,-0.00460)
  (0.03326,0.04230) (0.00943,-0.00836) (0.00832,-0.00557) (0.00851,0.00015) (0.01943,-0.01343)
  (0.01431,-0.00468) (0.06490,0.05821) (0.03097,0.00841) (0.04343,0.02351) (0.01787,0.00377)
  (0.01868,0.02484) (0.05262,0.01876) (-0.01150,-0.02700) (-0.00063,0.00235) (0.03615,0.01768)
  (0.01290,0.00869) (0.03392,0.01576) (0.03256,0.02102) (0.00257,-0.02489) (0.02859,0.00906)
  (-0.00899,-0.01287) (0.02002,0.01506) (0.02511,0.00782) (0.00640,-0.00477) (0.01165,-0.01257)
  (0.00498,0.00953) (0.02137,-0.01121) (0.00490,0.00342) (0.03076,0.00538) (0.03106,0.00419)
  (0.02797,0.00263) (0.00435,-0.00476) (0.02220,-0.00352) (0.00967,-0.00253) (0.02630,0.00552)
  (0.02610,0.01182) (0.00668,-0.01851) (0.03774,0.00644) (0.01785,-0.00297) (0.01465,0.01465)
  (0.02365,0.00895) (0.01011,-0.00334) (0.04252,0.02745) (0.03134,0.01893) (0.03070,0.01260)
  (0.02524,0.01048) (0.00704,-0.01042) (0.05844,0.03990) (0.00404,-0.00326) (0.03442,0.01471)
  (0.03981,0.00825) (0.03526,0.02742) (0.02179,0.02227) (0.05711,0.04560) (0.03956,0.03081)
  (0.04107,0.01284) (-0.00120,-0.00114) (-0.00519,-0.01763) (0.03719,0.02839) (0.01176,-0.00147)
  (-0.02808,-0.03580) (0.00299,0.00771) (0.05415,0.02032) (-0.00386,-0.02191) (0.02245,0.00215)
  (0.05705,0.05753) (0.02371,0.00975) (0.00333,-0.01387) (0.00616,-0.00347) (0.03213,0.02301)
  (0.01779,0.00592) (0.01384,0.00009) (0.01645,-0.00670) (0.01947,0.00395) (0.00411,-0.02175)
  (0.02588,0.01480) (-0.01431,-0.03138) (0.02167,0.00292) (-0.00536,-0.01507) (0.04513,0.02263)
  (0.02870,0.01504) (-0.00094,-0.01098) (0.00288,-0.00074) (0.00649,-0.00340) (0.02352,0.00039)
  (0.00151,-0.00653) (0.03692,0.01757) (0.01649,0.02210) (0.00770,-0.00143) (-0.00831,-0.01629)
  (0.03776,0.01354) (0.10166,0.06194) (0.01548,0.00630) (0.03304,0.03092) (0.01627,-0.00504)
  (0.01599,-0.01116) (0.00103,0.00539) (0.00771,0.00188) (0.01327,0.00940) (0.01600,0.00263)
  (0.01252,0.01848) (0.04733,0.01334) (-0.01987,-0.02841) (0.00385,-0.00772) (0.02573,0.00662)
  (0.00689,-0.01253) (0.00962,-0.00239) (0.01820,0.00190) (-0.00778,0.00157) (0.01348,0.00421)
  (-0.00240,-0.00639) (0.01426,0.00515) (0.02081,0.00012) (0.02487,0.01460) (0.01125,-0.01494)
  (0.01174,-0.00555) (0.03462,0.01043) (0.01949,-0.00754) (0.01776,0.01990) (-0.00079,-0.01266)
  (0.02706,0.00702) (0.05202,0.01531) (0.04490,0.02350) (0.04089,0.02998) (-0.00516,-0.03179)
  (0.03004,0.01077) (0.02466,0.01070) (0.03745,0.01572) (0.02239,0.00998) (0.06990,0.05676)
  (0.03761,0.02446) (0.02963,0.01833) (0.02289,0.00978) (-0.01528,-0.02937) (0.02540,0.00793)
  (0.04193,0.02049) (0.02066,0.00358) (0.01554,0.00426) (0.03200,0.00313) (0.00075,-0.00573)
  (0.00396,-0.01457) (0.00417,-0.00680) (0.00838,-0.00044) (0.00936,-0.00795) (0.01750,-0.00712)
  (-0.00052,-0.00874) (0.02255,-0.01289) (0.01820,-0.00594) (0.01853,0.00345) (0.05614,0.03764)
  (0.03696,0.02035) (0.02641,-0.01073) (0.02735,-0.00386) (0.00219,-0.00160) (0.01370,-0.00962)
  (0.00853,-0.01123) (-0.02137,-0.03379) (0.01040,-0.00763) (0.01634,0.00157) (0.04371,0.01152)
  (0.02175,0.00982) (-0.00443,-0.00875) (0.04356,0.03369) (0.01479,0.01606) (0.01494,0.02247)
  (-0.00392,-0.00268) (0.03625,0.03243) (0.05158,0.04086) (0.01834,0.01887) (0.00676,0.01002)
  (-0.00038,-0.01429) (0.00162,-0.01184) (0.00812,0.00414) (-0.00529,-0.02329) (0.00511,-0.00435)
  (0.00067,-0.00624) (0.02839,0.00255) (0.01085,-0.00823) (0.01033,0.01032) (0.02085,-0.00519)
  (0.00804,0.00729) (-0.02888,-0.04388) (0.03057,0.02750) (0.02223,0.01349) (-0.00335,-0.00867)
  (0.04530,0.02454) (0.00502,0.00558) (0.03083,0.01272) (-0.01937,-0.01555) (0.03625,-0.00055)
  (0.03455,0.00427) (0.01857,-0.00689) (0.01911,0.00309) (0.01718,0.00266) (0.01040,0.02275)
  (0.07507,0.07163) (0.06865,0.03995) (0.04928,0.03963) (0.02669,0.02028) (0.03863,0.01382)
  (0.04755,0.02177) (0.01247,-0.00519) (0.01701,0.01963) (0.00265,0.00804) (0.02759,0.01475)
  (0.03017,0.00901) (0.02613,0.00395) (0.01287,-0.01123) (0.01170,0.00222) (0.01740,-0.00751)
  (0.03512,0.00221) (0.00617,0.01972) (0.01190,-0.00047) (0.00618,-0.00275) (-0.01076,-0.01107)
  (0.06236,0.02303) (0.01538,-0.00354) (-0.00947,-0.01078) (-0.00351,-0.02069) (-0.01053,-0.01921)
  (0.00451,0.01347) (0.02212,0.00981) (0.07385,0.03766) (0.02387,-0.00080) (0.02325,-0.00271)
  (-0.00583,-0.01257) (-0.01472,-0.02016) (0.02446,0.00259) (0.00978,0.00077) (-0.01659,-0.00106)
  (0.02009,-0.00759) (-0.01835,-0.02323) (0.02841,0.01274) (0.03978,0.03227) (0.01328,0.01379)
  (0.01196,0.01520) (0.04238,0.02535) (0.03197,0.00234) (0.02883,0.01625) (0.02062,0.00444)
  (0.01885,0.01518) (0.02013,0.00627) (0.06625,0.03710) (0.00145,-0.01468) (0.01319,0.00482)
  (0.01378,-0.01151) (0.03267,0.01411) (0.03196,-0.00259) (-0.01117,-0.01730) (0.01986,-0.00653)
  (0.02546,0.01273) (0.01350,-0.00817) (0.01894,0.00373) (0.00772,-0.00670) (0.00338,-0.00442)
  (-0.00746,-0.01988) (0.00941,-0.00149) (0.01750,0.00796) (0.01445,0.00700) (0.01677,0.01201)
  (-0.00869,-0.02869) (0.02005,-0.01676) (0.03188,0.01547) (0.01332,0.00567) (0.03833,0.00356)
  (0.03739,0.02247) (0.02852,0.00247) (0.01963,0.00459) (0.02783,0.01093) (0.02475,0.01285)
  (0.03168,0.00801) (0.02100,-0.00043) (0.00296,-0.00439) (0.02546,0.02504) (0.01503,0.01837)
  (0.00494,-0.00651) (0.03559,0.01712) (0.00918,0.01526) (0.04420,0.02824) (0.01626,0.02782)
};
\addplot[zline, mag, opacity=0.85, line width=1.3pt, forget plot]
        coordinates {(-0.118,0.00521) (0.01905,0.00521)};
\addplot[zline, mag, opacity=0.85, line width=1.3pt, forget plot]
        coordinates {(0.01905,-0.05) (0.01905,0.00521)};
\addplot[chalo, forget plot] coordinates {(0.01905,0.00521)};
\addplot[cmark, forget plot] coordinates {(0.01905,0.00521)};
\addplot[cdot, forget plot]  coordinates {(0.01905,0.00521)};
\node[font=\scriptsize, anchor=north west, inner sep=2pt] at (axis cs:-0.1145,0.0755)
     {\textcolor{mag}{$(0.019,\,0.005)$}};
\end{axis}
\end{tikzpicture}
\subcaption{Qwen3-0.6B rollouts (\name)}\label{fig:away-lad-06b}
\end{subfigure}
\begin{subfigure}[b]{0.33\textwidth}
\centering
\begin{tikzpicture}
\begin{axis}[ldacc,
  scale only axis=true, width=4.0cm, height=2.265cm,
  xmin=-0.118, xmax=0.108, ymin=-0.05, ymax=0.078,
  xtick={-0.08,0,0.08},
  ytick={-0.04,0,0.04},
  scaled ticks=false,
  xticklabel style={/pgf/number format/fixed, /pgf/number format/precision=2},
  yticklabel style={/pgf/number format/fixed, /pgf/number format/precision=2},
  enlargelimits=false,
  xlabel={NOR ($\uparrow$)},
  yticklabels={},
  clip mode=individual,
  label style={font=\footnotesize},
  ticklabel style={font=\scriptsize}
]
\addplot[zline, forget plot] coordinates {(-0.118,0) (0.108,0)};
\addplot[zline, forget plot] coordinates {(0,-0.05) (0,0.078)};
\addplot[idl, forget plot, domain=-0.05:0.078, samples=2] {x}
        node[pos=0.8, sloped, below, mut, font=\scriptsize, inner sep=2pt] {$y=x$};
\addplot[ptmark, forget plot] coordinates {
  (-0.03172,-0.00191) (-0.00087,-0.00990) (-0.02270,0.01660) (-0.04898,-0.02024) (-0.05212,0.00816)
  (-0.05514,-0.00734) (-0.05315,0.00411) (0.02262,0.01767) (0.01450,0.00772) (-0.02808,-0.00345)
  (-0.02397,-0.00125) (-0.02287,0.00140) (0.02642,-0.00288) (-0.04554,-0.00741) (0.01269,0.02291)
  (0.07057,0.05794) (0.01061,0.02262) (-0.03475,0.00912) (0.01707,0.02208) (-0.03706,-0.00460)
  (0.03062,0.04230) (-0.02428,-0.00836) (-0.04722,-0.00557) (-0.05315,0.00015) (-0.05008,-0.01343)
  (-0.01876,-0.00468) (0.04570,0.05821) (-0.01076,0.00841) (0.01759,0.02351) (-0.00438,0.00377)
  (-0.00632,0.02484) (0.03913,0.01876) (-0.06552,-0.02700) (0.00468,0.00235) (0.05894,0.01768)
  (-0.01594,0.00869) (-0.00706,0.01576) (0.00062,0.02102) (-0.04896,-0.02489) (-0.00839,0.00906)
  (-0.02970,-0.01287) (-0.00031,0.01506) (-0.05107,0.00782) (-0.01157,-0.00477) (-0.03908,-0.01257)
  (-0.02163,0.00953) (-0.02716,-0.01121) (-0.06706,0.00342) (0.00183,0.00538) (-0.04544,0.00419)
  (-0.00584,0.00263) (-0.04085,-0.00476) (-0.01335,-0.00352) (-0.03313,-0.00253) (0.02002,0.00552)
  (-0.01967,0.01182) (-0.01960,-0.01851) (-0.00338,0.00644) (-0.02854,-0.00297) (0.02384,0.01465)
  (0.00671,0.00895) (-0.02199,-0.00334) (0.01326,0.02745) (0.01553,0.01893) (-0.02955,0.01260)
  (-0.02729,0.01048) (-0.01653,-0.01042) (0.05169,0.03990) (-0.01559,-0.00326) (-0.00096,0.01471)
  (0.01367,0.00825) (0.00862,0.02742) (0.02662,0.02227) (0.05838,0.04560) (0.04570,0.03081)
  (0.02318,0.01284) (-0.02068,-0.00114) (-0.05524,-0.01763) (0.01505,0.02839) (-0.04126,-0.00147)
  (-0.07304,-0.03580) (-0.02073,0.00771) (0.05422,0.02032) (-0.07956,-0.02191) (-0.02267,0.00215)
  (0.08094,0.05753) (-0.04711,0.00975) (0.00279,-0.01387) (-0.01823,-0.00347) (0.00595,0.02301)
  (0.00571,0.00592) (-0.03672,0.00009) (-0.00649,-0.00670) (-0.04141,0.00395) (-0.03204,-0.02175)
  (-0.00263,0.01480) (-0.03782,-0.03138) (-0.02210,0.00292) (-0.04532,-0.01507) (0.04911,0.02263)
  (-0.00941,0.01504) (-0.05832,-0.01098) (-0.03103,-0.00074) (-0.00232,-0.00340) (0.00720,0.00039)
  (-0.00604,-0.00653) (0.01649,0.01757) (-0.00348,0.02210) (-0.00917,-0.00143) (-0.02696,-0.01629)
  (-0.02029,0.01354) (0.07346,0.06194) (0.00542,0.00630) (0.01873,0.03092) (-0.01211,-0.00504)
  (-0.00469,-0.01116) (-0.05399,0.00539) (-0.03826,0.00188) (-0.02616,0.00940) (-0.03527,0.00263)
  (-0.02721,0.01848) (0.02083,0.01334) (-0.07677,-0.02841) (-0.05979,-0.00772) (-0.02660,0.00662)
  (-0.00248,-0.01253) (-0.04313,-0.00239) (-0.03591,0.00190) (-0.00408,0.00157) (-0.01865,0.00421)
  (-0.04792,-0.00639) (-0.01845,0.00515) (-0.00730,0.00012) (-0.00841,0.01460) (-0.02508,-0.01494)
  (-0.03197,-0.00555) (0.02521,0.01043) (-0.04446,-0.00754) (-0.04394,0.01990) (-0.04370,-0.01266)
  (0.00181,0.00702) (-0.01344,0.01531) (-0.02379,0.02350) (0.01661,0.02998) (-0.03060,-0.03179)
  (0.00440,0.01077) (-0.02248,0.01070) (-0.02171,0.01572) (0.00580,0.00998) (0.05967,0.05676)
  (0.00426,0.02446) (-0.02998,0.01833) (-0.01836,0.00978) (-0.05375,-0.02937) (0.01921,0.00793)
  (0.02393,0.02049) (-0.00158,0.00358) (-0.04637,0.00426) (-0.01417,0.00313) (-0.06809,-0.00573)
  (-0.03886,-0.01457) (-0.03631,-0.00680) (-0.02133,-0.00044) (-0.01004,-0.00795) (-0.01872,-0.00712)
  (-0.06969,-0.00874) (-0.02265,-0.01289) (-0.01116,-0.00594) (-0.01019,0.00345) (0.05501,0.03764)
  (0.01380,0.02035) (-0.04102,-0.01073) (-0.01654,-0.00386) (-0.05001,-0.00160) (-0.03179,-0.00962)
  (-0.02894,-0.01123) (-0.05971,-0.03379) (-0.05394,-0.00763) (-0.00156,0.00157) (-0.03265,0.01152)
  (0.00954,0.00982) (-0.02670,-0.00875) (0.01239,0.03369) (-0.00646,0.01606) (-0.01285,0.02247)
  (-0.02249,-0.00268) (0.01624,0.03243) (0.05169,0.04086) (0.00556,0.01887) (-0.03484,0.01002)
  (-0.03499,-0.01429) (-0.05789,-0.01184) (-0.02242,0.00414) (-0.06733,-0.02329) (-0.03792,-0.00435)
  (-0.02965,-0.00624) (-0.00498,0.00255) (-0.07085,-0.00823) (-0.00553,0.01032) (0.00694,-0.00519)
  (-0.01569,0.00729) (-0.08811,-0.04388) (0.04089,0.02750) (-0.00655,0.01349) (-0.04843,-0.00867)
  (0.02751,0.02454) (-0.03063,0.00558) (0.00808,0.01272) (-0.04454,-0.01555) (-0.00786,-0.00055)
  (-0.02756,0.00427) (-0.03660,-0.00689) (-0.01420,0.00309) (-0.03905,0.00266) (-0.02359,0.02275)
  (0.06516,0.07163) (0.02550,0.03995) (0.05065,0.03963) (0.03026,0.02028) (-0.00471,0.01382)
  (0.02091,0.02177) (0.00064,-0.00519) (-0.10936,0.01963) (-0.02314,0.00804) (-0.01405,0.01475)
  (-0.01112,0.00901) (0.03448,0.00395) (-0.06078,-0.01123) (-0.02048,0.00222) (-0.03650,-0.00751)
  (-0.02799,0.00221) (-0.03764,0.01972) (-0.02984,-0.00047) (-0.05215,-0.00275) (-0.06306,-0.01107)
  (0.05254,0.02303) (-0.00331,-0.00354) (-0.03073,-0.01078) (-0.03598,-0.02069) (-0.07265,-0.01921)
  (0.02232,0.01347) (0.00859,0.00981) (0.02872,0.03766) (0.02317,-0.00080) (-0.00495,-0.00271)
  (-0.05478,-0.01257) (-0.04771,-0.02016) (-0.01178,0.00259) (-0.02278,0.00077) (-0.05825,-0.00106)
  (-0.01838,-0.00759) (-0.04572,-0.02323) (-0.03044,0.01274) (0.03644,0.03227) (-0.03185,0.01379)
  (-0.01472,0.01520) (0.03535,0.02535) (-0.00339,0.00234) (0.00828,0.01625) (0.00164,0.00444)
  (0.01054,0.01518) (-0.02840,0.00627) (0.05128,0.03710) (-0.02842,-0.01468) (-0.03092,0.00482)
  (-0.03219,-0.01151) (-0.00115,0.01411) (-0.01758,-0.00259) (-0.06043,-0.01730) (-0.02475,-0.00653)
  (0.00089,0.01273) (-0.02821,-0.00817) (-0.04456,0.00373) (-0.05811,-0.00670) (-0.01718,-0.00442)
  (-0.07317,-0.01988) (-0.00354,-0.00149) (-0.00568,0.00796) (0.01544,0.00700) (-0.00504,0.01201)
  (-0.04613,-0.02869) (-0.01566,-0.01676) (-0.01011,0.01547) (-0.00709,0.00567) (0.02968,0.00356)
  (-0.00184,0.02247) (-0.01961,0.00247) (0.00295,0.00459) (0.03938,0.01093) (0.03230,0.01285)
  (0.00009,0.00801) (0.05214,-0.00043) (-0.02107,-0.00439) (0.01731,0.02504) (-0.03660,0.01837)
  (-0.03982,-0.00651) (0.01504,0.01712) (-0.01888,0.01526) (0.04695,0.02824) (-0.01717,0.02782)
};
\addplot[zline, mag, opacity=0.85, line width=1.3pt, forget plot]
        coordinates {(-0.118,0.00521) (-0.01359,0.00521)};
\addplot[zline, mag, opacity=0.85, line width=1.3pt, forget plot]
        coordinates {(-0.01359,-0.05) (-0.01359,0.00521)};
\addplot[chalo, forget plot] coordinates {(-0.01359,0.00521)};
\addplot[cmark, forget plot] coordinates {(-0.01359,0.00521)};
\addplot[cdot, forget plot]  coordinates {(-0.01359,0.00521)};
\node[font=\scriptsize, anchor=north west, inner sep=2pt] at (axis cs:-0.1145,0.0755)
     {\textcolor{mag}{$(-0.014,\,0.005)$}};
\end{axis}
\end{tikzpicture}
\subcaption{Qwen3-1.7B rollouts}\label{fig:away-lad-17b}
\end{subfigure}

\begin{subfigure}[b]{0.40\textwidth}
\centering
\begin{tikzpicture}
\begin{axis}[ldacc,
  scale only axis=true, width=4.0cm, height=2.265cm,
  xmin=-0.118, xmax=0.108, ymin=-0.05, ymax=0.078,
  xtick={-0.08,0,0.08},
  ytick={-0.04,0,0.04},
  scaled ticks=false,
  xticklabel style={/pgf/number format/fixed, /pgf/number format/precision=2},
  yticklabel style={/pgf/number format/fixed, /pgf/number format/precision=2},
  enlargelimits=false,
  xlabel={NOR ($\uparrow$)},
  ylabel={NOR under OPD ($\uparrow$)},
  ylabel style={yshift=-5pt},
  clip mode=individual,
  label style={font=\footnotesize},
  ticklabel style={font=\scriptsize}
]
\addplot[zline, forget plot] coordinates {(-0.118,0) (0.108,0)};
\addplot[zline, forget plot] coordinates {(0,-0.05) (0,0.078)};
\addplot[idl, forget plot, domain=-0.05:0.078, samples=2] {x}
        node[pos=0.8, sloped, below, mut, font=\scriptsize, inner sep=2pt] {$y=x$};
\addplot[ptmark, forget plot] coordinates {
  (-0.01973,-0.00191) (-0.01360,-0.00990) (-0.02321,0.01660) (-0.03551,-0.02024) (-0.06935,0.00816)
  (-0.05952,-0.00734) (-0.04840,0.00411) (0.02124,0.01767) (-0.00457,0.00772) (-0.04888,-0.00345)
  (-0.05191,-0.00125) (-0.03227,0.00140) (0.01198,-0.00288) (-0.06272,-0.00741) (0.00756,0.02291)
  (0.02761,0.05794) (-0.00321,0.02262) (-0.06661,0.00912) (-0.00635,0.02208) (-0.04147,-0.00460)
  (0.00095,0.04230) (-0.04370,-0.00836) (-0.07390,-0.00557) (-0.06545,0.00015) (-0.06973,-0.01343)
  (-0.01801,-0.00468) (0.02873,0.05821) (-0.01410,0.00841) (0.00436,0.02351) (-0.00074,0.00377)
  (-0.00790,0.02484) (-0.01347,0.01876) (-0.06513,-0.02700) (-0.00619,0.00235) (0.03202,0.01768)
  (-0.01857,0.00869) (-0.01529,0.01576) (-0.01296,0.02102) (-0.05336,-0.02489) (0.00506,0.00906)
  (-0.04414,-0.01287) (-0.01611,0.01506) (-0.05870,0.00782) (-0.01735,-0.00477) (-0.03444,-0.01257)
  (-0.03191,0.00953) (-0.05718,-0.01121) (-0.07562,0.00342) (-0.01996,0.00538) (-0.05772,0.00419)
  (-0.01880,0.00263) (-0.05328,-0.00476) (-0.00945,-0.00352) (-0.04471,-0.00253) (-0.01865,0.00552)
  (-0.02663,0.01182) (-0.03977,-0.01851) (-0.01679,0.00644) (-0.03906,-0.00297) (-0.00235,0.01465)
  (-0.00074,0.00895) (-0.03596,-0.00334) (-0.00366,0.02745) (0.02105,0.01893) (-0.01902,0.01260)
  (-0.01953,0.01048) (-0.01072,-0.01042) (0.03053,0.03990) (-0.02252,-0.00326) (-0.00619,0.01471)
  (-0.01753,0.00825) (-0.01172,0.02742) (0.00841,0.02227) (0.03240,0.04560) (0.02130,0.03081)
  (-0.01387,0.01284) (-0.02418,-0.00114) (-0.05091,-0.01763) (-0.01482,0.02839) (-0.03943,-0.00147)
  (-0.06461,-0.03580) (-0.02617,0.00771) (-0.00981,0.02032) (-0.07037,-0.02191) (-0.03598,0.00215)
  (0.04258,0.05753) (-0.07758,0.00975) (-0.00915,-0.01387) (-0.03843,-0.00347) (-0.01955,0.02301)
  (-0.00887,0.00592) (-0.06027,0.00009) (-0.02840,-0.00670) (-0.04467,0.00395) (-0.03376,-0.02175)
  (-0.02287,0.01480) (-0.05800,-0.03138) (-0.02764,0.00292) (-0.06253,-0.01507) (0.01137,0.02263)
  (-0.02050,0.01504) (-0.02906,-0.01098) (-0.02660,-0.00074) (-0.01303,-0.00340) (-0.01092,0.00039)
  (0.00361,-0.00653) (-0.00972,0.01757) (-0.00872,0.02210) (-0.02425,-0.00143) (-0.00666,-0.01629)
  (-0.01227,0.01354) (0.01243,0.06194) (0.00328,0.00630) (0.00520,0.03092) (-0.02964,-0.00504)
  (-0.02509,-0.01116) (-0.05045,0.00539) (-0.05680,0.00188) (-0.03885,0.00940) (-0.05225,0.00263)
  (-0.05711,0.01848) (0.01256,0.01334) (-0.08567,-0.02841) (-0.04592,-0.00772) (-0.03209,0.00662)
  (-0.00940,-0.01253) (-0.05127,-0.00239) (-0.05130,0.00190) (0.00461,0.00157) (-0.04297,0.00421)
  (-0.05560,-0.00639) (-0.04658,0.00515) (-0.02651,0.00012) (-0.02293,0.01460) (-0.01544,-0.01494)
  (-0.03832,-0.00555) (-0.00981,0.01043) (-0.06042,-0.00754) (-0.02842,0.01990) (-0.01698,-0.01266)
  (0.00224,0.00702) (0.00392,0.01531) (-0.00560,0.02350) (-0.00635,0.02998) (-0.03116,-0.03179)
  (-0.00278,0.01077) (-0.01538,0.01070) (-0.00474,0.01572) (0.00087,0.00998) (0.03360,0.05676)
  (-0.00532,0.02446) (-0.01678,0.01833) (-0.03846,0.00978) (-0.04522,-0.02937) (0.02175,0.00793)
  (-0.00742,0.02049) (-0.03552,0.00358) (-0.04668,0.00426) (-0.03055,0.00313) (-0.06715,-0.00573)
  (-0.04343,-0.01457) (-0.05009,-0.00680) (-0.02703,-0.00044) (-0.01064,-0.00795) (-0.03335,-0.00712)
  (-0.06679,-0.00874) (-0.04249,-0.01289) (-0.04406,-0.00594) (-0.02649,0.00345) (0.00439,0.03764)
  (-0.00243,0.02035) (-0.06665,-0.01073) (-0.02168,-0.00386) (-0.04561,-0.00160) (-0.04197,-0.00962)
  (-0.03511,-0.01123) (-0.06506,-0.03379) (-0.04558,-0.00763) (0.00581,0.00157) (-0.00570,0.01152)
  (0.00393,0.00982) (-0.01351,-0.00875) (0.00258,0.03369) (-0.00195,0.01606) (-0.00621,0.02247)
  (-0.04041,-0.00268) (0.00787,0.03243) (0.03392,0.04086) (0.00956,0.01887) (-0.04364,0.01002)
  (-0.05635,-0.01429) (-0.04815,-0.01184) (-0.02946,0.00414) (-0.06674,-0.02329) (-0.03428,-0.00435)
  (-0.05730,-0.00624) (-0.02990,0.00255) (-0.06535,-0.00823) (0.00069,0.01032) (-0.01318,-0.00519)
  (-0.02354,0.00729) (-0.06962,-0.04388) (-0.00922,0.02750) (-0.03041,0.01349) (-0.03904,-0.00867)
  (0.00025,0.02454) (-0.02477,0.00558) (-0.00681,0.01272) (-0.04601,-0.01555) (-0.02443,-0.00055)
  (-0.04186,0.00427) (-0.05119,-0.00689) (-0.03185,0.00309) (-0.05587,0.00266) (-0.02324,0.02275)
  (0.05000,0.07163) (0.01432,0.03995) (0.01967,0.03963) (0.00575,0.02028) (-0.02149,0.01382)
  (0.02393,0.02177) (-0.00807,-0.00519) (-0.02440,0.01963) (-0.01718,0.00804) (0.00341,0.01475)
  (-0.00934,0.00901) (0.00700,0.00395) (-0.07111,-0.01123) (-0.02733,0.00222) (-0.05042,-0.00751)
  (-0.03661,0.00221) (-0.04175,0.01972) (-0.04493,-0.00047) (-0.06122,-0.00275) (-0.07215,-0.01107)
  (-0.00054,0.02303) (-0.01404,-0.00354) (-0.04623,-0.01078) (-0.02744,-0.02069) (-0.05397,-0.01921)
  (-0.00875,0.01347) (0.00182,0.00981) (-0.00331,0.03766) (0.00734,-0.00080) (-0.01966,-0.00271)
  (-0.04297,-0.01257) (-0.05527,-0.02016) (-0.02987,0.00259) (-0.04076,0.00077) (-0.07727,-0.00106)
  (-0.04976,-0.00759) (-0.05427,-0.02323) (-0.01053,0.01274) (0.02132,0.03227) (-0.01971,0.01379)
  (-0.00306,0.01520) (0.02344,0.02535) (-0.01598,0.00234) (0.01223,0.01625) (-0.01999,0.00444)
  (0.00149,0.01518) (-0.01597,0.00627) (0.01482,0.03710) (-0.03906,-0.01468) (-0.03459,0.00482)
  (-0.02618,-0.01151) (0.01339,0.01411) (-0.03794,-0.00259) (-0.06998,-0.01730) (-0.02442,-0.00653)
  (-0.01057,0.01273) (-0.02729,-0.00817) (-0.04419,0.00373) (-0.03822,-0.00670) (-0.02708,-0.00442)
  (-0.06593,-0.01988) (-0.01025,-0.00149) (-0.02844,0.00796) (-0.01764,0.00700) (-0.02256,0.01201)
  (-0.03430,-0.02869) (-0.04407,-0.01676) (-0.01573,0.01547) (-0.02417,0.00567) (0.00108,0.00356)
  (-0.02140,0.02247) (-0.05755,0.00247) (-0.02128,0.00459) (0.02032,0.01093) (0.02106,0.01285)
  (-0.00362,0.00801) (0.01403,-0.00043) (-0.00727,-0.00439) (0.00733,0.02504) (-0.02234,0.01837)
  (-0.03156,-0.00651) (0.00904,0.01712) (-0.00113,0.01526) (0.01124,0.02824) (-0.01462,0.02782)
};
\addplot[zline, mag, opacity=0.85, line width=1.3pt, forget plot]
        coordinates {(-0.118,0.00521) (-0.02334,0.00521)};
\addplot[zline, mag, opacity=0.85, line width=1.3pt, forget plot]
        coordinates {(-0.02334,-0.05) (-0.02334,0.00521)};
\addplot[chalo, forget plot] coordinates {(-0.02334,0.00521)};
\addplot[cmark, forget plot] coordinates {(-0.02334,0.00521)};
\addplot[cdot, forget plot]  coordinates {(-0.02334,0.00521)};
\node[font=\scriptsize, anchor=north west, inner sep=2pt] at (axis cs:-0.1145,0.0755)
     {\textcolor{mag}{$(-0.023,\,0.005)$}};
\end{axis}
\end{tikzpicture}
\subcaption{Qwen3-4B rollouts}\label{fig:away-lad-4b}
\end{subfigure}
\begin{subfigure}[b]{0.33\textwidth}
\centering
\begin{tikzpicture}
\begin{axis}[ldacc,
  scale only axis=true, width=4.0cm, height=2.265cm,
  xmin=-0.118, xmax=0.108, ymin=-0.05, ymax=0.078,
  xtick={-0.08,0,0.08},
  ytick={-0.04,0,0.04},
  scaled ticks=false,
  xticklabel style={/pgf/number format/fixed, /pgf/number format/precision=2},
  yticklabel style={/pgf/number format/fixed, /pgf/number format/precision=2},
  enlargelimits=false,
  xlabel={NOR ($\uparrow$)},
  yticklabels={},
  clip mode=individual,
  label style={font=\footnotesize},
  ticklabel style={font=\scriptsize}
]
\addplot[zline, forget plot] coordinates {(-0.118,0) (0.108,0)};
\addplot[zline, forget plot] coordinates {(0,-0.05) (0,0.078)};
\addplot[idl, forget plot, domain=-0.05:0.078, samples=2] {x}
        node[pos=0.8, sloped, below, mut, font=\scriptsize, inner sep=2pt] {$y=x$};
\addplot[ptmark, forget plot] coordinates {
  (-0.02854,-0.00191) (-0.01889,-0.00990) (-0.03506,0.01660) (-0.04271,-0.02024) (-0.08385,0.00816)
  (-0.06615,-0.00734) (-0.05614,0.00411) (0.01283,0.01767) (-0.00417,0.00772) (-0.05674,-0.00345)
  (-0.05923,-0.00125) (-0.04211,0.00140) (0.01255,-0.00288) (-0.06304,-0.00741) (0.00626,0.02291)
  (0.02341,0.05794) (-0.01071,0.02262) (-0.08986,0.00912) (-0.01254,0.02208) (-0.04363,-0.00460)
  (-0.00169,0.04230) (-0.05573,-0.00836) (-0.08504,-0.00557) (-0.07599,0.00015) (-0.07515,-0.01343)
  (-0.02650,-0.00468) (0.02521,0.05821) (-0.01721,0.00841) (-0.00294,0.02351) (-0.00820,0.00377)
  (0.00116,0.02484) (-0.02543,0.01876) (-0.07229,-0.02700) (-0.01754,0.00235) (0.03061,0.01768)
  (-0.02446,0.00869) (-0.02640,0.01576) (-0.02031,0.02102) (-0.05905,-0.02489) (-0.00174,0.00906)
  (-0.05064,-0.01287) (-0.03394,0.01506) (-0.06816,0.00782) (-0.01439,-0.00477) (-0.03912,-0.01257)
  (-0.05056,0.00953) (-0.07133,-0.01121) (-0.08694,0.00342) (-0.02544,0.00538) (-0.06665,0.00419)
  (-0.02713,0.00263) (-0.06716,-0.00476) (-0.01248,-0.00352) (-0.05356,-0.00253) (-0.02570,0.00552)
  (-0.02679,0.01182) (-0.04375,-0.01851) (-0.02633,0.00644) (-0.04197,-0.00297) (-0.00757,0.01465)
  (-0.00070,0.00895) (-0.04600,-0.00334) (-0.00742,0.02745) (0.01614,0.01893) (-0.03008,0.01260)
  (-0.02399,0.01048) (-0.01760,-0.01042) (0.03175,0.03990) (-0.02665,-0.00326) (-0.01125,0.01471)
  (-0.02474,0.00825) (-0.02942,0.02742) (0.00417,0.02227) (0.01868,0.04560) (0.01750,0.03081)
  (-0.01919,0.01284) (-0.03063,-0.00114) (-0.05949,-0.01763) (-0.02595,0.02839) (-0.05011,-0.00147)
  (-0.07280,-0.03580) (-0.02866,0.00771) (-0.02082,0.02032) (-0.07902,-0.02191) (-0.04558,0.00215)
  (0.03600,0.05753) (-0.08317,0.00975) (-0.01045,-0.01387) (-0.04523,-0.00347) (-0.03096,0.02301)
  (-0.00885,0.00592) (-0.06776,0.00009) (-0.03218,-0.00670) (-0.05523,0.00395) (-0.03922,-0.02175)
  (-0.03070,0.01480) (-0.06162,-0.03138) (-0.03713,0.00292) (-0.07439,-0.01507) (0.00952,0.02263)
  (-0.02940,0.01504) (-0.03311,-0.01098) (-0.03713,-0.00074) (-0.01603,-0.00340) (-0.01623,0.00039)
  (0.00212,-0.00653) (-0.01939,0.01757) (-0.00934,0.02210) (-0.02490,-0.00143) (-0.00404,-0.01629)
  (-0.02094,0.01354) (0.00519,0.06194) (0.00075,0.00630) (-0.00076,0.03092) (-0.03484,-0.00504)
  (-0.03679,-0.01116) (-0.06725,0.00539) (-0.06866,0.00188) (-0.05079,0.00940) (-0.06374,0.00263)
  (-0.06680,0.01848) (0.01051,0.01334) (-0.09808,-0.02841) (-0.05389,-0.00772) (-0.04170,0.00662)
  (-0.01214,-0.01253) (-0.06039,-0.00239) (-0.06399,0.00190) (0.00834,0.00157) (-0.04822,0.00421)
  (-0.06251,-0.00639) (-0.05711,0.00515) (-0.03000,0.00012) (-0.03620,0.01460) (-0.01929,-0.01494)
  (-0.04232,-0.00555) (-0.01417,0.01043) (-0.06212,-0.00754) (-0.05431,0.01990) (-0.01838,-0.01266)
  (-0.00541,0.00702) (-0.00731,0.01531) (-0.01220,0.02350) (-0.01996,0.02998) (-0.02839,-0.03179)
  (-0.02014,0.01077) (-0.02980,0.01070) (-0.01044,0.01572) (-0.00348,0.00998) (0.03158,0.05676)
  (-0.01096,0.02446) (-0.02911,0.01833) (-0.04491,0.00978) (-0.04881,-0.02937) (0.02073,0.00793)
  (-0.01043,0.02049) (-0.04292,0.00358) (-0.05349,0.00426) (-0.03852,0.00313) (-0.07954,-0.00573)
  (-0.04668,-0.01457) (-0.05713,-0.00680) (-0.02696,-0.00044) (-0.01737,-0.00795) (-0.03694,-0.00712)
  (-0.07953,-0.00874) (-0.05324,-0.01289) (-0.05321,-0.00594) (-0.02789,0.00345) (-0.00260,0.03764)
  (-0.00788,0.02035) (-0.07638,-0.01073) (-0.02702,-0.00386) (-0.05304,-0.00160) (-0.04716,-0.00962)
  (-0.04447,-0.01123) (-0.07351,-0.03379) (-0.05373,-0.00763) (0.00436,0.00157) (-0.02389,0.01152)
  (0.00185,0.00982) (-0.01759,-0.00875) (-0.00013,0.03369) (-0.00826,0.01606) (-0.01197,0.02247)
  (-0.04705,-0.00268) (0.01129,0.03243) (0.03030,0.04086) (0.00081,0.01887) (-0.05023,0.01002)
  (-0.06182,-0.01429) (-0.05615,-0.01184) (-0.04173,0.00414) (-0.07075,-0.02329) (-0.03951,-0.00435)
  (-0.05963,-0.00624) (-0.03877,0.00255) (-0.08002,-0.00823) (-0.00314,0.01032) (-0.01853,-0.00519)
  (-0.03836,0.00729) (-0.07317,-0.04388) (-0.00952,0.02750) (-0.03550,0.01349) (-0.04739,-0.00867)
  (-0.00266,0.02454) (-0.03253,0.00558) (-0.01468,0.01272) (-0.05567,-0.01555) (-0.02496,-0.00055)
  (-0.05058,0.00427) (-0.06224,-0.00689) (-0.03733,0.00309) (-0.07016,0.00266) (-0.03254,0.02275)
  (0.04629,0.07163) (0.00572,0.03995) (0.00649,0.03963) (0.00090,0.02028) (-0.03023,0.01382)
  (0.02361,0.02177) (-0.01071,-0.00519) (-0.05122,0.01963) (-0.01876,0.00804) (-0.00286,0.01475)
  (-0.01011,0.00901) (-0.00292,0.00395) (-0.08661,-0.01123) (-0.03395,0.00222) (-0.06134,-0.00751)
  (-0.04271,0.00221) (-0.05541,0.01972) (-0.05188,-0.00047) (-0.07357,-0.00275) (-0.08683,-0.01107)
  (-0.00326,0.02303) (-0.01441,-0.00354) (-0.04778,-0.01078) (-0.03054,-0.02069) (-0.06106,-0.01921)
  (-0.01089,0.01347) (0.00516,0.00981) (-0.00738,0.03766) (0.00784,-0.00080) (-0.02711,-0.00271)
  (-0.05170,-0.01257) (-0.06617,-0.02016) (-0.03307,0.00259) (-0.04708,0.00077) (-0.09156,-0.00106)
  (-0.05978,-0.00759) (-0.06013,-0.02323) (-0.02603,0.01274) (0.01152,0.03227) (-0.02515,0.01379)
  (-0.01147,0.01520) (0.02103,0.02535) (-0.02247,0.00234) (0.00944,0.01625) (-0.02537,0.00444)
  (-0.01052,0.01518) (-0.01612,0.00627) (0.01167,0.03710) (-0.04652,-0.01468) (-0.04600,0.00482)
  (-0.03153,-0.01151) (0.00618,0.01411) (-0.05278,-0.00259) (-0.08112,-0.01730) (-0.02517,-0.00653)
  (-0.01594,0.01273) (-0.03522,-0.00817) (-0.05644,0.00373) (-0.04574,-0.00670) (-0.03709,-0.00442)
  (-0.08354,-0.01988) (-0.01261,-0.00149) (-0.03849,0.00796) (-0.02329,0.00700) (-0.02662,0.01201)
  (-0.03421,-0.02869) (-0.04839,-0.01676) (-0.02494,0.01547) (-0.03628,0.00567) (-0.00148,0.00356)
  (-0.03300,0.02247) (-0.07000,0.00247) (-0.02904,0.00459) (0.02032,0.01093) (0.02297,0.01285)
  (-0.00805,0.00801) (0.01144,-0.00043) (-0.01217,-0.00439) (0.00185,0.02504) (-0.03737,0.01837)
  (-0.03751,-0.00651) (0.01176,0.01712) (-0.00766,0.01526) (0.00064,0.02824) (-0.02840,0.02782)
};
\addplot[zline, mag, opacity=0.85, line width=1.3pt, forget plot]
        coordinates {(-0.118,0.00521) (-0.03028,0.00521)};
\addplot[zline, mag, opacity=0.85, line width=1.3pt, forget plot]
        coordinates {(-0.03028,-0.05) (-0.03028,0.00521)};
\addplot[chalo, forget plot] coordinates {(-0.03028,0.00521)};
\addplot[cmark, forget plot] coordinates {(-0.03028,0.00521)};
\addplot[cdot, forget plot]  coordinates {(-0.03028,0.00521)};
\node[font=\scriptsize, anchor=north west, inner sep=2pt] at (axis cs:-0.1145,0.0755)
     {\textcolor{mag}{$(-0.030,\,0.005)$}};
\end{axis}
\end{tikzpicture}
\subcaption{Qwen3-8B rollouts (teacher)}\label{fig:away-lad-8b}
\end{subfigure}
\caption{
\textbf{Reduction in overlap across rollout policies.}
Same quantity as \cref{fig:away-a}, with the rollout policy varied over the Qwen3 family:
(a) 0.6B (\name policy), (b) 1.7B, (c) 4B, and (d) 8B (teacher).
Each point compares trajectory-level NOR under OPD (y-axis) with NOR under the corresponding rollouts (x-axis).
Only 0.6B rollouts place most trajectories below the diagonal and
the paired gap decreases monotonically with the size of the rollout policy.
}
\label{app:away-ladder}
\end{figure}

%% file: figs/heldout.tex
\definecolor{ink}{HTML}{1A1F24}  \definecolor{mut}{HTML}{8A929B}
\definecolor{mag}{HTML}{B03AA0}

\pgfplotsset{
  ptmark/.style={only marks, color=mut, mark=*, mark size=1.1pt,
                 mark options={solid, draw=mut, fill=mut, opacity=0.45}},
  idl/.style={mut, dashed, line width=1.0pt},
  zline/.style={mut, opacity=0.55, dotted, line width=0.8pt},
  chalo/.style={only marks, color=white, mark=*, mark size=3.4pt,
                mark options={draw=white, fill=white, line width=0.6pt}},
  cmark/.style={only marks, color=mag, mark=+, mark size=6.2pt,
                mark options={mag, line width=1.9pt}},
  cdot/.style={only marks, color=mag, mark=*, mark size=2.4pt,
               mark options={draw=white, line width=0.9pt, fill=mag}},
}

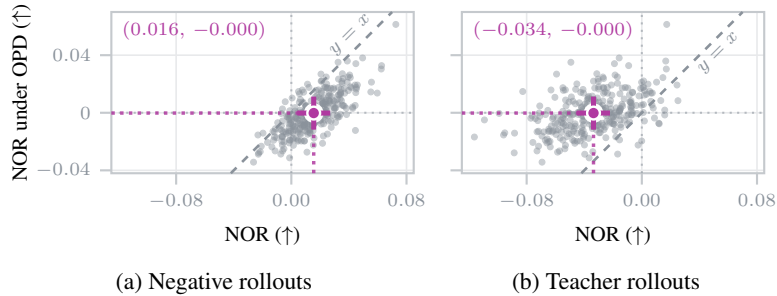
\begin{figure}[h]
\centering
\begin{subfigure}[b]{0.40\textwidth}
\centering
\begin{tikzpicture}
\begin{axis}[ldacc,
  scale only axis=true, width=4.0cm, height=2.133cm,
  xmin=-0.125, xmax=0.085, ymin=-0.042, ymax=0.07,
  xtick={-0.08,0,0.08},
  ytick={-0.04,0,0.04},
  scaled ticks=false,
  xticklabel style={/pgf/number format/fixed, /pgf/number format/precision=2},
  yticklabel style={/pgf/number format/fixed, /pgf/number format/precision=2},
  enlargelimits=false,
  xlabel={NOR ($\uparrow$)},
  ylabel={NOR under OPD ($\uparrow$)},
  ylabel style={yshift=-5pt},
  clip mode=individual,
  label style={font=\footnotesize},
  ticklabel style={font=\scriptsize}
]
\addplot[zline, forget plot] coordinates {(-0.125,0) (0.085,0)};
\addplot[zline, forget plot] coordinates {(0,-0.042) (0,0.07)};
\addplot[idl, forget plot, domain=-0.042:0.07, samples=2] {x}
        node[pos=0.8, sloped, above, mut, font=\scriptsize, inner sep=2pt] {$y=x$};
\addplot[ptmark, forget plot] coordinates {
  (0.01661,-0.00646) (0.00233,-0.00956) (0.01029,0.00515) (0.00687,-0.00220) (0.03678,0.01691)
  (0.02339,0.00044) (0.00308,-0.00357) (0.02508,-0.01479) (0.01576,-0.01755) (0.03083,0.00336)
  (0.00125,-0.01282) (0.02120,0.00826) (0.02623,0.00680) (-0.00205,-0.01396) (0.00462,-0.01973)
  (-0.00614,-0.00573) (0.02077,-0.00789) (0.01305,-0.01388) (0.00950,-0.01229) (0.02221,0.01052)
  (0.02199,-0.00366) (-0.00345,-0.01454) (0.00138,-0.01776) (0.03813,0.00315) (0.01131,-0.00773)
  (0.03278,0.01760) (0.01401,0.00121) (0.01673,0.00494) (0.01466,0.00478) (0.00817,0.00056)
  (0.03480,0.00844) (0.01482,-0.00755) (0.02581,0.01158) (0.00111,-0.00959) (-0.00654,-0.01990)
  (0.01493,0.00098) (0.00500,0.01477) (0.03544,0.02711) (0.01086,-0.02736) (0.00483,-0.01335)
  (0.05969,0.01368) (0.03603,0.00511) (0.03196,-0.00204) (0.02687,0.00805) (0.06247,0.03039)
  (0.03600,0.02378) (0.00626,0.00922) (0.00550,-0.01190) (0.03392,0.00959) (0.02115,0.00081)
  (0.02578,-0.00577) (0.02154,0.00623) (0.01960,-0.00545) (-0.00114,-0.01903) (0.02932,0.00849)
  (0.01028,0.00465) (-0.00050,-0.00758) (0.00891,-0.00867) (0.01243,-0.00608) (0.00143,-0.02261)
  (0.02671,0.01221) (0.00281,-0.00928) (0.02604,0.01374) (0.02765,-0.00571) (0.02286,0.01760)
  (0.03300,0.01246) (-0.02292,-0.02879) (0.01194,-0.00206) (0.01356,0.00166) (0.00333,-0.00402)
  (0.02977,0.00344) (0.04049,0.02624) (0.02359,0.00612) (0.01594,-0.00766) (0.01664,0.01100)
  (0.02945,0.01717) (0.05565,0.02127) (0.03190,0.00925) (0.00113,-0.00832) (-0.00642,-0.01656)
  (0.02296,0.00042) (-0.01350,-0.02371) (0.04255,0.00401) (0.01509,-0.01249) (0.02070,0.02195)
  (0.00470,-0.01894) (-0.00040,-0.00577) (0.02398,0.00885) (0.03675,0.00468) (0.01110,-0.00408)
  (-0.00318,-0.00254) (0.01937,-0.00133) (0.03278,0.02340) (0.03889,0.03593) (0.00567,-0.00155)
  (0.02240,-0.00765) (0.00502,-0.00733) (0.00858,-0.00867) (0.01393,-0.00498) (0.01450,-0.00884)
  (0.01795,-0.00048) (0.02329,-0.00510) (-0.00607,-0.01679) (0.00124,-0.01384) (-0.00165,-0.00768)
  (-0.00204,-0.00154) (0.00795,-0.00547) (0.01382,0.00832) (0.02558,-0.00070) (0.03227,-0.00313)
  (0.01562,-0.00727) (0.01537,0.00019) (-0.01273,-0.01297) (0.02202,0.00692) (0.02618,0.00839)
  (0.04096,0.02444) (0.00803,0.00747) (0.02376,0.00572) (0.02528,0.01719) (-0.02086,-0.02323)
  (0.04515,0.01650) (-0.02625,-0.03431) (0.03521,0.02049) (0.03836,-0.00123) (0.03014,0.00312)
  (-0.00701,-0.00046) (0.03878,0.02122) (0.01575,-0.00746) (0.04588,0.01929) (0.02226,-0.01615)
  (0.03413,0.01349) (0.02373,-0.00146) (0.03328,-0.00276) (0.00888,-0.01963) (0.03231,0.00589)
  (0.01440,-0.00885) (0.04269,0.00307) (0.00696,-0.01961) (0.02355,0.00741) (0.00596,-0.01100)
  (0.03067,-0.00665) (-0.00540,-0.01335) (-0.00425,0.00767) (0.02436,-0.00126) (0.02537,-0.00209)
  (0.02015,0.00541) (0.00260,-0.01180) (0.06269,0.03275) (0.01518,0.00058) (0.01375,0.00269)
  (0.00762,0.01583) (0.04951,0.03808) (0.01520,-0.00118) (0.01548,-0.00317) (0.01072,-0.01320)
  (-0.00626,-0.02238) (-0.01230,-0.01907) (0.02955,0.00212) (0.01761,-0.00187) (0.00974,0.00171)
  (0.00290,-0.00307) (0.02168,-0.02120) (-0.01170,-0.01684) (-0.00606,-0.02650) (-0.01968,-0.01415)
  (0.02218,0.00647) (-0.00053,-0.01280) (0.02114,0.02210) (0.02858,0.00622) (0.03453,0.01631)
  (0.00942,-0.00802) (-0.00118,-0.03132) (0.01017,-0.00154) (0.00950,-0.00561) (0.03513,0.02174)
  (0.04436,0.00509) (0.03246,0.03508) (0.00412,-0.00439) (0.02372,0.01781) (0.01776,0.01168)
  (-0.00106,0.00142) (0.02106,0.01498) (0.01550,-0.00543) (0.00630,-0.00742) (0.03424,0.01415)
  (0.00079,-0.01816) (0.00256,0.00360) (-0.00618,0.00268) (-0.00261,-0.00288) (0.02412,-0.00019)
  (0.02064,-0.00069) (0.00740,-0.00271) (0.00616,0.00014) (0.01290,-0.01164) (0.02483,0.01722)
  (0.00416,-0.00012) (0.03935,0.02489) (0.02746,0.01698) (0.01589,-0.00154) (0.01920,0.01073)
  (0.03267,0.01233) (0.01881,0.01589) (0.03531,0.00529) (-0.00730,-0.02870) (0.03395,-0.00100)
  (0.00694,-0.00839) (-0.00106,-0.01883) (0.01181,-0.00908) (-0.00400,-0.01322) (-0.00173,-0.00378)
  (0.00323,-0.01778) (0.00610,0.00172) (0.01896,-0.00232) (0.01704,0.01557) (0.00771,-0.00962)
  (0.01296,-0.01228) (0.00856,0.01427) (-0.02160,-0.02703) (0.00056,0.01025) (-0.00153,-0.02093)
  (0.04477,0.01398) (0.05521,0.01446) (0.02453,0.02005) (0.02506,0.00414) (0.01978,0.00322)
  (0.00525,-0.00534) (-0.01110,-0.01106) (0.01188,-0.00711) (0.01225,0.00301) (0.01014,0.00476)
  (-0.02265,-0.02785) (0.00649,-0.01505) (0.04170,0.01740) (-0.01182,-0.01330) (0.00920,-0.01956)
  (0.01183,-0.00106) (0.02596,0.00905) (-0.00514,-0.02503) (0.02403,0.00186) (-0.00509,-0.02019)
  (-0.00651,-0.01633) (0.00095,-0.00098) (0.02800,0.01140) (0.02288,0.00226) (0.01260,-0.00001)
  (0.03702,0.01364) (-0.00171,-0.01259) (0.00033,-0.00498) (0.02083,0.00967) (0.00262,0.00117)
  (0.02216,0.00455) (0.01224,0.01761) (0.02088,0.01361) (0.01294,-0.01424) (-0.00797,-0.01139)
  (0.00555,0.00721) (0.02934,0.01620) (0.02417,0.00186) (0.04422,0.01927) (0.00973,0.00419)
  (-0.02321,-0.01544) (0.02423,0.02485) (0.04481,0.03411) (-0.00150,-0.01026) (-0.00135,-0.00503)
  (0.01809,0.00206) (-0.01653,-0.02060) (0.03001,0.00770) (0.02901,-0.00182) (0.00914,-0.00937)
  (0.03416,0.01284) (0.02768,-0.01012) (0.02079,-0.00052) (0.01193,-0.01134) (0.02045,-0.00503)
  (0.04516,0.01762) (-0.00946,-0.00690) (0.00819,-0.00013) (0.01713,-0.00275) (-0.01015,-0.01732)
  (0.00983,-0.00002) (0.01170,0.00019) (0.01139,-0.01494) (0.00604,-0.01434) (0.04452,-0.01227)
  (0.01529,-0.02246) (-0.01126,-0.02651) (0.00988,-0.00575) (0.01360,0.00257) (0.01123,0.01987)
  (0.00931,-0.00116) (-0.00687,-0.00905) (0.00568,0.01317) (0.03962,0.01545) (0.07258,0.06138)
  (0.00932,-0.00991) (0.01640,0.00797) (-0.00682,-0.01209) (0.02059,0.02686) (0.04103,0.01809)
};
\addplot[zline, mag, opacity=0.85, line width=1.3pt, forget plot]
        coordinates {(-0.125,-0.00020) (0.01555,-0.00020)};
\addplot[zline, mag, opacity=0.85, line width=1.3pt, forget plot]
        coordinates {(0.01555,-0.042) (0.01555,-0.00020)};
\addplot[chalo, forget plot] coordinates {(0.01555,-0.00020)};
\addplot[cmark, forget plot] coordinates {(0.01555,-0.00020)};
\addplot[cdot, forget plot]  coordinates {(0.01555,-0.00020)};
\node[font=\scriptsize, anchor=north west, inner sep=2pt] at (axis cs:-0.1215,0.0675)
     {\textcolor{mag}{$(0.016,\,-0.000)$}};
\end{axis}
\end{tikzpicture}
\subcaption{\Name rollouts}\label{fig:away-ho-ours}
\end{subfigure}
\begin{subfigure}[b]{0.33\textwidth}
\centering
\begin{tikzpicture}
\begin{axis}[ldacc,
  scale only axis=true, width=4.0cm, height=2.133cm,
  xmin=-0.125, xmax=0.085, ymin=-0.042, ymax=0.07,
  xtick={-0.08,0,0.08},
  ytick={-0.04,0,0.04},
  scaled ticks=false,
  xticklabel style={/pgf/number format/fixed, /pgf/number format/precision=2},
  yticklabel style={/pgf/number format/fixed, /pgf/number format/precision=2},
  enlargelimits=false,
  xlabel={NOR ($\uparrow$)},
  yticklabels={},
  clip mode=individual,
  label style={font=\footnotesize},
  ticklabel style={font=\scriptsize}
]
\addplot[zline, forget plot] coordinates {(-0.125,0) (0.085,0)};
\addplot[zline, forget plot] coordinates {(0,-0.042) (0,0.07)};
\addplot[idl, forget plot, domain=-0.042:0.07, samples=2] {x}
        node[pos=0.8, sloped, below, mut, font=\scriptsize, inner sep=2pt] {$y=x$};
\addplot[ptmark, forget plot] coordinates {
  (-0.00848,-0.00646) (-0.05476,-0.00956) (-0.03304,0.00515) (-0.07616,-0.00220) (-0.00080,0.01691)
  (-0.03030,0.00044) (-0.05487,-0.00357) (-0.04901,-0.01479) (-0.05669,-0.01755) (-0.02496,0.00336)
  (-0.07177,-0.01282) (-0.02701,0.00826) (-0.00740,0.00680) (-0.09519,-0.01396) (-0.04488,-0.01973)
  (-0.02431,-0.00573) (-0.03617,-0.00789) (-0.02966,-0.01388) (-0.04077,-0.01229) (-0.03747,0.01052)
  (0.00877,-0.00366) (-0.05061,-0.01454) (-0.04057,-0.01776) (-0.01500,0.00315) (-0.01947,-0.00773)
  (0.00074,0.01760) (-0.01555,0.00121) (-0.02959,0.00494) (-0.00682,0.00478) (-0.02191,0.00056)
  (-0.02179,0.00844) (-0.05853,-0.00755) (-0.03444,0.01158) (-0.04887,-0.00959) (-0.06376,-0.01990)
  (-0.05490,0.00098) (-0.08271,0.01477) (-0.03126,0.02711) (-0.05850,-0.02736) (-0.06222,-0.01335)
  (-0.04053,0.01368) (-0.02732,0.00511) (-0.02446,-0.00204) (-0.04786,0.00805) (0.01700,0.03039)
  (-0.02661,0.02378) (-0.03485,0.00922) (-0.03154,-0.01190) (-0.01230,0.00959) (-0.01549,0.00081)
  (-0.03465,-0.00577) (-0.04609,0.00623) (-0.03163,-0.00545) (-0.05480,-0.01903) (0.01318,0.00849)
  (-0.01334,0.00465) (-0.02033,-0.00758) (-0.02672,-0.00867) (-0.02257,-0.00608) (-0.05303,-0.02261)
  (-0.02079,0.01221) (-0.04621,-0.00928) (0.00730,0.01374) (0.00178,-0.00571) (-0.02674,0.01760)
  (-0.00860,0.01246) (-0.03967,-0.02879) (-0.01556,-0.00206) (-0.06529,0.00166) (-0.07403,-0.00402)
  (-0.01780,0.00344) (-0.01859,0.02624) (-0.03942,0.00612) (-0.00996,-0.00766) (-0.04893,0.01100)
  (-0.01361,0.01717) (0.01023,0.02127) (-0.03318,0.00925) (-0.04048,-0.00832) (-0.05201,-0.01656)
  (-0.04374,0.00042) (-0.06360,-0.02371) (-0.02895,0.00401) (-0.06167,-0.01249) (-0.03356,0.02195)
  (-0.04803,-0.01894) (-0.03640,-0.00577) (-0.05296,0.00885) (-0.05267,0.00468) (-0.06200,-0.00408)
  (-0.07094,-0.00254) (-0.00929,-0.00133) (-0.03032,0.02340) (0.01661,0.03593) (-0.05419,-0.00155)
  (0.01207,-0.00765) (-0.02749,-0.00733) (-0.03015,-0.00867) (-0.01505,-0.00498) (-0.05034,-0.00884)
  (-0.01951,-0.00048) (-0.05620,-0.00510) (-0.01807,-0.01679) (-0.02885,-0.01384) (-0.04635,-0.00768)
  (-0.01765,-0.00154) (-0.07228,-0.00547) (-0.00752,0.00832) (-0.02604,-0.00070) (-0.02097,-0.00313)
  (0.01175,-0.00727) (0.00100,0.00019) (-0.06486,-0.01297) (-0.04345,0.00692) (0.02509,0.00839)
  (0.01463,0.02444) (-0.05058,0.00747) (0.00109,0.00572) (-0.03268,0.01719) (-0.07171,-0.02323)
  (-0.03005,0.01650) (-0.06930,-0.03431) (-0.01690,0.02049) (-0.03005,-0.00123) (-0.02547,0.00312)
  (-0.09911,-0.00046) (-0.00017,0.02122) (-0.06029,-0.00746) (-0.02112,0.01929) (-0.06938,-0.01615)
  (-0.02656,0.01349) (-0.00692,-0.00146) (-0.01975,-0.00276) (-0.04282,-0.01963) (-0.01382,0.00589)
  (-0.05293,-0.00885) (-0.00730,0.00307) (-0.00008,-0.01961) (-0.02966,0.00741) (-0.03434,-0.01100)
  (-0.03027,-0.00665) (-0.05572,-0.01335) (-0.03228,0.00767) (-0.06688,-0.00126) (-0.04045,-0.00209)
  (0.00706,0.00541) (-0.05804,-0.01180) (-0.00915,0.03275) (-0.01797,0.00058) (-0.02145,0.00269)
  (-0.06308,0.01583) (-0.01573,0.03808) (-0.04425,-0.00118) (-0.05850,-0.00317) (-0.04192,-0.01320)
  (-0.03986,-0.02238) (-0.06702,-0.01907) (-0.02740,0.00212) (-0.03482,-0.00187) (-0.01542,0.00171)
  (-0.02004,-0.00307) (-0.04651,-0.02120) (-0.11630,-0.01684) (-0.07280,-0.02650) (-0.08840,-0.01415)
  (-0.03780,0.00647) (-0.04754,-0.01280) (0.00125,0.02210) (-0.02218,0.00622) (-0.04596,0.01631)
  (-0.01940,-0.00802) (-0.03661,-0.03132) (0.01591,-0.00154) (-0.04033,-0.00561) (-0.03024,0.02174)
  (-0.00382,0.00509) (-0.04927,0.03508) (-0.05404,-0.00439) (0.00135,0.01781) (-0.00197,0.01168)
  (-0.00751,0.00142) (0.00539,0.01498) (-0.03262,-0.00543) (-0.05448,-0.00742) (-0.04244,0.01415)
  (-0.06568,-0.01816) (-0.04191,0.00360) (-0.03055,0.00268) (-0.06150,-0.00288) (-0.05084,-0.00019)
  (-0.05322,-0.00069) (-0.04060,-0.00271) (-0.04449,0.00014) (-0.02625,-0.01164) (-0.00972,0.01722)
  (-0.05056,-0.00012) (-0.01697,0.02489) (-0.02642,0.01698) (-0.00744,-0.00154) (-0.06037,0.01073)
  (0.00181,0.01233) (-0.05309,0.01589) (-0.00024,0.00529) (-0.01679,-0.02870) (-0.04989,-0.00100)
  (-0.01489,-0.00839) (-0.01546,-0.01883) (-0.06285,-0.00908) (-0.07279,-0.01322) (-0.07708,-0.00378)
  (-0.03696,-0.01778) (-0.03546,0.00172) (-0.01070,-0.00232) (-0.05450,0.01557) (-0.03436,-0.00962)
  (-0.05251,-0.01228) (-0.00642,0.01427) (-0.03405,-0.02703) (-0.03087,0.01025) (-0.06594,-0.02093)
  (-0.00645,0.01398) (-0.02551,0.01446) (-0.04512,0.02005) (-0.00775,0.00414) (0.00792,0.00322)
  (-0.01948,-0.00534) (-0.04687,-0.01106) (-0.02215,-0.00711) (-0.03490,0.00301) (-0.07089,0.00476)
  (-0.08864,-0.02785) (-0.05958,-0.01505) (-0.05483,0.01740) (-0.06811,-0.01330) (-0.04699,-0.01956)
  (-0.04736,-0.00106) (-0.03788,0.00905) (-0.05345,-0.02503) (0.01591,0.00186) (-0.06465,-0.02019)
  (-0.04022,-0.01633) (-0.02369,-0.00098) (-0.03689,0.01140) (-0.04095,0.00226) (-0.00630,-0.00001)
  (0.00637,0.01364) (-0.06408,-0.01259) (-0.10933,-0.00498) (-0.03700,0.00967) (-0.04074,0.00117)
  (-0.03810,0.00455) (-0.01150,0.01761) (-0.02382,0.01361) (-0.04490,-0.01424) (-0.02873,-0.01139)
  (-0.00801,0.00721) (-0.04085,0.01620) (-0.02629,0.00186) (0.02442,0.01927) (-0.04175,0.00419)
  (-0.09604,-0.01544) (-0.02704,0.02485) (-0.01903,0.03411) (-0.04176,-0.01026) (-0.04977,-0.00503)
  (-0.03581,0.00206) (-0.07567,-0.02060) (-0.00833,0.00770) (-0.04195,-0.00182) (-0.03176,-0.00937)
  (0.00322,0.01284) (-0.01451,-0.01012) (-0.00339,-0.00052) (-0.06465,-0.01134) (-0.01215,-0.00503)
  (-0.02392,0.01762) (-0.05584,-0.00690) (-0.07130,-0.00013) (-0.03631,-0.00275) (-0.02498,-0.01732)
  (-0.02209,-0.00002) (-0.04343,0.00019) (-0.03973,-0.01494) (-0.04174,-0.01434) (-0.01856,-0.01227)
  (-0.00261,-0.02246) (-0.05718,-0.02651) (-0.04107,-0.00575) (-0.01477,0.00257) (0.00428,0.01987)
  (-0.04818,-0.00116) (-0.03059,-0.00905) (-0.01521,0.01317) (0.00200,0.01545) (0.01730,0.06138)
  (-0.02543,-0.00991) (0.00625,0.00797) (-0.05963,-0.01209) (-0.05923,0.02686) (-0.02995,0.01809)
};
\addplot[zline, mag, opacity=0.85, line width=1.3pt, forget plot]
        coordinates {(-0.125,-0.00020) (-0.03356,-0.00020)};
\addplot[zline, mag, opacity=0.85, line width=1.3pt, forget plot]
        coordinates {(-0.03356,-0.042) (-0.03356,-0.00020)};
\addplot[chalo, forget plot] coordinates {(-0.03356,-0.00020)};
\addplot[cmark, forget plot] coordinates {(-0.03356,-0.00020)};
\addplot[cdot, forget plot]  coordinates {(-0.03356,-0.00020)};
\node[font=\scriptsize, anchor=north west, inner sep=2pt] at (axis cs:-0.1215,0.0675)
     {\textcolor{mag}{$(-0.034,\,-0.000)$}};
\end{axis}
\end{tikzpicture}
\subcaption{Teacher rollouts}\label{fig:away-ho-8b}
\end{subfigure}
\caption{
\textbf{Reduction in overlap with the \name policy on held-out prompts.}
Same quantity as \cref{fig:away-a}, measured on 300 held-out prompts drawn from nine
benchmarks spanning mathematics, science, and code, none of which appear in training.
Each point is one trajectory, comparing NOR under OPD (y-axis) with (a) \name-policy or (b) teacher rollouts (x-axis).
}
\label{app:away-heldout}
\end{figure}